\documentclass[pmlr]{jmlr}

\usepackage{longtable}%
\usepackage[utf8]{inputenc} %
\usepackage[T1]{fontenc}    %
\usepackage{hyperref}       %
\usepackage{url}            %
\usepackage{booktabs}       %
\usepackage{amsfonts}       %
\usepackage{nicefrac}       %
\usepackage{microtype}      %
\usepackage{graphicx}
\usepackage{amsmath}
\usepackage{hhline}
\usepackage{wrapfig}
\usepackage{multirow}
\usepackage{dsfont}
\usepackage{overpic}
\usepackage{contour}
\contourlength{1pt}
\usepackage{xspace}
\usepackage{sidecap}
\usepackage{rotating}
\usepackage[normalem]{ulem}
\usepackage[font=small,labelfont=bf]{caption}
\usepackage{subcaption}
\usepackage{cleveref}
\usepackage{microtype}
\usepackage{enumitem}
\usepackage{wrapfig}
\usepackage{nicefrac}
\usepackage{algpseudocode}
\usepackage{apptools}
\usepackage[super]{nth}
\usepackage{comment}

\usepackage{booktabs}
\usepackage[load-configurations=version-1]{siunitx} %

\theorembodyfont{\upshape}
\theoremheaderfont{\scshape}
\theorempostheader{:}
\theoremsep{\newline}

\jmlrvolume{}
\firstpageno{1}
\editors{Sophia Sanborn, Christian Shewmake, Simone Azeglio, Nina Miolane}

\jmlryear{2023}
\jmlrworkshop{Symmetry and Geometry in Neural Representations}

\title[Explicit Neural Surfaces]{Explicit Neural Surfaces: Learning Continuous Geometry with Deformation Fields}

 \author{\Name{Thomas Walker} \Email{T.M.Walker@sms.ed.ac.uk}\\
 \Name{Octave Mariotti} \Email{omariott@ed.ac.uk}\\
 \Name{Amir Vaxman} \Email{avaxman@ed.ac.uk}\\
 \Name{Hakan Bilen} \Email{h.bilen@ed.ac.uk}\\
 \addr University Of Edinburgh}

\begin{document}
\pagenumbering{gobble}

\maketitle

\begin{abstract}
We introduce Explicit Neural Surfaces (ENS), an efficient smooth surface representation that directly encodes topology with a deformation field from a known base domain. We apply this representation to reconstruct explicit surfaces from multiple views, where we use a series of neural deformation fields to progressively transform the base domain into a target shape. By using meshes as discrete surface proxies, we train the deformation fields through efficient differentiable rasterization. Using a fixed base domain allows us to have Laplace-Beltrami eigenfunctions as an intrinsic positional encoding alongside standard extrinsic Fourier features, with which our approach can capture fine surface details. Compared to implicit surfaces, ENS trains faster and has several orders of magnitude faster inference times. The explicit nature of our approach also allows higher-quality mesh extraction whilst maintaining competitive surface reconstruction performance and real-time capabilities.
\end{abstract}

\section{Introduction}\label{sec:intro}
Reconstructing 3D objects from multiple-view images is a fundamental problem in computer vision and graphics. 
Classical 3D reconstruction methods~\citep{hartley2003multiple,furukawa2009accurate,agarwal2011building} follow a multi-stage pipeline, match either handcrafted or learned features across images, and then recover their 3D coordinates through mesh generation. Their success heavily relies on correctly identifying matching features from different images, where accumulated matching errors cannot be easily fixed in the later stage. A recent promising paradigm is based on analysis-by-synthesis, learning neural representations for 3D such that their rendering matches the image collection.

Successful methods in this group use continuous volumetric representations~\citep{nerf, NeuS, volsdf, yariv2020multiview} and render them in a differentiable manner. 
Neural radiance fields (NeRFs)~\citep{nerf} and its variants represent geometry using density fields obtained from volume rendering.
However, density-based geometries have undefined surface boundaries and therefore lead to noisy extracted surfaces. 
To mitigate this problem, recent techniques~\citep{NeuS, volsdf} use an implicit signed distance function within this volume rendering framework to recover exact geometries as level sets.
Despite their good performance in surface reconstruction, these methods have at least two important shortcomings.
First, they are slow to train, render views, and extract meshes from, since using an implicit function necessitates expensive sampling strategies to locate the surface.

Second, in order to extract meshes, implicit surfaces require a post-processing step such as marching cubes~\citep{lorensen1987marching}. However, this exposes such methods to reconstruction artifacts, generates an excessive number of elements, and produces low-quality meshing with grid aliasing. Whilst not being directly reflected in surface reconstruction performance, these drawbacks prevent neural implicit surfaces from being integrated into real-time geometric pipelines, and can be problematic for finite-element-methods (FEM) which are very sensitive to element quality \citep{shewchuk2002good}.
A possible mitigation is to explicitly learn a mesh through differentiable rasterization~\citep{thies2019deferred,worchel2022multi}. 
Such methods achieve promising speed/performance trade-offs and can be integrated into real-time mesh-based graphics engines which are widely used. 
However, the flexibility and reconstruction quality are limited by a fixed discrete mesh and heavily rely on the presence of strong regularization techniques~\citep{Nicolet2021Large}. 

Motivated by these shortcomings, we introduce an \emph{explicit} neural surface (ENS) representation based on a neural deformation field that maps an initial continuous domain (e.g., the unit sphere) into a target surface in a progressive coarse-to-fine strategy. The representation is trained through sampling discrete meshes, that are efficiently rendered by a jointly learned neural deferred shader.
Our approach comprises three unique components. First, our deformation field is smooth, and is decoupled from domain mesh resolution and connectivity. Secondly, we benefit from our explicit geometry by extracting shape-aware \textit{intrinsic} positional encodings to improve rendering and reconstruction performance. Specifically, we use a hybrid positional encoding mixing intrinsic Laplace-Beltrami eigenfunctions with extrinsic Fourier features. Finally, we sequentially combine neural deformation fields with progressively higher-frequency positional encodings, enabling gradual surface refinement through both coarse and then fine deformation fields.
The resulting method attains an \textit{explicitly} defined neural surface which demonstrates superior reconstruction performance to prior explicit methods while being a fully continuous representation. Compared to the implicit neural surfaces, our method is significantly faster at inference time while being competitive in reconstruction performance. In addition, our explicit surface definition allows for direct surface sampling, enabling higher-quality intrinsic remeshing across resolutions and with arbitrary connectivity, without requiring post-processing techniques or strong regularization.

\begin{figure}[t]
    \centering
    \includegraphics[width=0.75\textwidth]{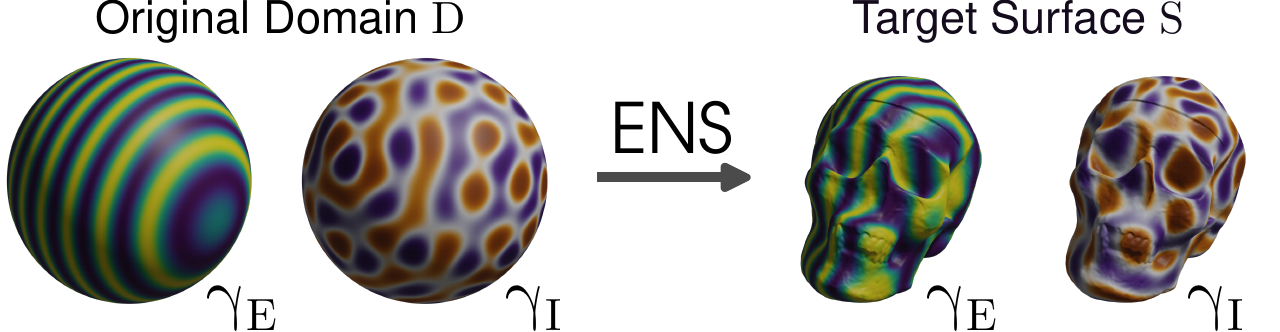}
    \caption{ENS uses both an extrinsic positional encoding $\gamma_E$ and an intrinsic positional encoding $\gamma_I$ of the original domain. By combining ambient as well as shape-aware surface embeddings we can attain fine surface details and textures.}
    \label{fig:embeddings}
\end{figure}

\section{Related Work}\label{sec:relwork}
\paragraph{Neural Surface Representations} Neural networks in 3D reconstruction have recently gained significant popularity for their ability to model continuous geometry. Numerous methods successfully represent surfaces as level sets of signed distance functions (SDF)~\citep{NeuS, yariv2020multiview,volsdf,Oechsle2021ICCV}. An \textit{implicit} surface is appealing as it can model arbitrary topological changes. NeuS and other implicit surface-based variants~\citep{NeuS, volsdf,Oechsle2021ICCV} have been successfully integrated into volume rendering pipelines. However, implicit surfaces cannot be directly sampled, and therefore necessitate expensive volumetric procedures to train, render, and mesh. Concurrently, PermutoSDF~\citep{rosu2023permutosdf} used learned spatial hash-encodings to facilitate shallower MLPs and greatly accelerate training times. However, expensive spatial queries are still necessary to render images and extract meshes. A separate line of literature proposed learning an atlas of neural parametric surfaces~\citep{atlas, deepgeomprior, diffsurfrep} for 3D reconstruction. This surface definition can facilitate sampling surfaces directly by virtue of a parameter space, however, using multiple disjoint mappings requires stitching together surface patches (softly), leading to poor-quality reconstructions. In contrast, our method enforces an initial topology as a prior and thus guarantees watertight surfaces, explicitly represented by a deformation field.

\paragraph{Neural Deferred Shading} Deferred shading is a real-time rendering approach based on meshes, in which the shading component is performed entirely in screen-space \citep{deering1988triangle}. The scene is rasterized to create a screen-space map containing geometric information, and then processed by a shader to predict pixel-wise RGB values. In a recent work \citep{dmt_sdf}, shade meshes were extracted from an implicit SDF with Deep Marching Tetrahedra (DMTet) \citep{shen2021dmtet}. While the deferred shading is faster relative to volume rendering, the need to use DMTet as a mesh-extraction step still limits the efficiency of their approach. Furthermore, the use of shaders that cannot handle arbitrary lighting limits their application to settings with fixed global illumination. Another approach \citep{thies2019deferred} proposed fully parameterizing a deferred shader using a neural network, and hence synthesizing novel views with arbitrary illumination and materials. \citet{worchel2022multi} (NDS) subsequently proposed to jointly optimize the mesh with the neural shader, and attain an efficient multi-view surface reconstruction pipeline. In our work, we extend this further to attain the benefits of neural surface representations. 

\section{Method}\label{sec:method}
\begin{figure}[t]
    \centering
    \includegraphics[width=\textwidth]{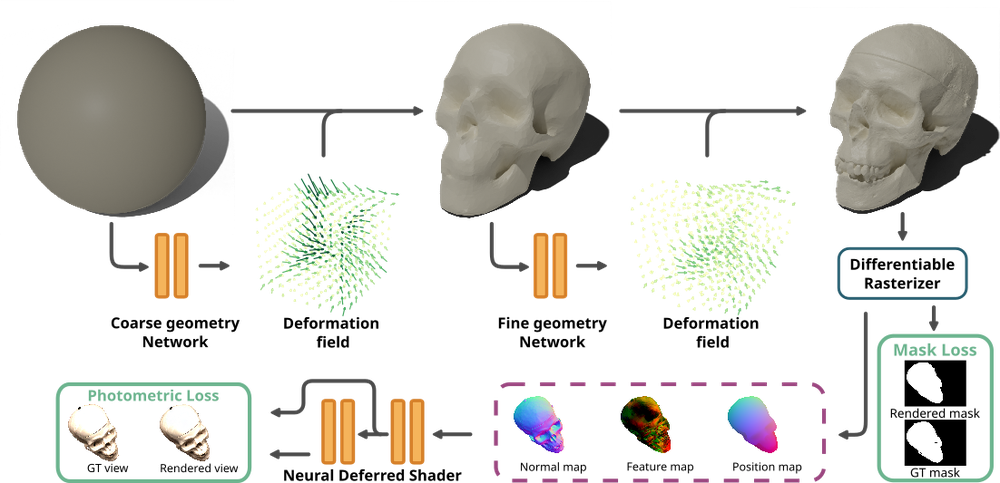}
    \vspace{5pt}
    \caption{The fixed base domain $D$ (here the unit sphere) is transformed by a low-frequency (top middle) and then a high-frequency neural deformation field, $f_{\text{coarse}}$ and $f_{\text{fine}}$ respectively, to create a surface (top-right skull). The final mesh is obtained by pullback to a meshing of the unit sphere. During training, such a mesh is rasterized and processed through a neural shader to produce a candidate image $\tilde{I}_i$ that is compared against a GT image $I_i$.}
    \label{fig:arch}
\end{figure}

Our objective is to learn a continuous surface representation $S$ of a specified object in a scene. As input, we consider $N$ images, along with their corresponding cameras, $\mathcal{I} = \{I_i, C_i\}^N_{i=1}$ and binary masks $M=\{m_i\}^N_{i=1}$ segmenting the object of interest. Our architecture and pipeline are depicted in \Cref{fig:arch}. We first introduce our surface representation, and then our training procedure.

\subsection{Neural Deformation Field}
We compute an ambient mapping $f:\mathbb{R}^3 \to \mathbb{R}^3$. 
This is used to \textit{explicitly} define a continuous surface $S$ when applied to a known and fixed continuous input domain $D \subset \mathbb{R}^3$, \mbox{\emph{i.e.} $S = \{f(\textbf{x}) | \textbf{x} \in D\}$}.
We parameterize $f$ by a deformation neural field $f_\theta(\textbf{x})$, with parameters $\theta$. We use meshes $\mathcal{M}_D=\left\{V_D,E,F\right\}$ defined on $D$, that are deformed into compatible meshes $\mathcal{M}_S=\left\{V_S,E,F\right\}$ on $S$. They are then used for training $f$, where we employ fast rasterization and deferred shading, in order to produce predicted images $\tilde{I}_i$ (\Cref{sec:shading}). This representation effectively decouples the geometry of $S$, determined only by the network parameters $\theta$, from any specific discrete mesh $\mathcal{M}$. Thus, it allows us to freely construct (or refine) any mesh $\mathcal{M}_D$ for the purpose of training $f$. Extracting a mesh during training and inference is extremely fast, using only a single forward pass of the deformation network, while the continuity of the mapping allows for meshes of any chosen size or connectivity~(\figureref{fig:connectivity}). Targeting $S$ with spherical topology, we set $D$ to be the unit sphere, from which sampling can be done in simple closed form. Throughout training we sample meshes $\mathcal{M}_D$ (and consequently $\mathcal{M}_S$) of increased vertex density; please see appendix for details. 

\subsection{Intrinsic/Extrinsic Surface Embeddings}
\label{sec:embeddings}
Due to the low-frequency spectral bias of neural networks on low-dimensional inputs \citep{gfft, nerf}, we use positional encodings of the input domain $D$ to learn fine surface deformations. Standard implicit representations \citep{nerf,NeuS,volsdf} encode positional coordinates with respect to a spectral basis defined on the ambient Euclidean space. 
Since $D$ is a known base domain in our case, we can augment the extrinsic encoding with an \textit{intrinsic} spectral basis (as the Laplacian eigenfunctions~\cite{levy2006laplace,koestler2022intrinsic}) of the input domain $D$. We approximate the continuous eigenfunctions of $D$ by a specific fine mesh $M_D$  with piecewise linear functions. For this mesh, we construct the cotan Laplacian $W$, and the Voronoi mass matrix $A$, and extract the eigenbasis $\left\{\phi_i\right\}$ that solves $W\phi_i = \lambda_i A \phi_i$ for eigenvalues $\lambda_i$. Following ~\citet{koestler2022intrinsic,rustamov2007gps}, we define an intrinsic embedding $\gamma_I: D \to \mathbb{R}^d$ as 

where $\phi_1, \dots, \phi_d$ are the lowest $d$ eigenfunctions on $D$. As we demonstrate in \Cref{fig:intrinsic} and \Cref{fig:ablations}, the deformation network benefits from having an intrinsic encoding to learn fine surface details and produce higher-quality renders. However, using solely intrinsic information only captures the metric information of the surface, and would prevent the deformation network from having awareness of location of $D$ in the scene. For this reason we use a hybrid embedding $\gamma_H = [\gamma_I, \gamma_E]$, combining both intrinsic $\gamma_I$ and extrinsic positional embeddings $\gamma_E$ based on random Fourier feature (RFF) encoding $\gamma_E: \mathbb{R}^3 \to \mathbb{R}^d$ defined as
\begin{equation}
    \gamma_E(\textbf{x}) = [\cos(\textbf{b}_1^{\top} \textbf{x}), \sin(\textbf{b}_1^{\top} \textbf{x}), \dots, \cos(\textbf{b}_{d/2}^{\top} \textbf{x}), \sin(\textbf{b}_{d/2}^{\top} \textbf{x})],
\end{equation}
where coefficients $\textbf{b}_i \in \mathbb{R}^3$ are sampled randomly from a multivariate Gaussian distribution. By controlling the standard deviation $\sigma$, one can control the distribution of basis function frequencies to target low or high-frequency deformations.  

\subsection{Coarse-To-Fine Optimization}
\label{sec:coarse-to-fine}
Surface fitting can fall into bad minima in the early stages of training \citep{Nicolet2021Large}. This makes it problematic to bias deformation fields towards fine details, as it introduces high-entropy noise into the optimization process. To address this, we construct a coarse-to-fine optimization approach by decomposing the deformation field into a composition of two mappings. The first, $f_{\text{coarse}}$, comprises a low-frequency extrinsic RFF encoding $\gamma_E$ to learn a correctly centered general shape. The second, $f_{\text{fine}}$,  uses a high-frequency hybrid encoding $\gamma_H = [\gamma_E, \gamma_I]$ to learn fine details:
\begin{equation}
f(\textbf{x}) = f_{\text{fine}} \circ \gamma_H \circ f_{\text{coarse}}  \circ \gamma_E(\textbf{x}).
\end{equation}
By removing the onus of learning low frequencies from $f_{\text{fine}}$, we obtain a stable coarse-to-fine optimization process. While we could potentially use more functions in the composition, we found the two-stage deformation is sufficient to capture surface detail in our scenes.

\subsection{Neural Deferred Shader}
\label{sec:shading}
Similar to NDS~\citep{worchel2022multi}, we render images using the real-time graphics approach known as \emph{deferred shading}. For a mesh $\mathcal{M}_S$ predicted by the deformation field, given a camera $C_i$, we rasterize and shade it into an image $\tilde{I}_i$ that is compared against the ground-truth image $I_i$. %
We diverge from NDS, which considers strictly mesh data for shading, to benefit from our neural deformation field $f_\theta(\textbf{x})$. Namely, we extending it to produce a feature vector $z_\theta(\textbf{x})$ alongside deformed vertex locations, such that we output $(f(\textbf{x}), z(\textbf{x}))$. This is in direct analogy to the use of learned feature vectors in implicit SDF methods \citep{NeuS, yariv2020multiview}. As argued by \citet{yariv2020multiview}, this feature vector could learn to encode global effects such as shadows and secondary light reflections during training. We define the differentiable rasterizer,

\begin{equation}
    r(V_S,z(V_D), C_i)=(x_i, z_i, n_i, \tilde{m}_i),
\end{equation}

to generate a 2D image with per-pixel linearly-interpolated surface positions $x_i$, normals $n_i$, predicted masks $\tilde{m}_i$, and features $z_i$. Finally, this is fed to the \emph{neural shader} $h_\sigma(x_i,n_i, z_i, C_i)$---a small learned MLP with parameters $\sigma$---that predicts per-pixel RGB values to render a final predicted image $\tilde I_i$. The use of a learned feature $z$ facilitates fast and accurate colour learning. This is likely in-part due to its increased representation ability, but also because the deformation network benefits from the hybrid positional encoding $\gamma_H(\textbf{x})$, allowing it learn high-frequency feature vectors $z$ from intrinsic information. However, as discussed in a concurrent work by \citet{wu2022voxurf}, the use of feature vectors $z(\textbf{x})$ to aid rendering can disturb normal-color dependency, overshadowing the learning of the geometry. For this reason we compute photometric losses with two shaders, a feature-based shader $h_z$ and a geometry-based shader $h_g$ which doesn't receive $z$. We refer the reader to \Cref{geo_based_shader} for implementation details.
 
\subsection{Loss function}
\label{sec:losses}
As our objective function we use a combination of a photometric L1 loss $L_c$, a mask-based loss $L_m$, and a normal regularization loss $L_n$ computed on the mesh $\mathcal{M}_S$, 
\begin{equation}
    \underset{\theta, \sigma}{\min} \, \lambda_c L_c(\mathcal{I}; \theta, \sigma) + \lambda_m L_m({M_S; \theta, \sigma}) + \lambda_n L_n(S; \theta),
\end{equation}
with hyperparameters $\lambda_c, \lambda_m, \lambda_n \in \mathbb{R}$.
\paragraph{Mesh Regularization} The low-frequency bias of neural networks has been successfully used as a geometric smoothness prior \citep{deepgeomprior}. However, the positional encoding, promoting higher-frequencies, can counter these advantages and result in noisy surfaces. For this reason, we use a normal-smoothness loss on $M_S=(V_S,E,F)$ to induce an underlying smooth $f$. Following \citet{worchel2022multi}, we place a soft constraint on the consistency of face normals $\textbf{n} \in S^2$ using a cosine similarity, 
\begin{equation}
     L_n = \frac{\lambda_n}{|E|} \sum_{(i,j) \in E} {(1 - \textbf{n}_i \cdot \textbf{n}_j)^2},
\end{equation}
where $(i,j)$ are the indices of the left and right faces of each edge in $E$. Note that, unlike the fixed-mesh method NDS, we do not use a Laplacian loss on the mesh (to promote triangle regularity), since eventually, $M_S$ is just a proxy to learning $f$, where we, in fact, want to allow for triangles to deform sufficiently to fit narrow regions of the target geometry.

\paragraph{Appearance Loss} In our deferred shading pipeline, predicted masks $\tilde{m}_i$ are produced by the rasterizer $r$ before any shading. Our neural shader further produces two image predictions, $\tilde{I}^z_i$ and $\tilde{I}_i$ from the shader modules $h_{z}$ and $h_{g}$ respectively. With these quantities, we compute a combined photometric loss $L_c$ and a mask loss as follows:
\begin{equation}
    L_c =  \frac{1}{|\mathcal{I}|}\sum^{|\mathcal{I}|}_{i=1} \|I_i -\tilde{I}_i^{z}\| + \lambda_g \|I_i -\tilde{I}_i\|, \quad L_m = \frac{1}{|\mathcal{I}|}\sum^{|\mathcal{I}|}_{i=1} \|m_i -\tilde{m}_i\|,
\label{eq:loss}
\end{equation}
where $\lambda_g$ controls the contribution of $h_g$ and consequently its impact on the learned geometry. Details about $L_n$ are in the supplementary material.

\section{Experiments}\label{sec:exp}
For shading, we use the code provided by~\citet{worchel2022multi} in PyTorch~\citep{pytorch}.
For differentiable rasterization, we use the high-performance modular primitives provided by~\citet{Laine2020diffrast}.

\subsection{Surface Reconstruction}
\label{sec:surface_rec}

\begin{figure}[t]
    \centering
    \setlength{\tabcolsep}{0pt}
    \resizebox{\textwidth}{!}{
    \begin{tabular}{ccccccc}
        & \multicolumn{2}{c}{24 Red House}& \multicolumn{2}{c}{122 Owl}& \multicolumn{2}{c}{114 Buddha}\\
     \rotatebox{90}{\parbox[t]{2cm}{\hspace*{\fill}NDS\hspace*{\fill}}}\hspace*{5pt}
                            &  \includegraphics[width=2cm]{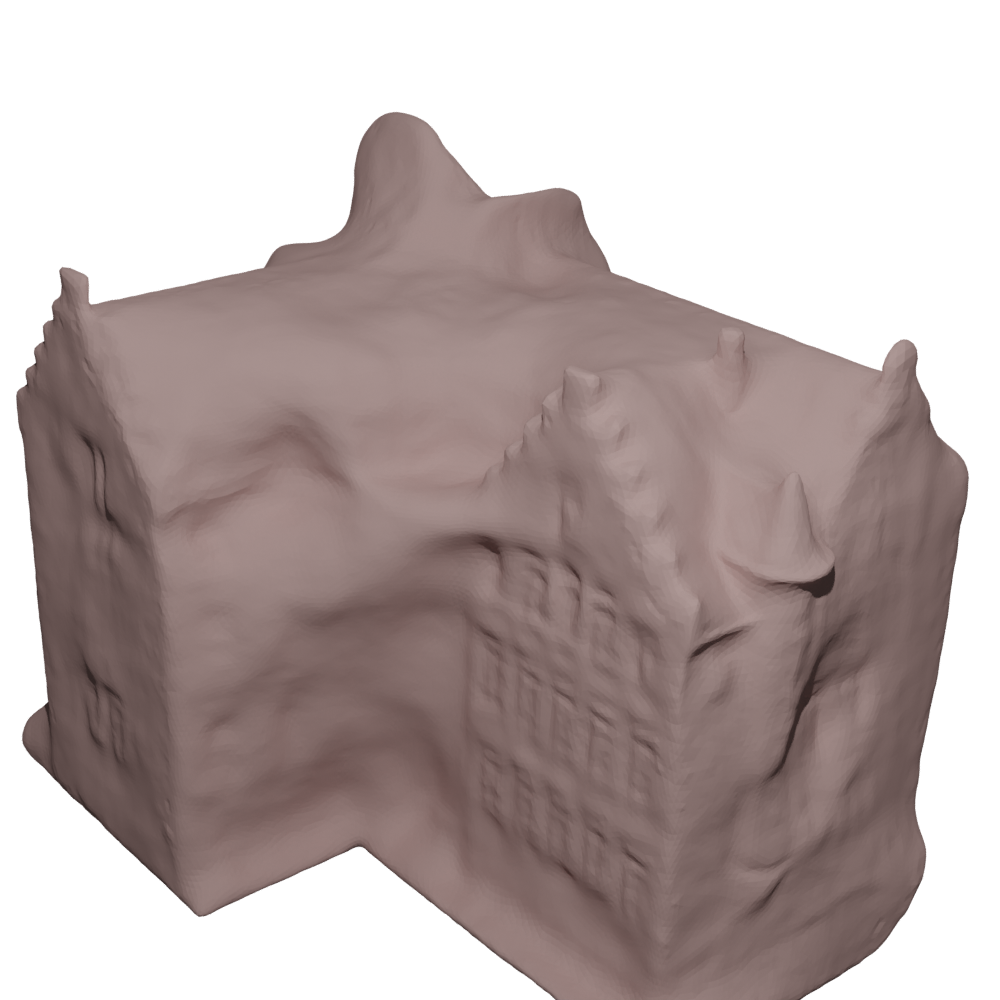}
                            &  \includegraphics[width=2cm]{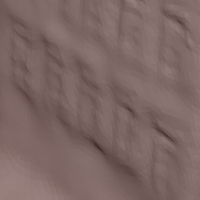}
                            &  \includegraphics[width=2cm]{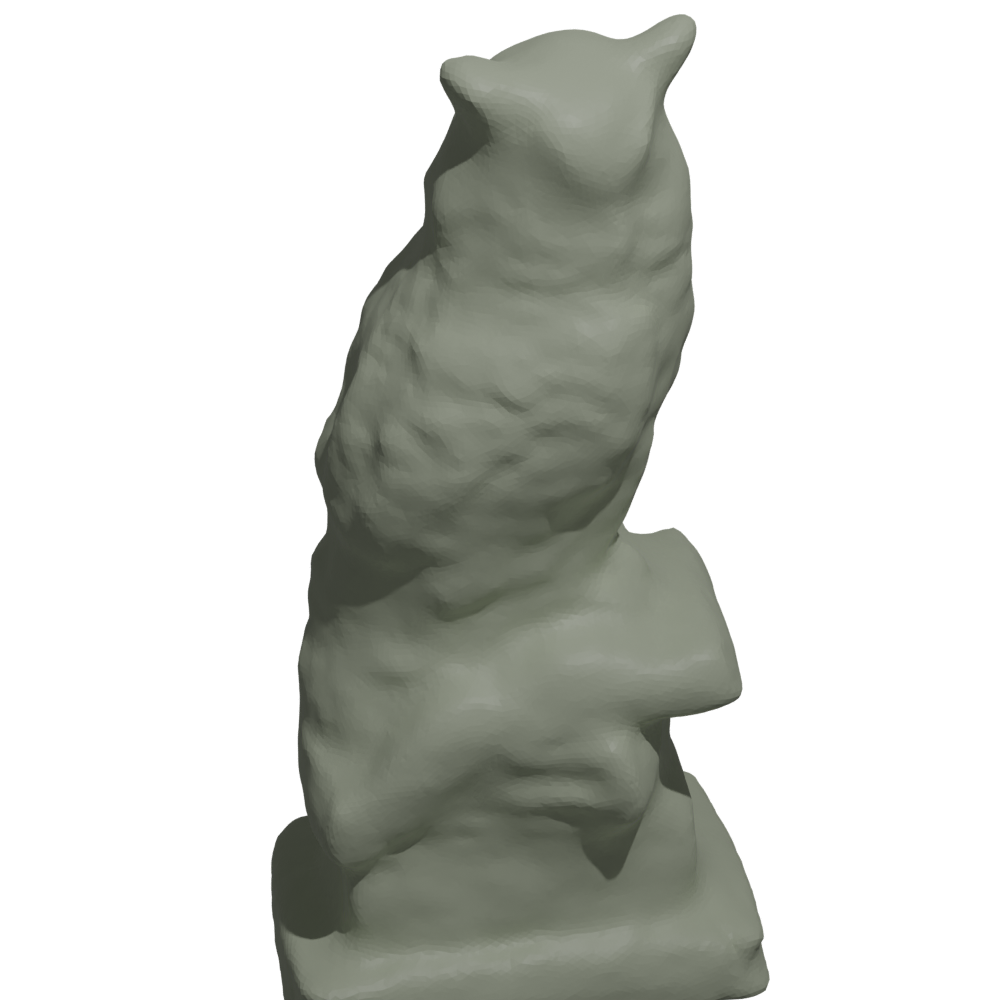}
                            &  \includegraphics[width=2cm]{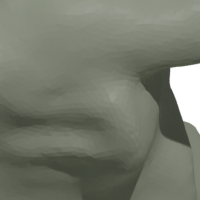}
                            &  \includegraphics[width=2cm]{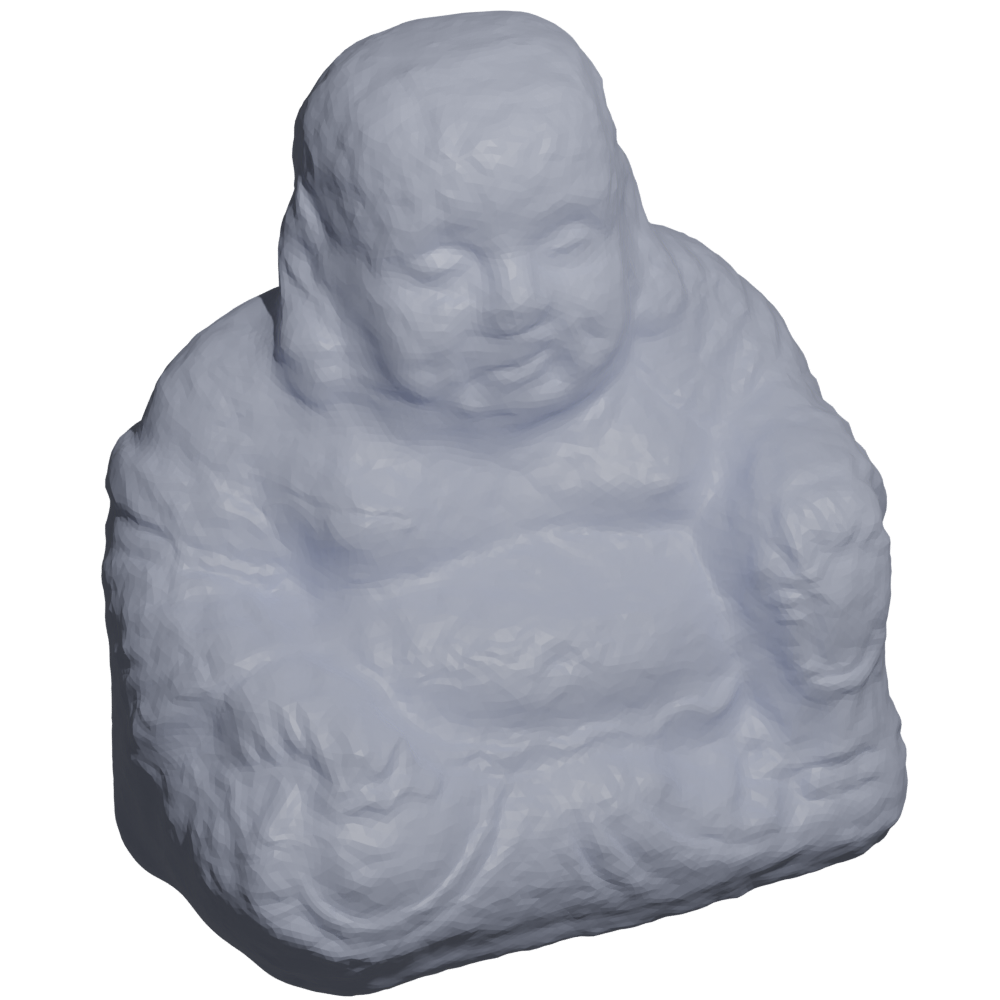}
                            &  \includegraphics[width=2cm]{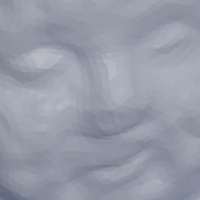}\\
     \rotatebox{90}{\parbox[t]{2cm}{\hspace*{\fill}NeuS\hspace*{\fill}}}\hspace*{5pt}
                            &  \includegraphics[width=2cm]{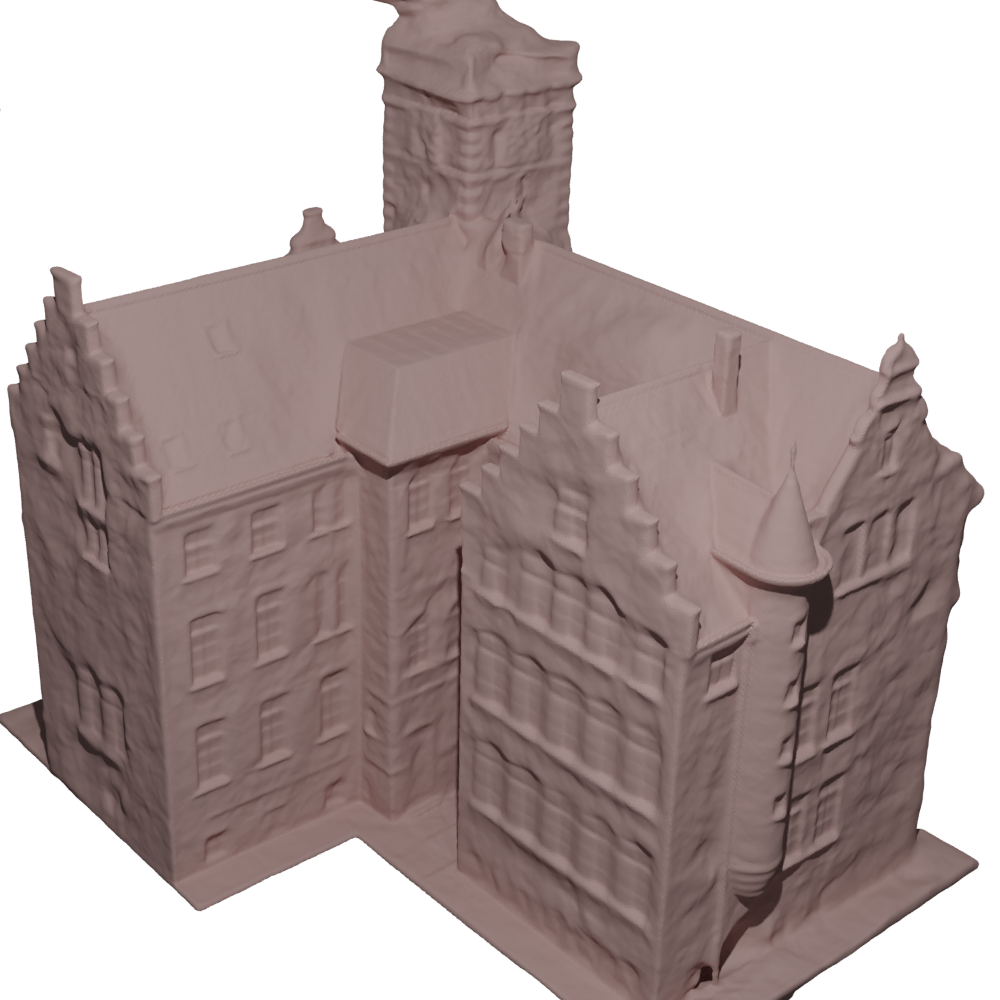}
                            &  \includegraphics[width=2cm]{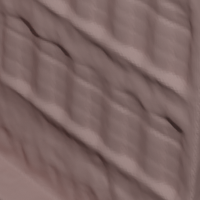}
                            &  \includegraphics[width=2cm]{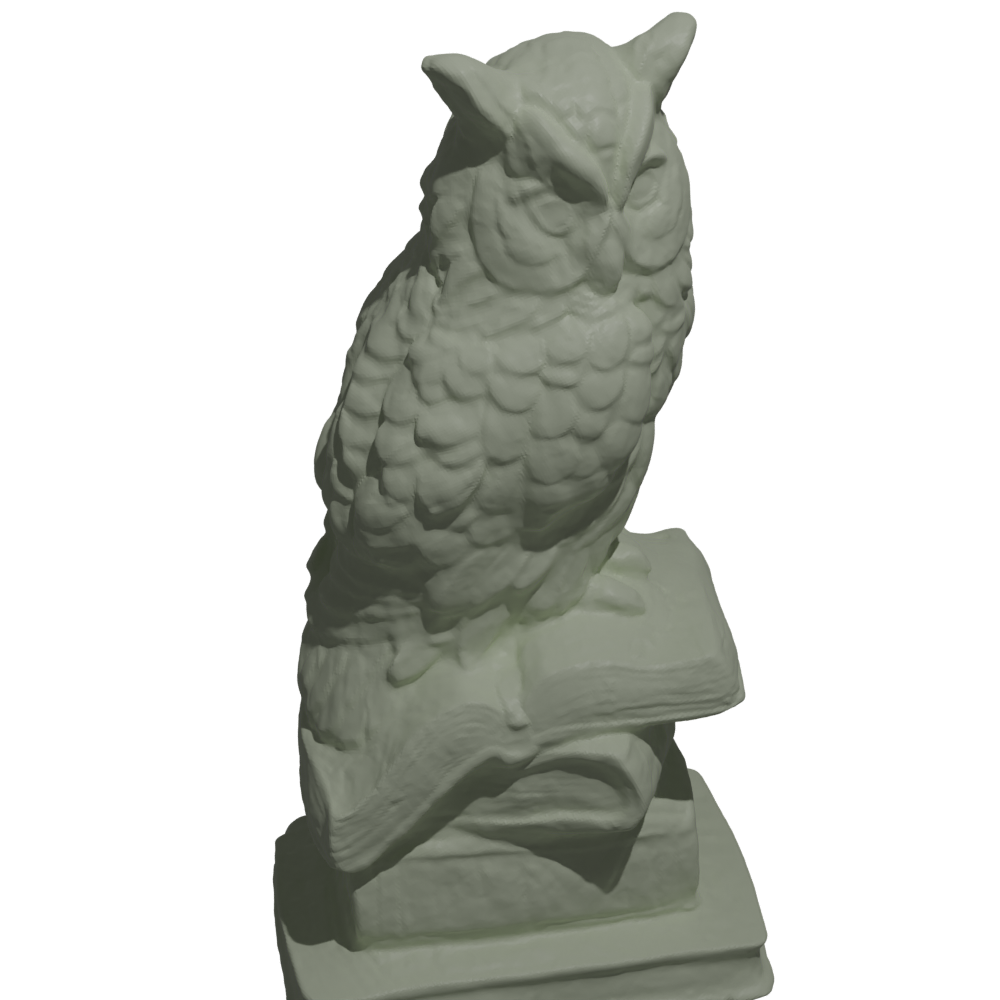}
                            &  \includegraphics[width=2cm]{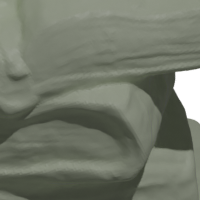}
                            &  \includegraphics[width=2cm]{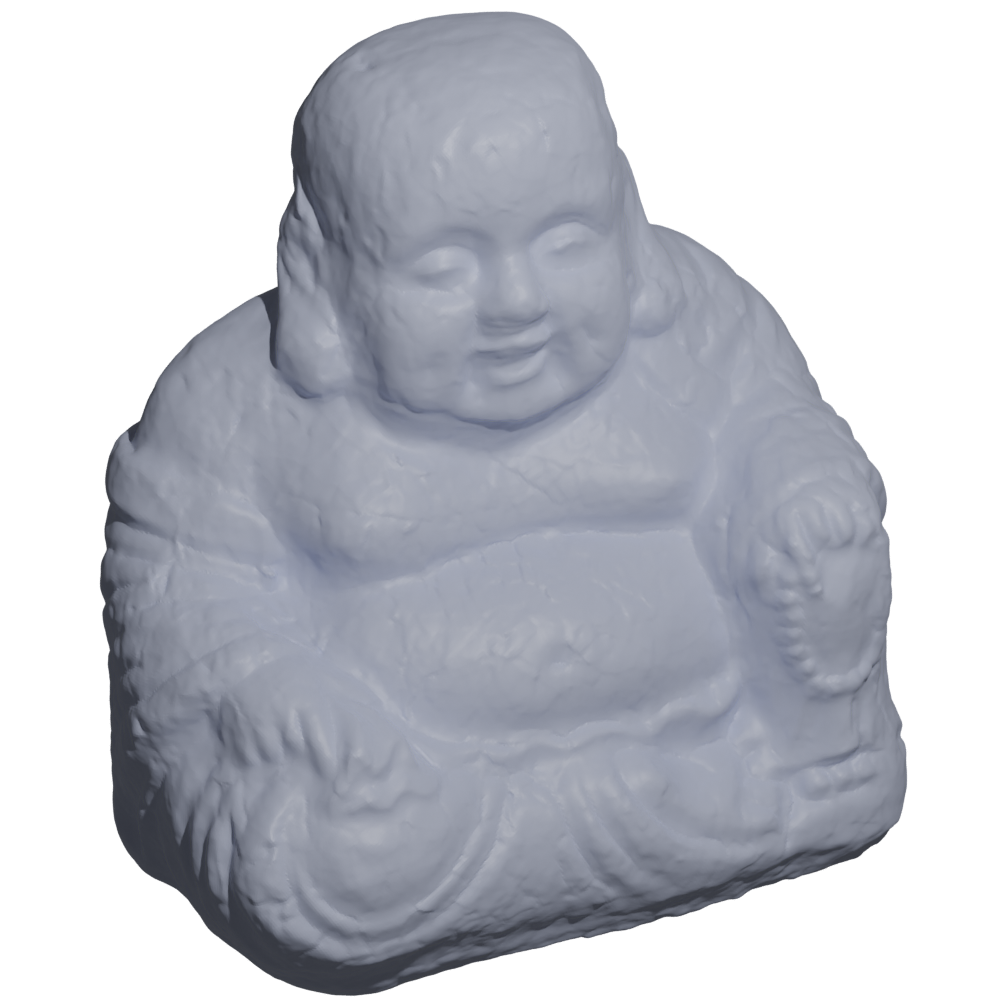}
                            &  \includegraphics[width=2cm]{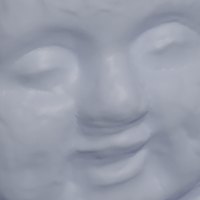}\\
      
     \rotatebox{90}{\parbox[t]{2cm}{\hspace*{\fill}ENS\hspace*{\fill}}}\hspace*{5pt}
                            &  \includegraphics[width=2cm]{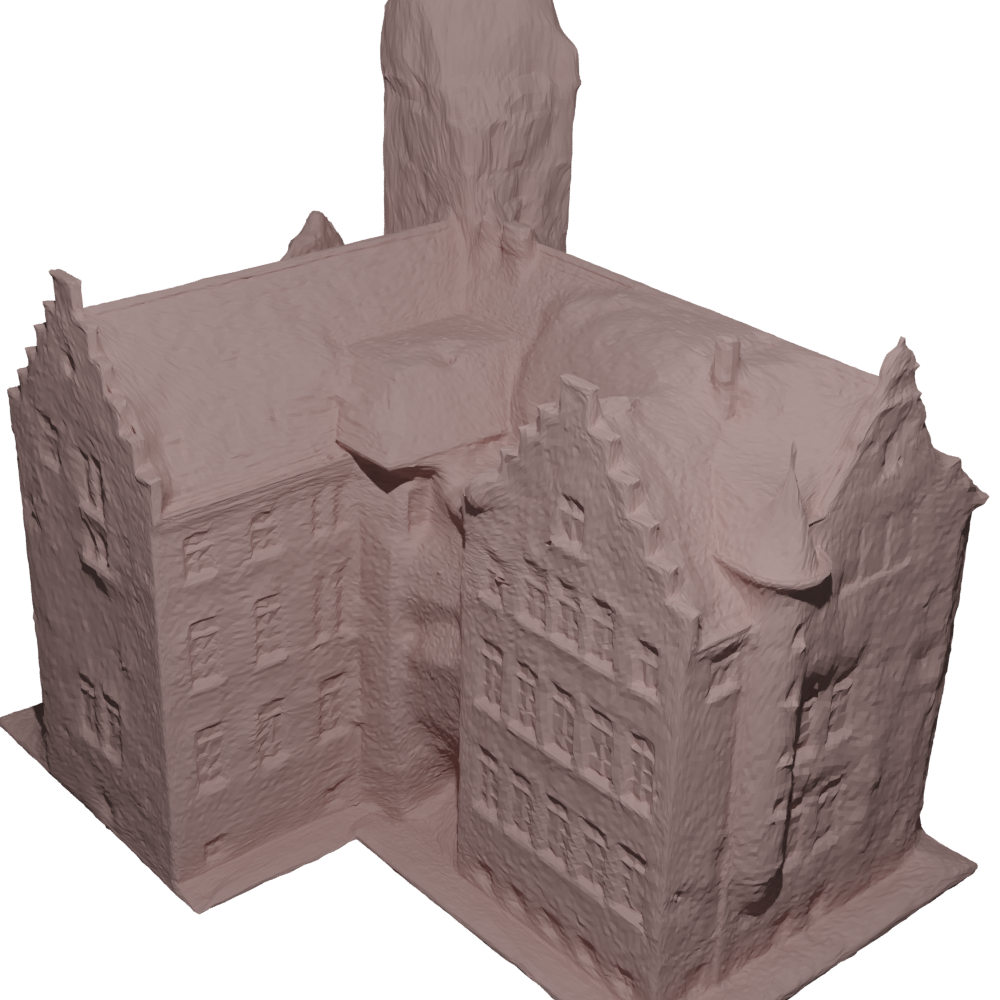}
                            &  \includegraphics[width=2cm]{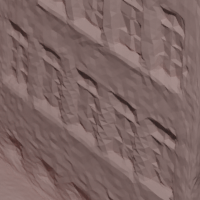}
                            &  \includegraphics[width=2cm]{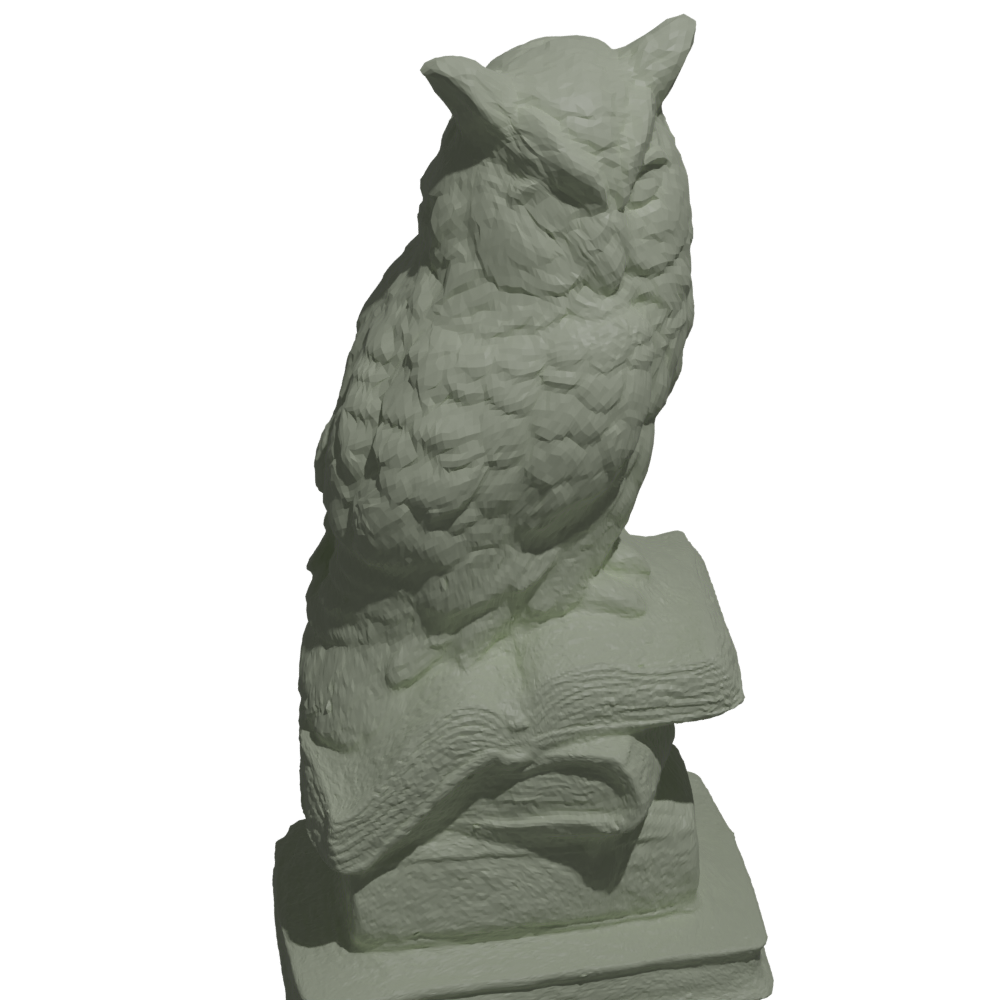}
                            &  \includegraphics[width=2cm]{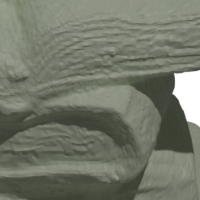}
                            &  \includegraphics[width=2cm]{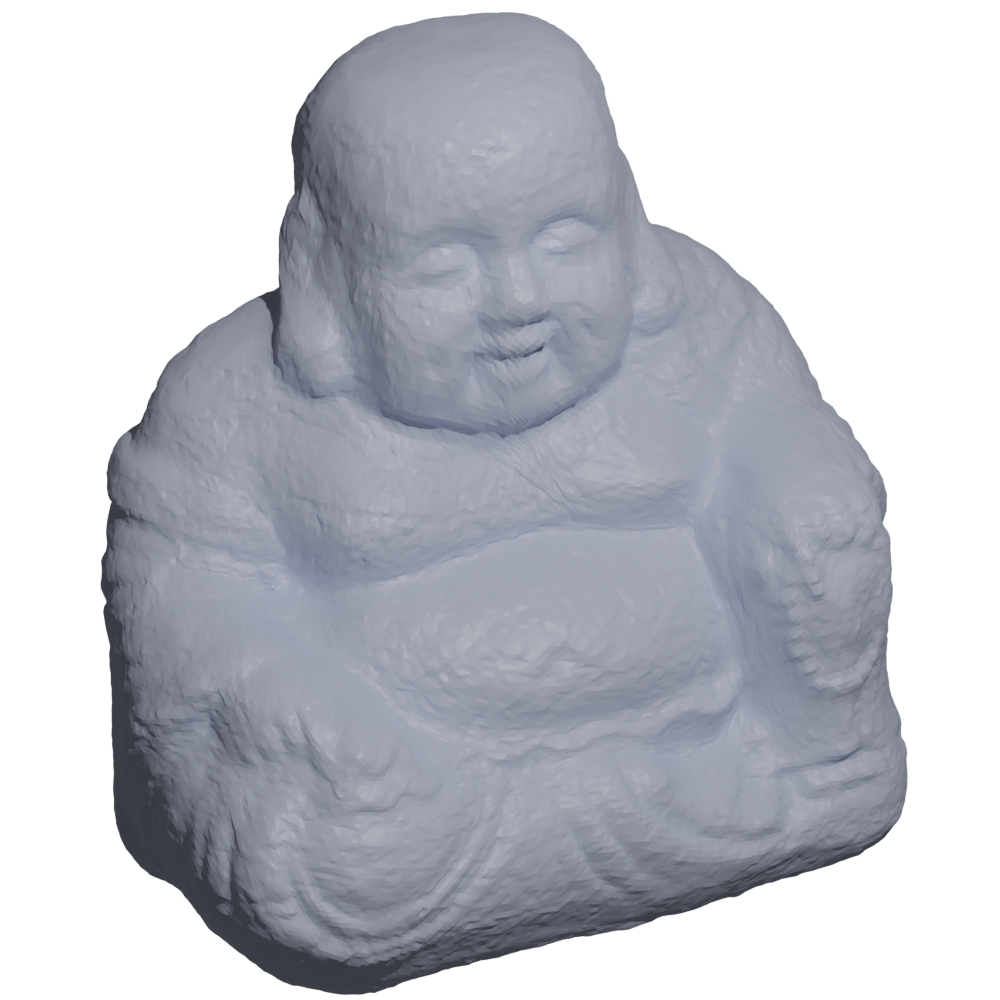}
                            &  \includegraphics[width=2cm]{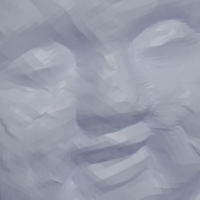}\\ %
    \end{tabular}
    }
    \caption{Qualitative evaluation for multi-view surface reconstruction. We compare our method ENS to the state-of-the-art implicit and explicit methods in three diverse scenes of the DTU dataset.}
    \label{fig:dtu_recons}
\end{figure}

\begin{table}[t]
\hspace{-15pt}
\centering
\scriptsize
\resizebox{1\linewidth}{!}{
\setlength{\tabcolsep}{3pt}
    \renewcommand{\arraystretch}{1.2}
    \renewcommand{\b}{\bfseries}
    \renewcommand{\i}{\it}
    \begin{tabular}{lrrrrrrrrrrrrrrrr|ccc}
     Scan   &     24 &     37 &     40 &     55 &     63 &     65 &     69 &     83 &     97 &    105 &    106 &    110 &    114 &    118 &    122 &     Avg & TT & MT & RT  \\
     \hline
    
    IDR   &   1.58 &   2.06 &   0.75 &   0.43 &\b 1.06 &   0.68 &   0.68 &\b 1.38 &   1.17 &   0.88 &   0.63 &\b 0.99 &   0.37 &   0.50 &   0.52 &    0.92 &   5hrs  & 5s     & 30s*\\
    NeuS  &\b 0.83 &\b 0.98 &\b 0.56 &\b 0.37 &   1.13 &\b 0.59 &\b 0.60 &   1.45 &\b 0.95 &\b 0.78 &\b 0.52 &   1.43 &\b 0.36 &\b 0.45 &\b 0.45 & \b 0.77 &   5hrs  & 5s     & 30s*\\
    \hline
    NDS   &   4.24 &   5.25 &   1.30 &   0.53 &   1.72 &   1.22 &   1.35 &   1.59 &   2.77 &   1.15 &   1.02 &   3.18 &   0.62 &   1.65 &   0.91 &    1.95 &\b 3min  & \b N/A & \b 5ms \\
    ENS   &   1.66 &   3.30 &\i 0.88 &   0.40 &   1.59 &   1.04 &   0.84 &   1.50 &   1.22 &   0.91 &\i 0.75 &   2.67 &\b 0.36 &   0.59 &   0.53 &    1.22 &\b 5min  & \b 2ms & \b 5ms \\
    \end{tabular}
}
\caption{Chamfer scores (in millimeters) and average single scene times for multi-view surface reconstruction on the DTU dataset. TT: training time, MT: mesh extraction time, RT: rendering time. Results in \textit{italics} use a different original topology (\Cref{nonsphere}). Results followed by an $\ast$ are estimations based on the author's comments.}
\label{tab:dtu_chamfer}
\end{table}

We evaluate our approach on the task of surface reconstruction from multi-view raster images.
We compare our method to the explicit NDS~\citep{worchel2022multi}, and two implicit surface approaches, IDR~\citep{yariv2020multiview}, and NeuS~\citep{NeuS}. 
While both are SDF-based, IDR is more closely related to ours in that it uses surface rather than volume rendering.
\paragraph{Dataset}
We use 15 scenes from the DTU dataset~\citep{aanaes2016large} for comparison. Each scene contains 1600 x1200 resolution images, each paired with camera parameters and binary masks, from 49 or 64 poses. 
Collectively, the scenes are challenging for reconstruction algorithms as they contain non-Lambertian materials, high-frequency geometry, inconsistent lighting, and geometry/texture ambiguities. 
We follow the official DTU evaluation in surface mode to generate Chamfer-L1 scores in comparison to ground truth point clouds (\Cref{tab:dtu_chamfer}).

ENS significantly outperforms NDS, the benchmark explicit approach, across all scenes, while maintaining the extremely efficient training, rendering and mesh extraction. On sphere topology, we approach state-of-the-art reconstruction quality in a fraction of the time, and attain a preferable \textit{explicit} neural representation from which meshes can be directly extracted and rendered in real-time. In \Cref{fig:dtu_recons}, we present reconstructions from our approach compared to NDS and NeuS. Compared to NeuS, our coarse-to-fine hybrid encoding enables us to capture high-frequency details that are not present in the compared methods. For scan 24, one of the more challenging tasks, where NDS fails, our method successfully recovers a surface and successfully learns sharp, high-frequency features. 

\newcommand\width{\textwidth}

\begin{figure}
    \centering
         \includegraphics[width=0.28\linewidth]{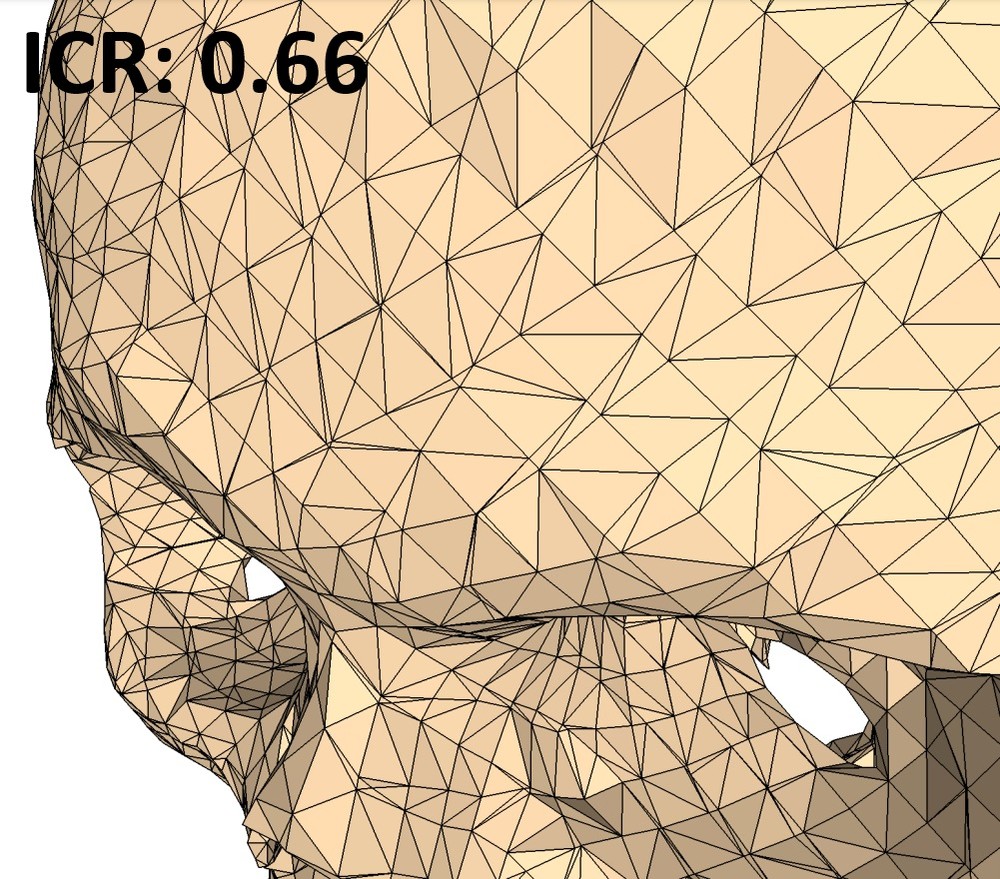}  \includegraphics[width=0.28\linewidth]{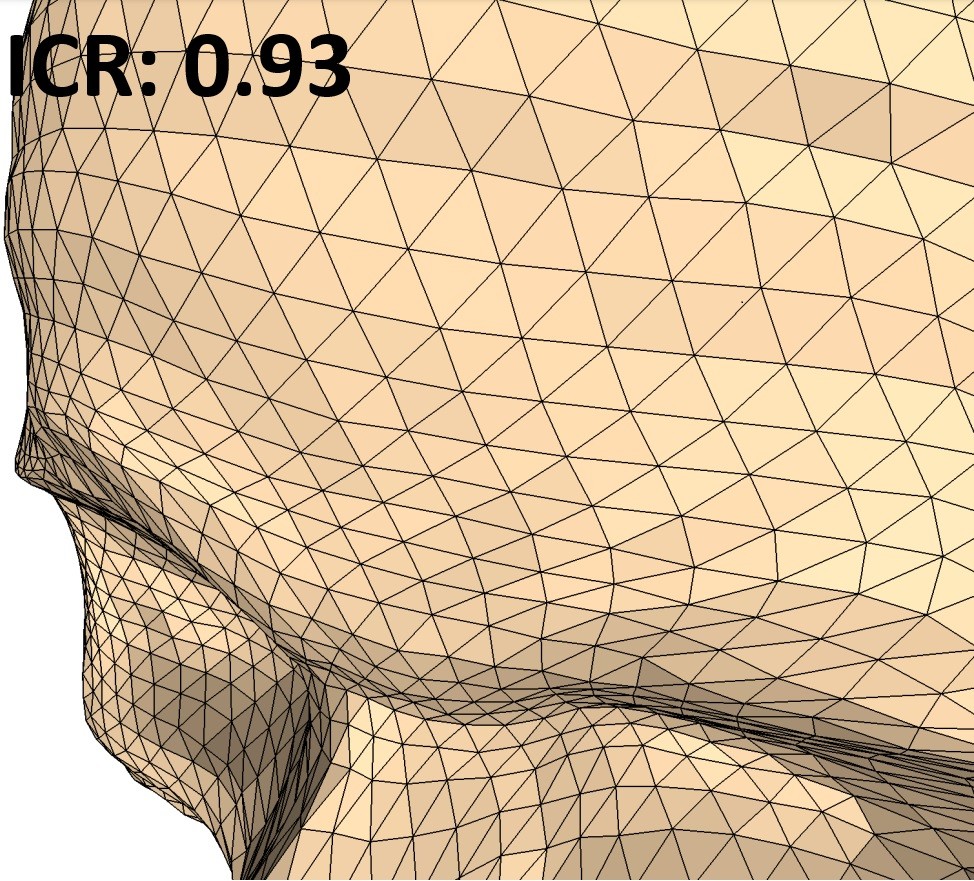}
         \includegraphics[width=0.28\linewidth]{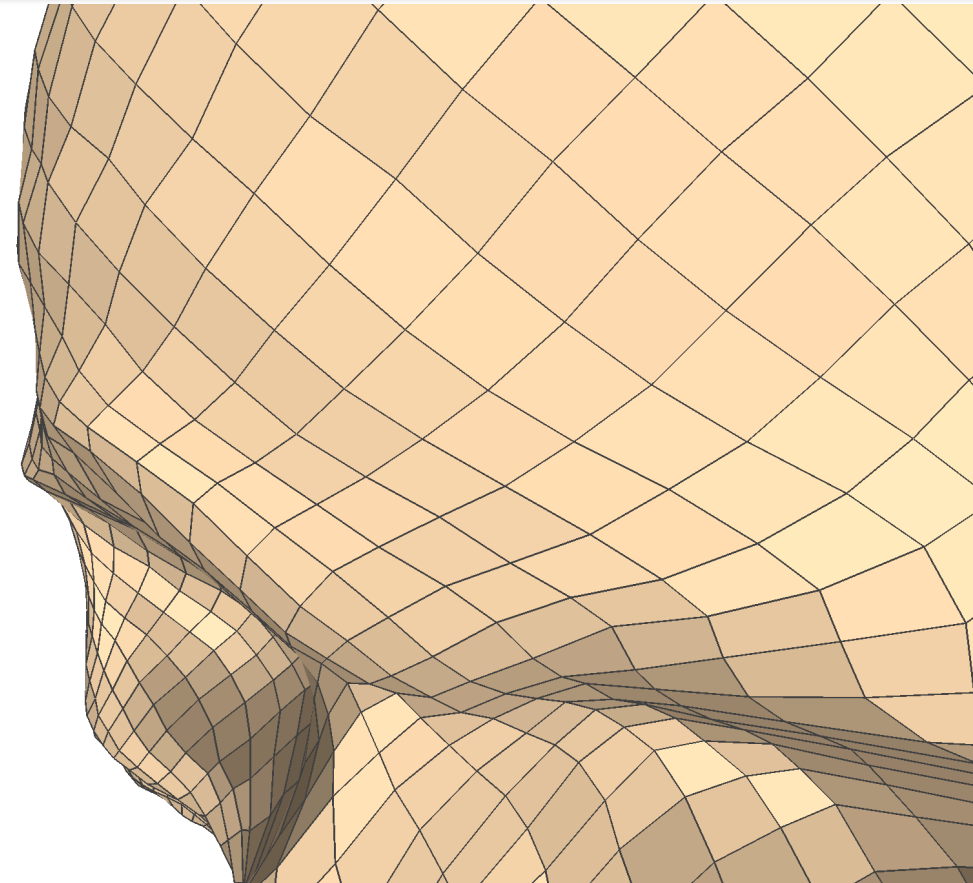}    
    \caption{Implicit (left, from NeuS), vs. explicit (ours) extracted meshes of comparable resolutions. The marching cubes artifacts are evident visually in unwanted holes and aliasing, as well as in the quality measures (ICR, where 1 is ideal). Our method can generate any mesh by pullback from the domain, here exemplified for both triangle (middle) and quad (right), and achieve higher quality elements.}
    \label{fig:connectivity}
\end{figure}

\paragraph{Mesh Quality}
In addition to efficiency, a major advantage of ENS is that we can attain higher-quality meshes for downstream tasks than implicit approaches, as the surface is modelled explicitly. 
In \Cref{fig:connectivity}, we compare meshes of $\approx 10$K vertices extracted from NeuS and ENS, where we generate both triangle and quadrilateral meshes from $f$ and $D$ with just inference, as they are decoupled from any fixed mesh $M$. It is evident that the marching-cubes mesh exhibits grid artifacts \citep{sawdayee2022orex}. See ~\Cref{tab:mesh_qual_tab} for quality analysis. 
While mesh-only methods like NDS require strong regularization to converge to a desirable surface \citep{Nicolet2021Large}, we can inherit the regularity of our neural network as a prior to attain good solutions \citep{deepgeomprior}. However, unlike implicit approaches, ENS can efficiently compute mesh-based losses and back-propagate them to the continuous representation. For example, in \Cref{tab:mesh_qual_tab} we present mesh quality statistics on a subset of DTU where we directly optimize the average inradius-to-circumradius \citep{shewchuk2002good} of the output mesh during training, and hence attain higher quality meshing for downstream tasks. For further mesh quality analysis as well as implementation details please see \Cref{remesh}.

\begin{table}[t]
    \centering
    \setlength{\tabcolsep}{2pt}
    \begin{tabular}[t]{cr|rrr|rrr|rrr}
    && \multicolumn{3}{c|}{Scene 65} & \multicolumn{3}{c|}{Scene 118} & \multicolumn{3}{c}{Scene 114} \\
    && NeuS & ENS & ENS* & NeuS & ENS & ENS* & NeuS & ENS & ENS* \\
    \midrule
    \rotatebox{90}{\footnotesize\hspace{-25pt}ICR}
    & Average ($\uparrow$)  & 0.66 & 0.84 & \textbf{0.95} & 0.65 & 0.72 & \textbf{0.92} & 0.65 & 0.79 &\textbf{0.94}\\
    & \% $<$ 0.10 ($\downarrow$)& 5.80 & 0.27 & \textbf{0.05} & 6.03 & 2.28 & \textbf{0.15} & 5.95 & 0.95& \textbf{0.11}\\
    & \% $<$ 0.25 ($\downarrow$)& 13.74 & 1.15 & \textbf{0.16} & 14.40 & 7.79 &\textbf{0.89} & 14.41 & 5.10 &\textbf{0.66} \\
    & \% $<$ 0.90 ($\downarrow$)& 83.20 & 49.39 &\textbf{14.49} & 82.99 & 65.29 &\textbf{24.84} & 83.01 & 53.88 &\textbf{18.32}\\
    \midrule
    \rotatebox{90}{\footnotesize\hspace{-4pt}CD}
    & Average ($\downarrow$) & \textbf{0.59} & 1.04 & 1.06 & \textbf{0.45} & 0.59 & 0.63 & \textbf{0.36} & \textbf{0.36} & \textbf{0.36}\\
    \end{tabular}
    \caption{Normalized Inradius-to-circumradius  ratio (ICR) quality and Chamfer Distance (CD) statistics for NeuS vs ENS (ours). For normalized ICR, 1 is ideal and 0 is degenerate. We include both ENS and a regularized ENS*, for which we optimize ICR. With this model we can extract very high quality meshes in real-time, with minimal impact on Chamfer distance.}
    \label{tab:mesh_qual_tab}
\end{table}

\begin{figure}[t]
    \centering
    \begin{tabular}{cccc}
        \includegraphics[width=0.23\textwidth]{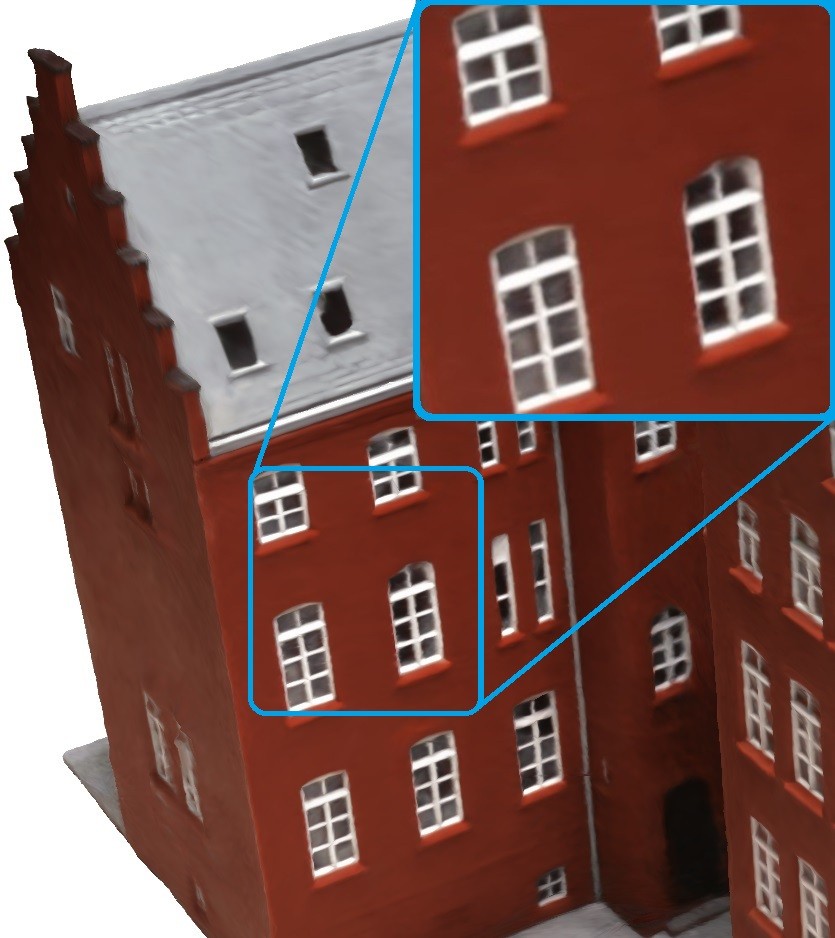} &
        \includegraphics[width=0.23\textwidth]{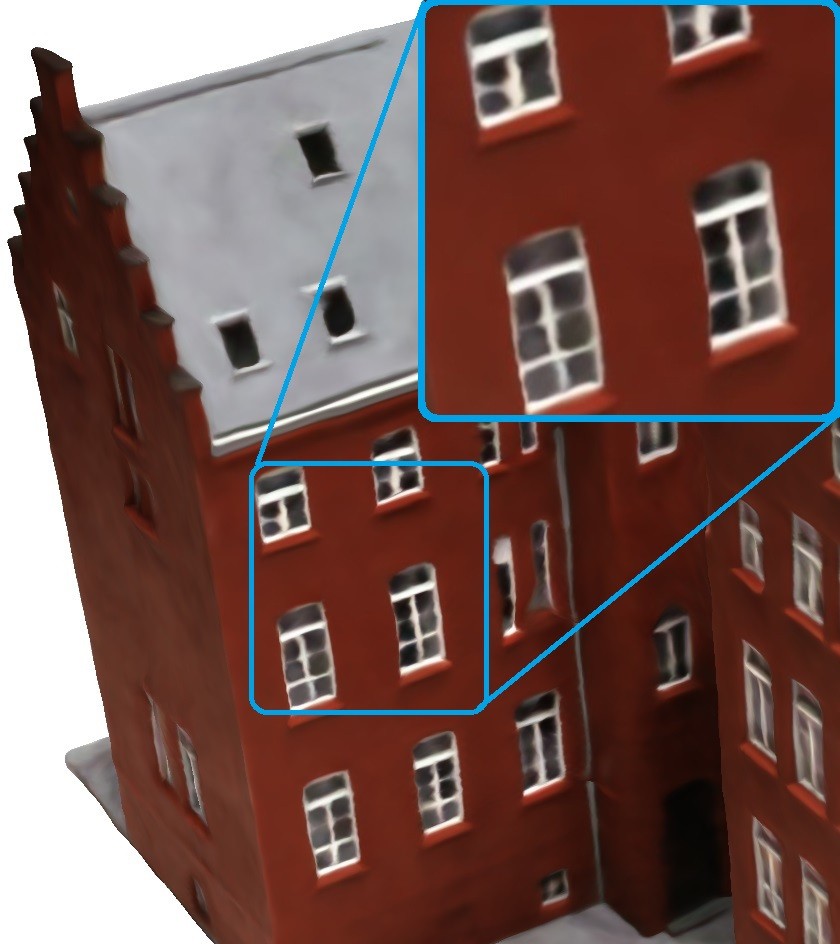} &
        \includegraphics[width=0.23\textwidth]{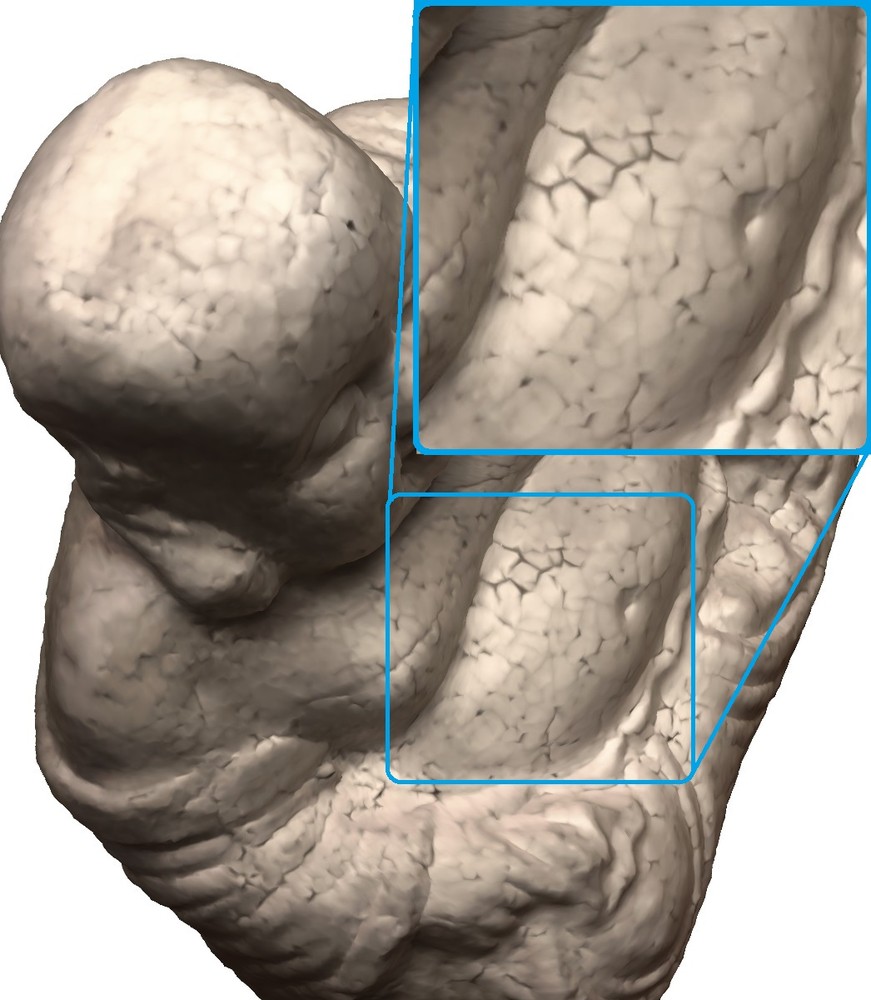} &
        \includegraphics[width=0.23\textwidth]{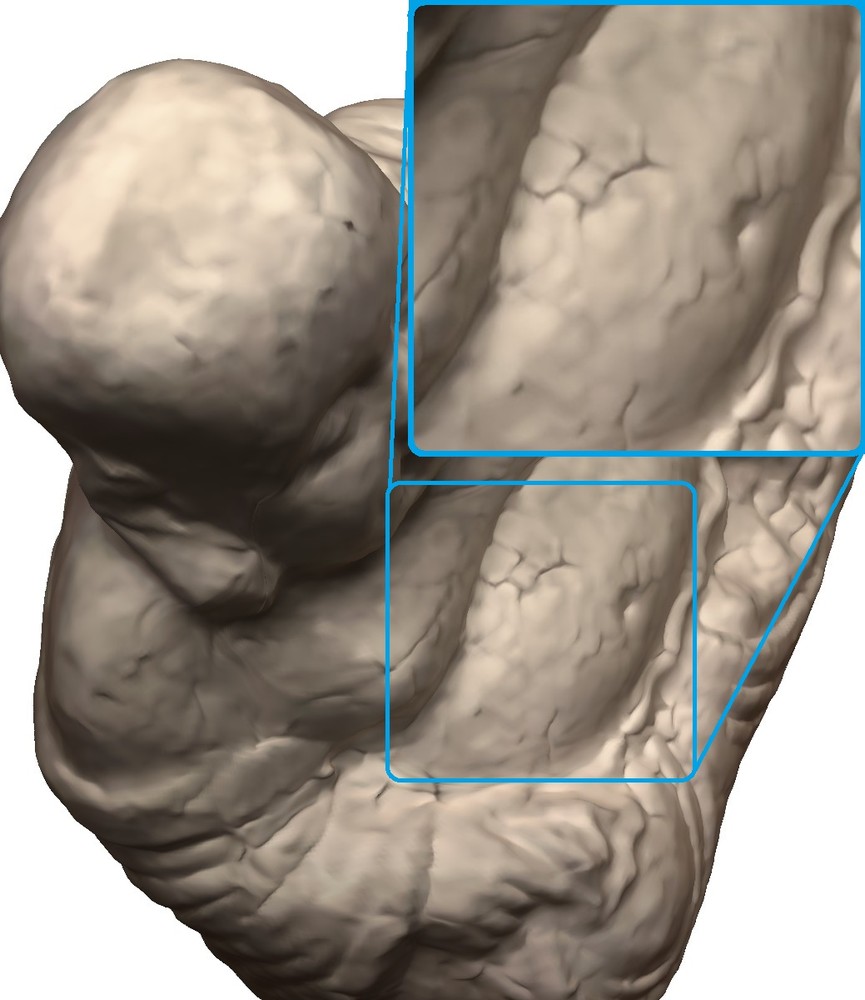} \\
        
        (a) with $\gamma_I$ & (b) without $\gamma_I$ & (c) with $\gamma_I$ & (d) without $\gamma_I$
    \end{tabular}
    \caption{Rendering ENS with and without intrinsic positional encoding $\gamma_I$. We clearly observe improved rendering details in the models which use intrinsic positional encoding.}
    \label{fig:intrinsic}
\end{figure}

\begin{figure}[t]
    \hspace{-10pt}
    \centering
    \setlength{\tabcolsep}{0pt}
    \begin{tabular}{cccc}
        \begin{minipage}{0.22\textwidth}
            \vspace{-10pt}
            \centering
            \setlength{\tabcolsep}{2pt}
            \resizebox{\textwidth}{!}{
                \begin{tabular}{lc}
                    Ablation & CD ($\downarrow$)\\
                    \midrule
                    w/o $\gamma_E$   & 3.39 \\
                    w/o $\gamma_I$   & 1.31 \\
                    w/o $f_{coarse}$ & DNC \\
                    w/o $h_g$        & 1.28 \\
                    Full model       & \textbf{1.22}\\
                \end{tabular}
            }
            \label{tab:ablated_chamfer}
        \end{minipage}
        &
        \hspace{1pt}

        \begin{minipage}{0.22\textwidth}
            \centering
            \includegraphics[width=\textwidth]{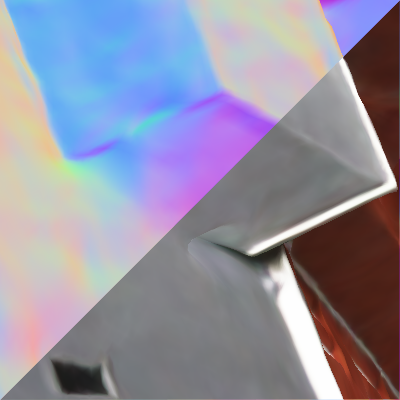}
            \label{fig:ablated_gamma_I}
        \end{minipage}
        &
        \hspace{1pt}

        \begin{minipage}{0.22\textwidth}
            \centering
            \includegraphics[width=\textwidth]{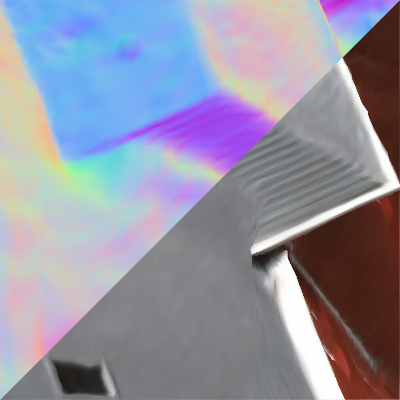}
            \label{fig:ablated_h_g}
        \end{minipage}
        &
        \hspace{1pt}
        \begin{minipage}{0.22\textwidth}
            \centering
            \includegraphics[width=\textwidth]{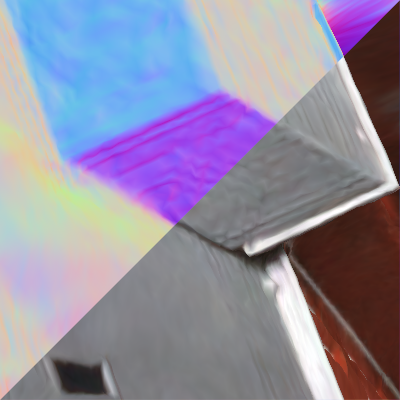}
            \label{fig:ablated_full_mod}
        \end{minipage} \\
       (a) CD & (b) w/o $\gamma_I$ & (c) w/o $h_g$ & (d) ENS

    \end{tabular}
    \caption{Quantitative and qualitative ablation study. (a): Average Chamfer distance (CD) on the DTU dataset for different ablated models. DNC: did not converge. (b-d): Comparison of renders and normal maps on the house scene for converging ablated models.}
    \label{fig:ablations}
\end{figure}

\paragraph{Ablation} In \Cref{fig:ablations} we ablate various model components by individually removing them from the full model. Without intrinsic encoding (``w/o $\gamma_I$''), the results in missing fine details such as the tiles on the roof of scan 24. In \Cref{fig:intrinsic} we further demonstrate that intrinsic encoding produces notably improved rendering quality, capturing fine structures such as window frames. Using only intrinsic encoding, (``w/o $\gamma_E$'') struggles to capture coarse shape in the initial stages of training and results in significantly depreciated surface accuracy. See \Cref{extended_ablation} for further examples. 
Removing the geometry-based shader (``w/o $h_g$'') results in a model which can capture high-frequencies. However, areas with strong colour-normal dependency, such as the roofs of the house, are prone to extraneous growths with painted on textures (see \Cref{fig:ablated_h_g}).  See \Cref{diverge} for further details.

\section{Limitations and Future work}\label{sec:conc}
The dominant limitation of our method is the same as surrounding explicit methods, in that we have to predetermine the topology of the base domain $D$ to fit the target object in question. However, the topology is only fixed at the level of choosing $D$, where we can use an explicit representation of any geometry that is assumingly topologically-equivalent to the solution. An exciting future direction could be integrating explicit neural surfaces into FEA-based shape optimization pipelines, for which the high mesh-quality produces of our representation is particularly attractive for accurately solving PDEs.

\section{Conclusion}
We proposed an explicit neural surface (ENS) for multi-view 3D reconstruction that combines the advantages of both explicit mesh and neural representations. Our approach produces high-quality meshing, and is extremely fast to train, render and mesh while being competitive with state-of-the-art implicit methods. This representation then has the potential to be used for downstream geometry-processing tasks such as physically-based optimization \citep{umetani2018learning}, computational geometric design \citep{neumesh} or exploring topologically consistent shape spaces \citep{kilian2007geometric}.

\bibliography{pmlr-sample.bib}

\newpage
\section{Appendix}

\appendix

We provide further implementation details in \Cref{sec:impdet}, extended ablation studies in \Cref{extended_ablation}, and quantitative/qualitative analysis of mesh quality in \Cref{remesh}.

\section{Implementation Details} \label{sec:impdet}
\paragraph{Network Details}
For both deformation field networks $f_\text{coarse}, f_\text{fine}$ we use a single MLP with one hidden layer of 400 softplus units, which outputs a  deformed vertex position $f(\textbf{x})$, and a 128-dimensional feature vector $z(\textbf{x})$. We use a residual connection on the vertex locations such that we have:
\begin{equation}
        f(\textbf{x}) = \textbf{x} + \delta \text{MLP}(\textbf{x}),
\end{equation}
where is $\delta$ is a scaling parameter. When $f_{fine}$ is introduced at iteration $500$, this parameter is scheduled to linearly increase from $0$ to $0.1$ over $100$ iterations (this was found to improve stability). For each neural shader $h_z, h_g$ we use a single MLP with 3 hidden layers of 256 ReLU units. 

In our experiments (see \Cref{tab:mesh_qual_tab}) we demonstrate that accessing a mesh during training allows us to learn a deformation field that optimizes for mesh quality alongside reconstruction accuracy. Specifically, we add an additional inradius-to-circumradius regularization term into our loss. The inradius-to-circumradius ratio is a commonly used mesh quality measure in computational geometry and finite element analysis \citep{shewchuk2002good}. The inradius (\(r\)) is defined as the radius of the unique incircle, while the circumradius (\(R\)) is the radius of the unique circumcircle. The normalized ratio ${2r}/{R}$ being closer to 1 indicates that the triangle is closer to an equilateral shape, which leads to more accurate numerical simulations and reduced interpolation errors. We compute a loss over the mesh by penalizing the deviation of the average triangle normalized inradius-to-circumradius from 1,
\begin{equation}
    L_{ICR} = \frac{\lambda_{ICR}}{|F|} \sum_{i \in F} {(1 - \frac{2r_i}{R_i}}),
\end{equation}
where $\lambda_{ICR}= 5\times10^{-3}$

\begin{figure}[b]
\resizebox{\textwidth}{!}{
\begin{tabular}{cccc}
    \includegraphics[width=.24\linewidth]{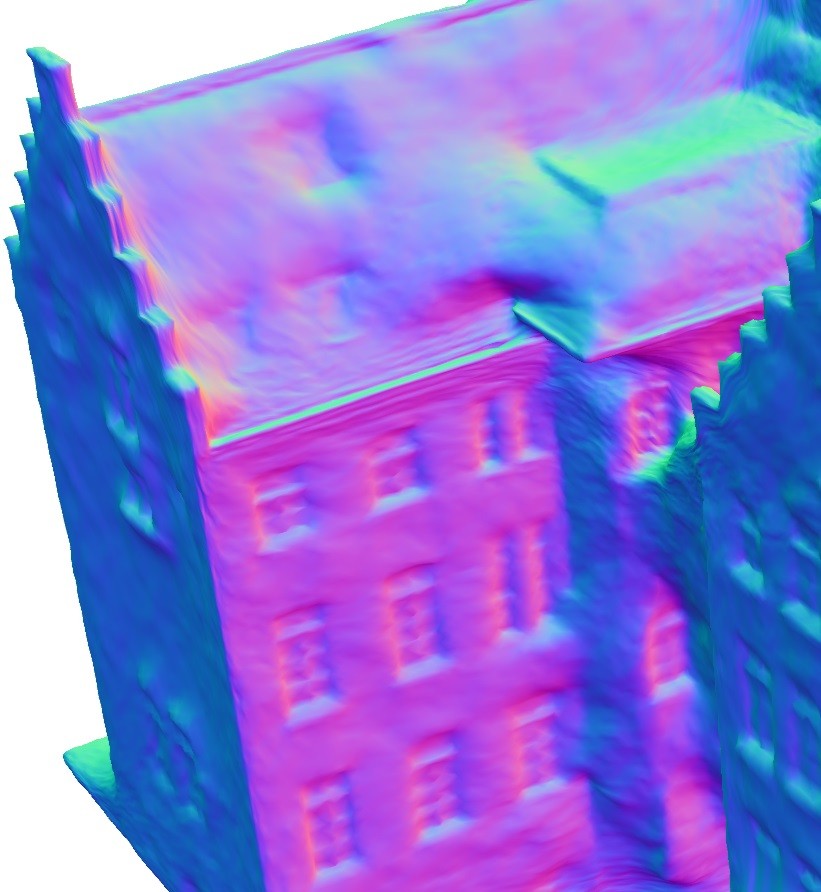} &
    \includegraphics[width=.24\linewidth]{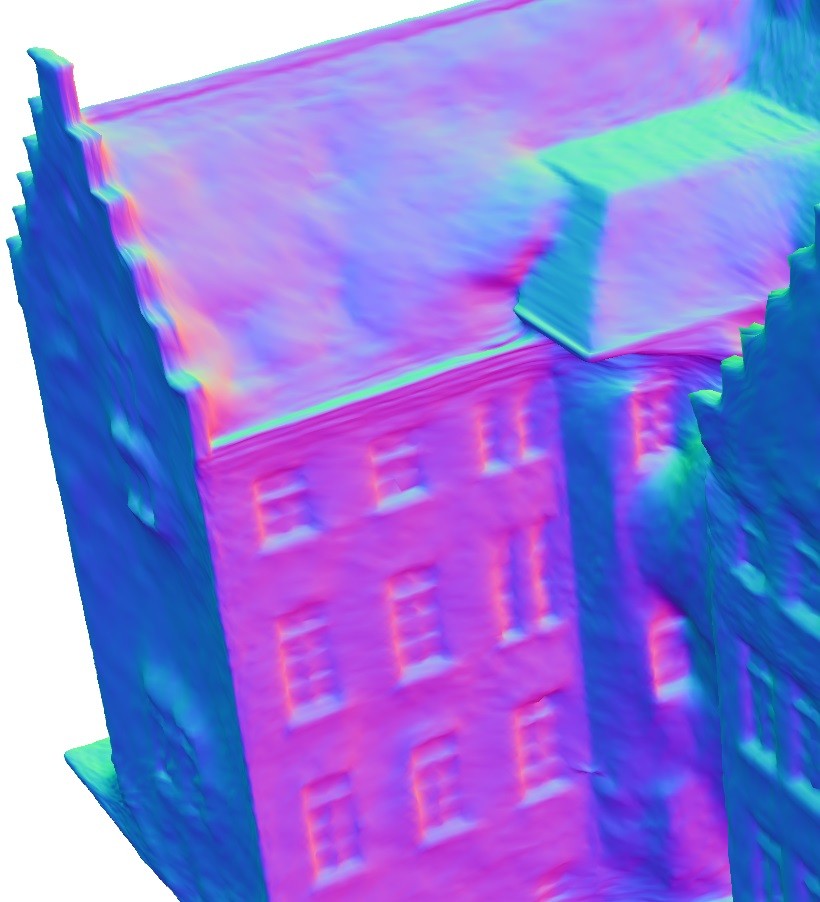} &
    \includegraphics[width=.24\linewidth]{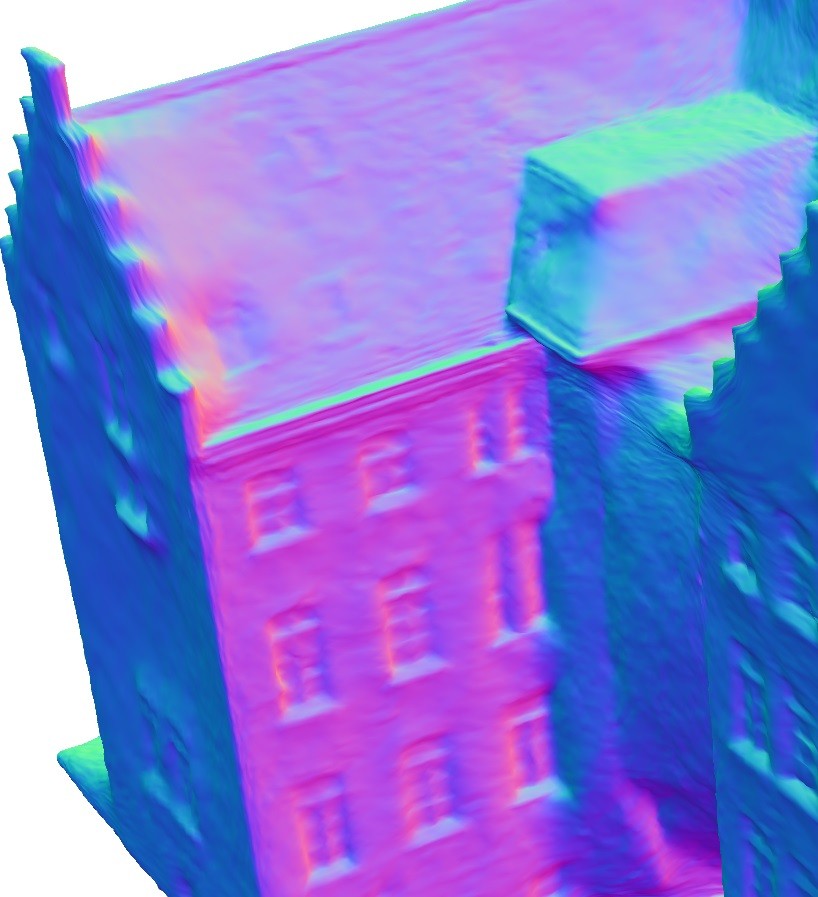} &
    \includegraphics[width=.24\linewidth]{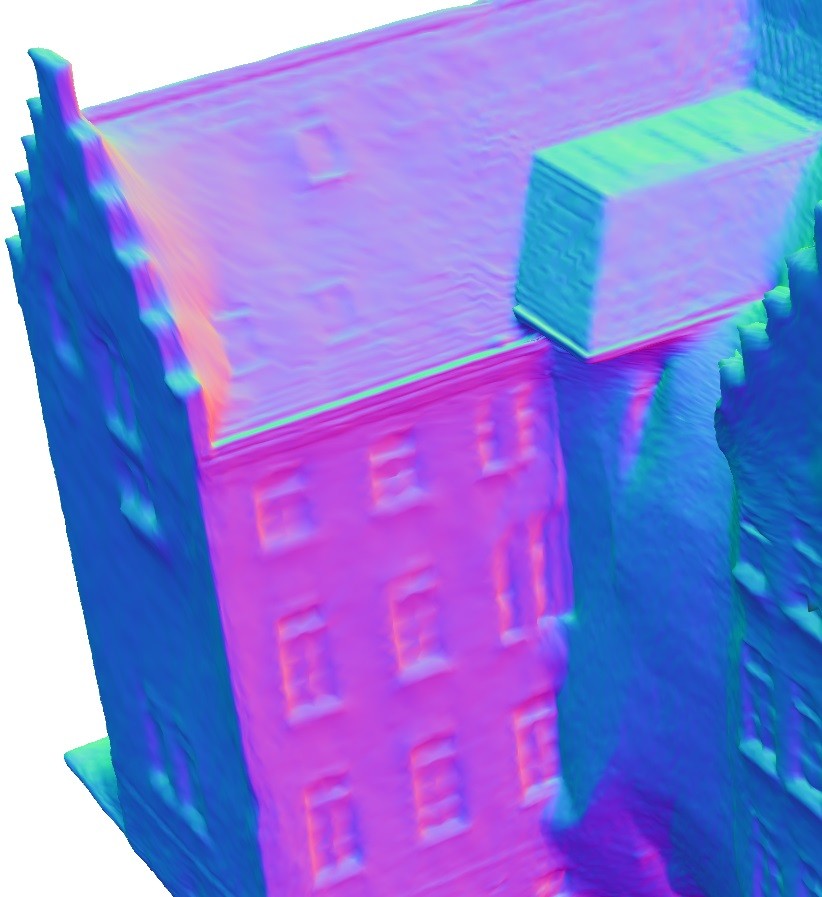} \\
    (a) $\lambda_g=0$ & (b) $\lambda_g=0.1$ & (c) $\lambda_g=0.3$ & (d) $\lambda_g=1$ \\
\end{tabular}
}
\caption{Qualitative comparison for multi-view surface reconstruction with varying geometry shader loss coefficient $\lambda_g$.}
\label{fig:qual_abl_geo}
\end{figure}

\paragraph{Geometry-Based Shader}
\label{geo_based_shader}
While it is beneficial to learn colour variations that are uncorrelated to surface normals, diffuse materials with a strong correlation can instead induce color variations from the feature vector $z(\textbf{x})$ rather than pushing the geometry of $f(x)$ to deform. To mitigate this, we decouple $h$ into two components: a feature-based shader $h_z(x_i,n_i,z_i,C_i)$ predicts a base color image $\tilde{I}_i^z$, which we subsequently feed into a geometry-based shader $h_g(x_i,n_i,\tilde{I}_i^z, C_i)=\tilde{I}_i$. The key important property is that the geometry-shader \emph{detaches} $\tilde{I}_i^z$; that is, we directly prescribe ${\partial \tilde{I}_i}/{\partial \tilde{I}_i^z}=0$ when back-propagating gradients through the architecture. Thus, the gradients of $h_g$ propagate through the geometry directly, and balance the independent training of $z$ and $f$ as illustrated in \Cref{fig:ablated_h_g} and \Cref{fig:qual_abl_geo}. 

\paragraph{Training Details}

For each neural network we use an ADAM \citep{kingma2014adam} optimizer. For the neural shaders $h_z, h_g$ we use a learning rate of $1\cdot 10^{-3}$, and for each deformation network we use a learning rate of $2\cdot 10^{-3}$, which is reduced by a factor of 0.75 at the mesh refinement step. At every iteration we shade $5\%$ of the pixels in the intersection of ground truth and predicted masks, from 6 different randomly sampled views. We set the geometry-based shader $h_g$ loss coefficient $\lambda_g = 0.1$ for all experiments (see \Cref{extended_ablation}). On DTU we train for 500 iterations with the coarse deformation field $f_\text{coarse}$ and coarse mesh, before training for a further 1500 iterations with the full resolution mesh and both deformation fields $f_\text{coarse}, f_\text{fine}$. Note that in this second stage, $f_\text{coarse}$ is ``frozen'' and not optimized for; in practice, we observed that optimizing both networks had no improvements to surface reconstruction, albeit increasing training times. 

In total, our approach takes only $\approx 5$ minutes to train on a single NVIDIA A100 40GB GPU. Rendering and mesh extraction times were measured using the maximum resolution mesh (163,842 vertices), requiring only a forward pass of shallow MLPs (taking 2--5ms).

\paragraph{Training Meshes} We use an icosahedral mesh on the unit sphere that is subdivided to the initial coarse $\mathcal{M}_D$ with 2,562 vertices. This mesh is eventually subdivided (with new vertices being normalized to lie on the sphere) to 163,842 vertices in the second stage (alongside the introduction of $f_{\text{fine}}$). We found that an additional level of division (to 655,362 vertices) offered no clear improvement in Chamfer distance whilst increasing training times and memory requirements.

\paragraph{Positional Encodings}
For the the low-frequency deformation field $f_{\text{coarse}}$ we use a random Fourier feature scale of $\sigma=0.5$. For the high-frequency network $f_{\text{fine}}$ we use a random Fourier features scale of $\sigma=4$, and 2320 eigenfunctions which are pre-computed on the highest resolution input mesh (icosahedron with 163,842 vertices). Specifically, we compute the first 10,000 eigenfunctions, and use a subset of 2320 eigenfunctions. This subset comprises the eigenfunctions selected empirically in a related work \citep{koestler2022intrinsic}, as well as the last 1500 high-frequency eigenfunctions (8500-10000). For the neural shaders $h_g, h_z$, we use Fourier feature positional encodings $\gamma_{\omega}, \gamma_{n}$ of $4$ and $3$ octaves for the view directions $\omega$ and normals $n$ respectively.

\section{Extended Ablations}
\label{extended_ablation}
\begin{figure}[p]
    \centering
    \setlength{\tabcolsep}{0pt}
    \resizebox{\textwidth}{!}{
    \begin{tabular}{cccccc}
        & \multicolumn{1}{c}{}& \multicolumn{1}{c}{}& \multicolumn{1}{c}{}\\
     \rotatebox{90}{\parbox[t]{2cm}{\hspace*{\fill} \text{ENS Render}\hspace*{\fill}}}\hspace*{5pt}
                            &  \includegraphics[width=2cm]{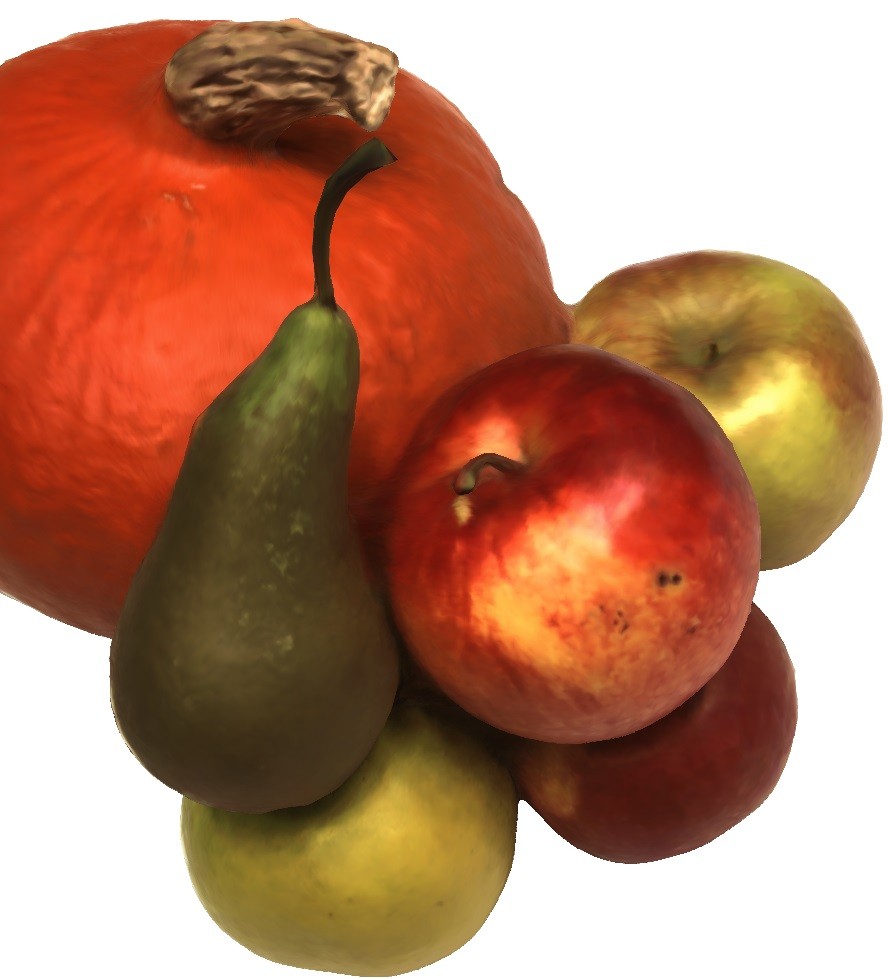}
                            &  \includegraphics[width=2cm]{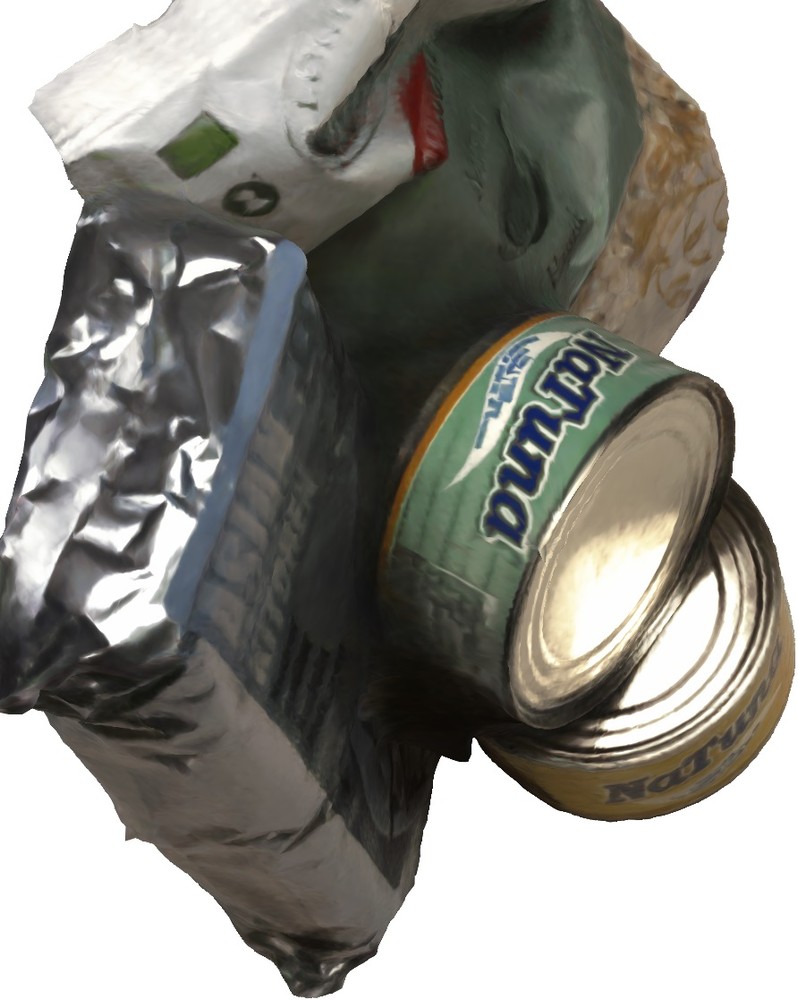}
                            &  \includegraphics[width=2cm]{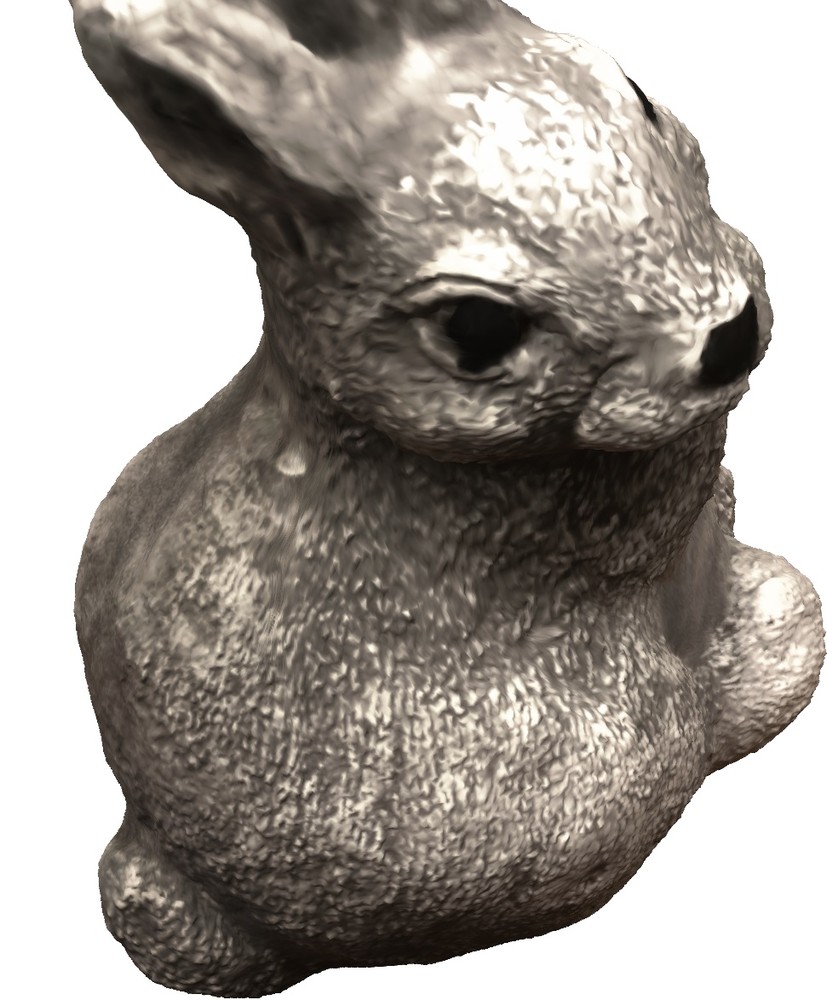}
                            & \includegraphics[width=2cm]{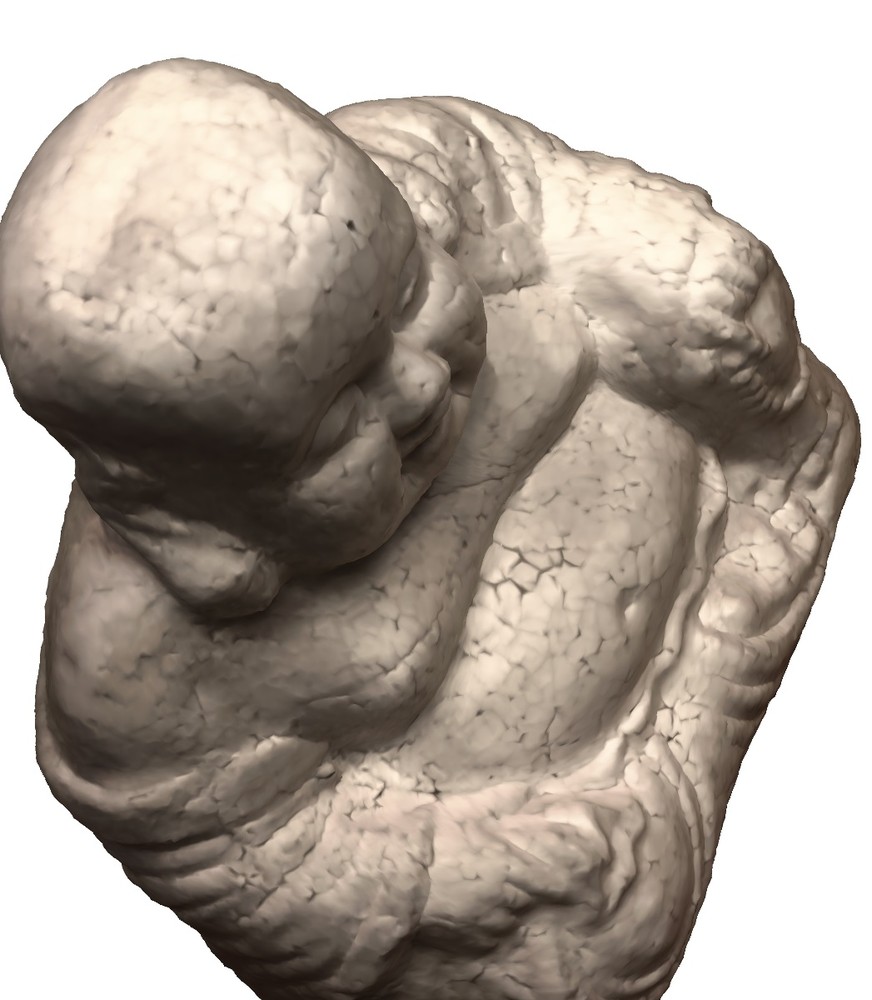}
                            &  \includegraphics[width=2cm]{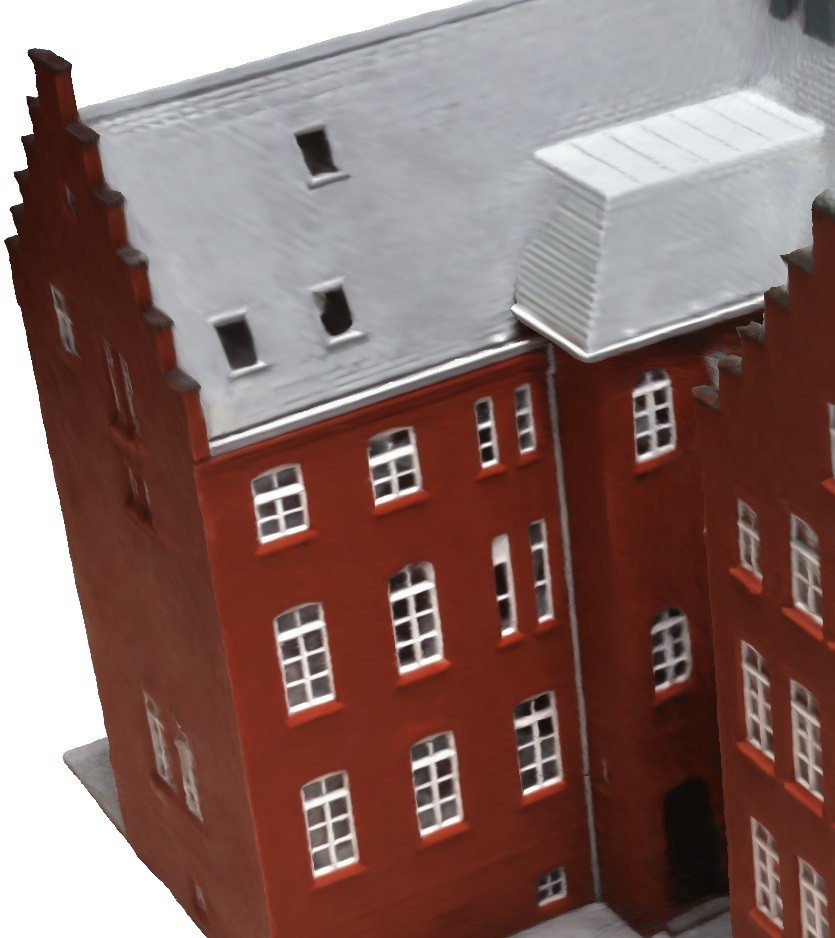}\\
     \rotatebox{90}{\parbox[t]{2cm}{\hspace*{\fill}$\gamma_E$ \text{Render}\hspace*{\fill}}}\hspace*{5pt}
                            &  \includegraphics[width=2cm]{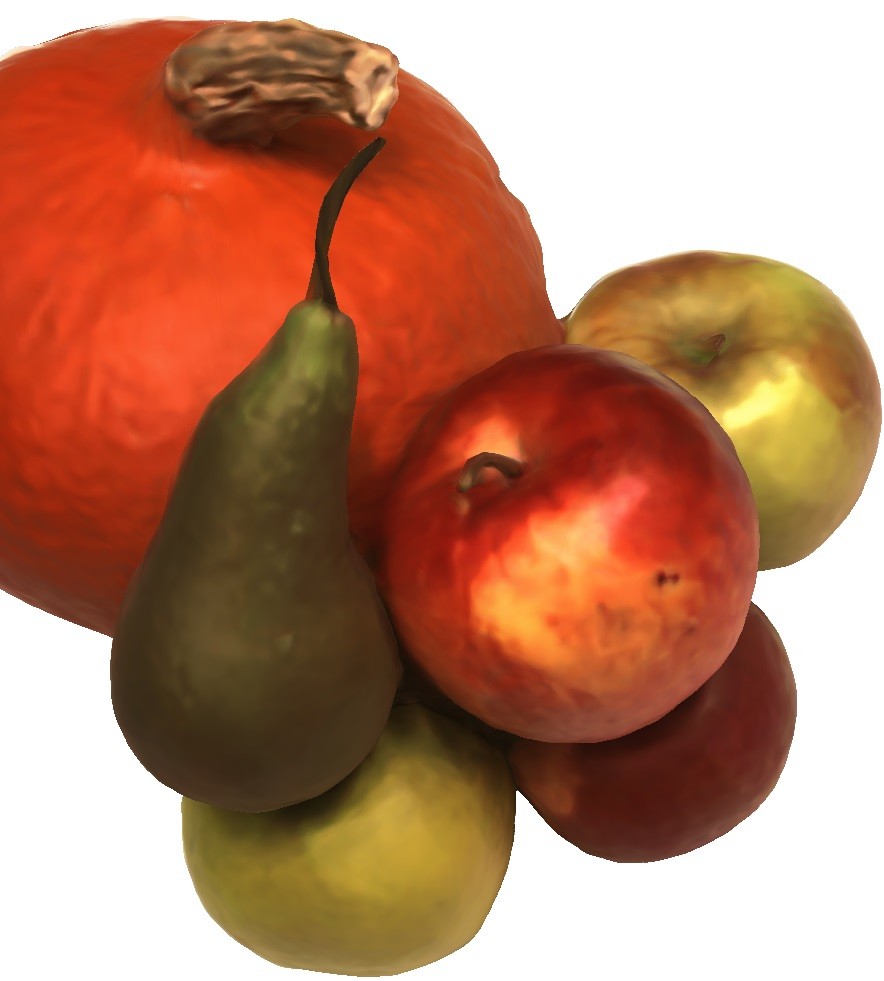}
                            &  \includegraphics[width=2cm]{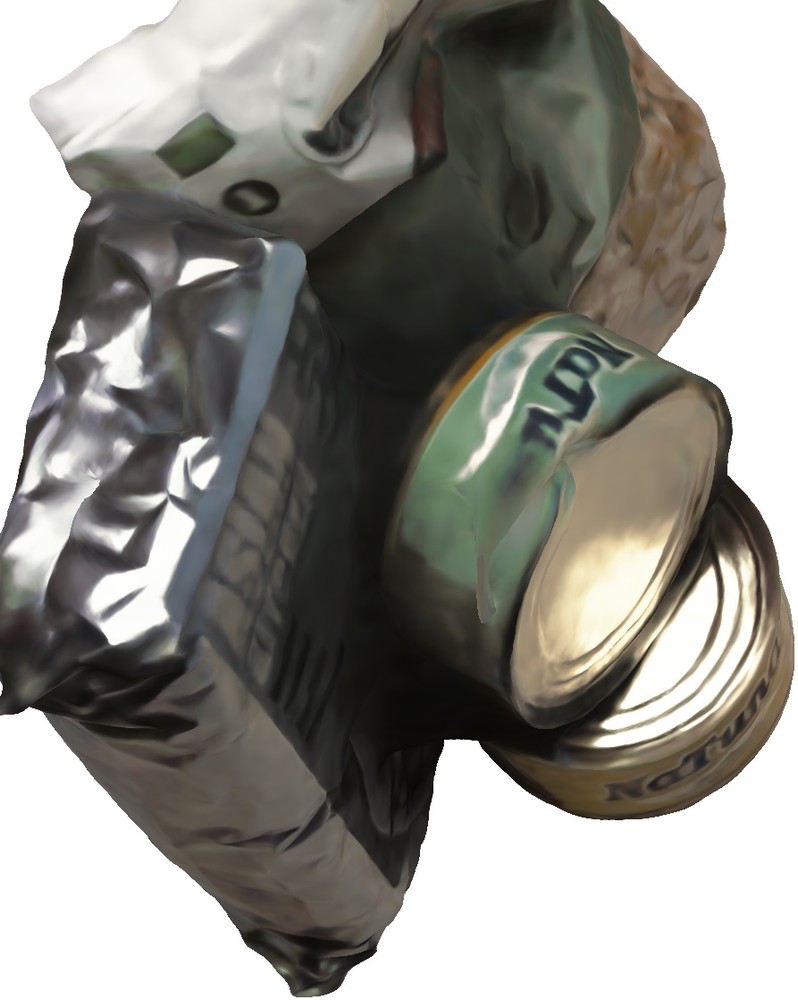}
                            &  \includegraphics[width=2cm]{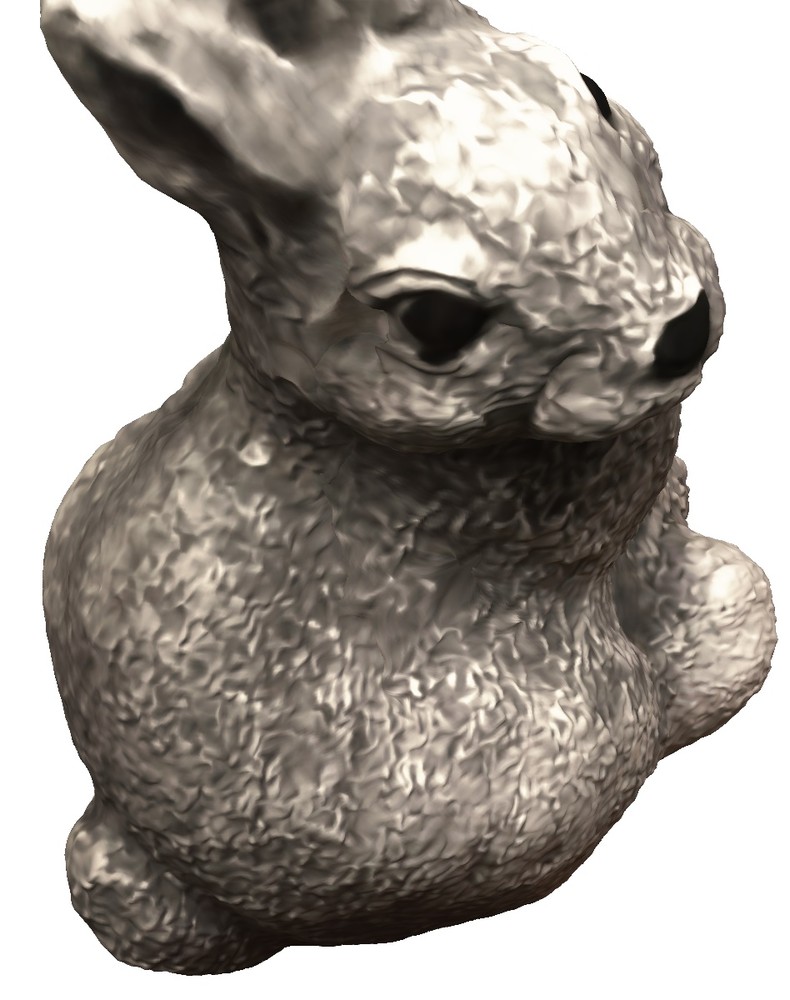}
                            & \includegraphics[width=2cm]{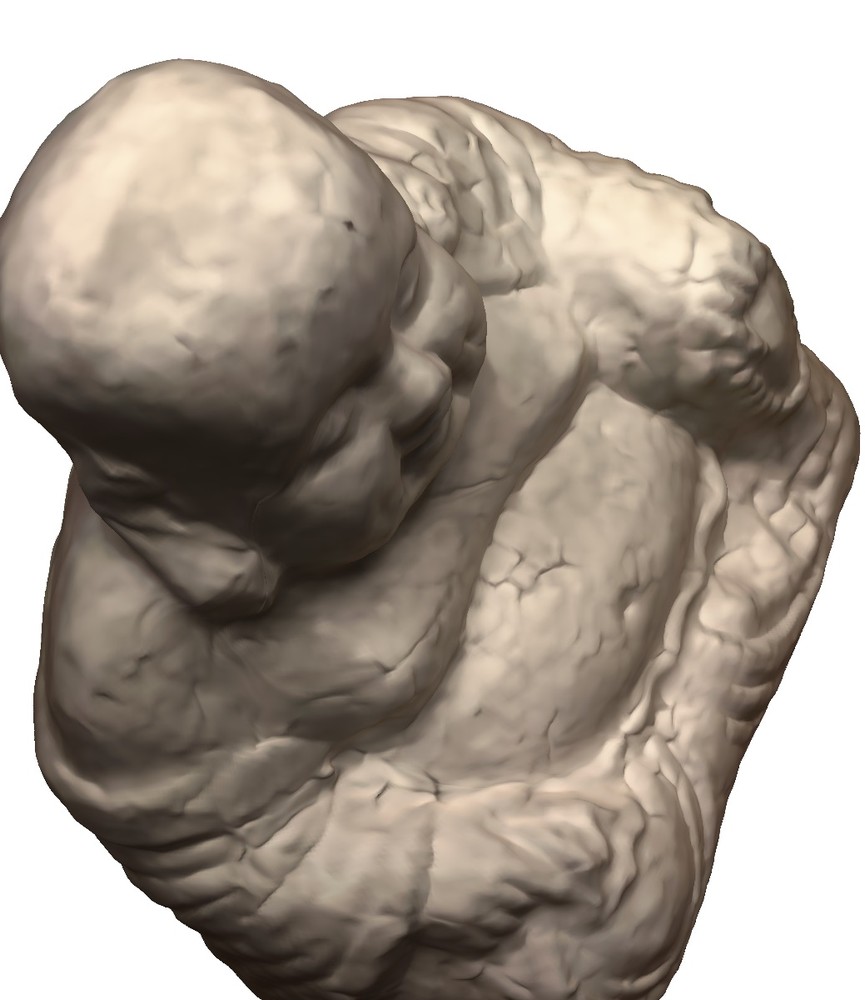}
                            &  \includegraphics[width=2cm]{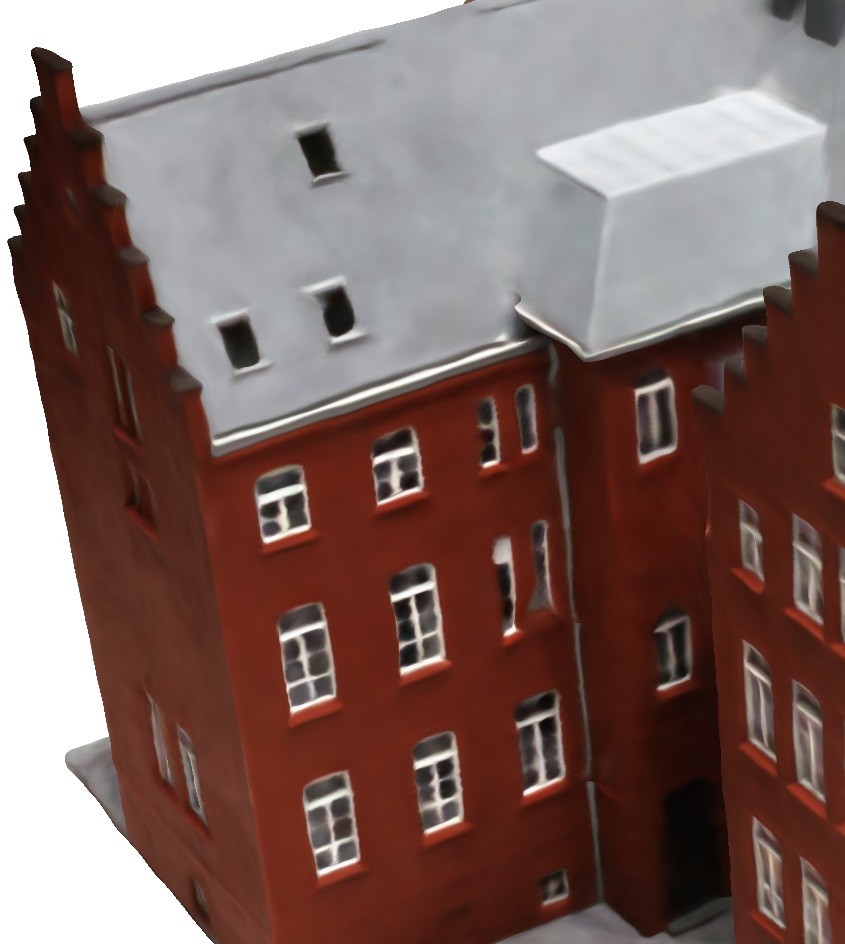}\\
      
     \rotatebox{90}{\parbox[t]{2cm}{\hspace*{\fill} \text{ENS Normal}\hspace*{\fill}}}\hspace*{5pt}
                            &  \includegraphics[width=2cm]{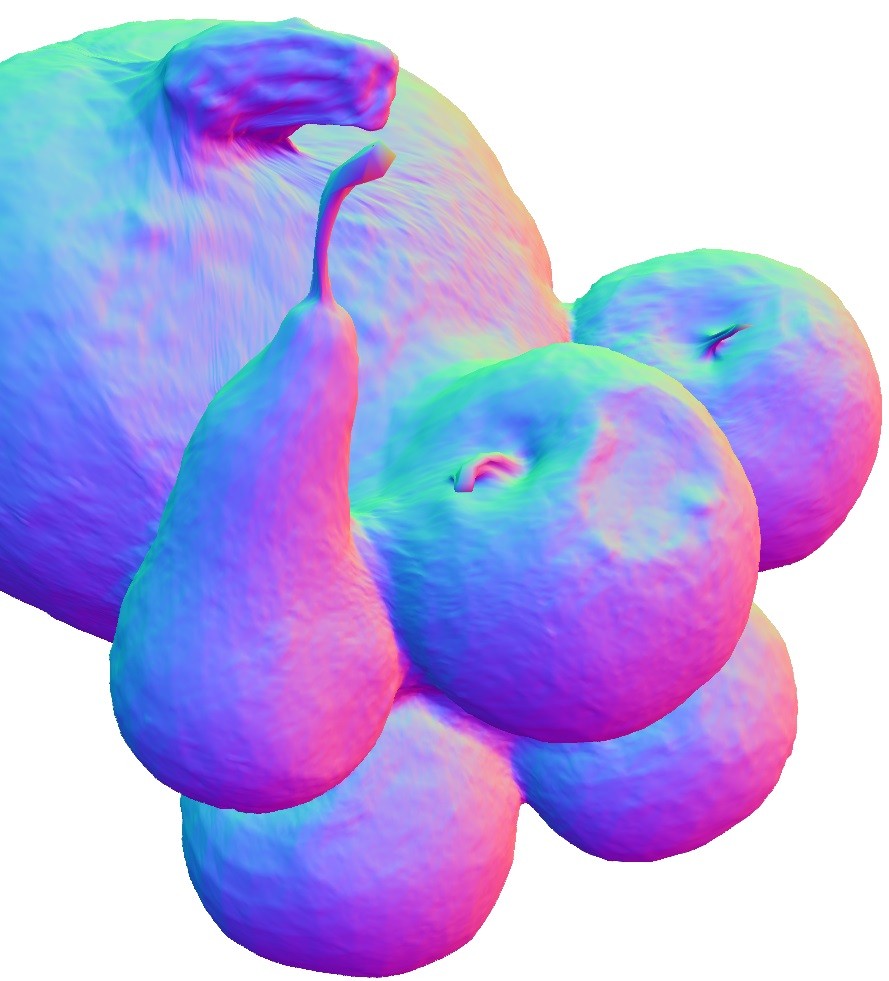}
                            &  \includegraphics[width=2cm]{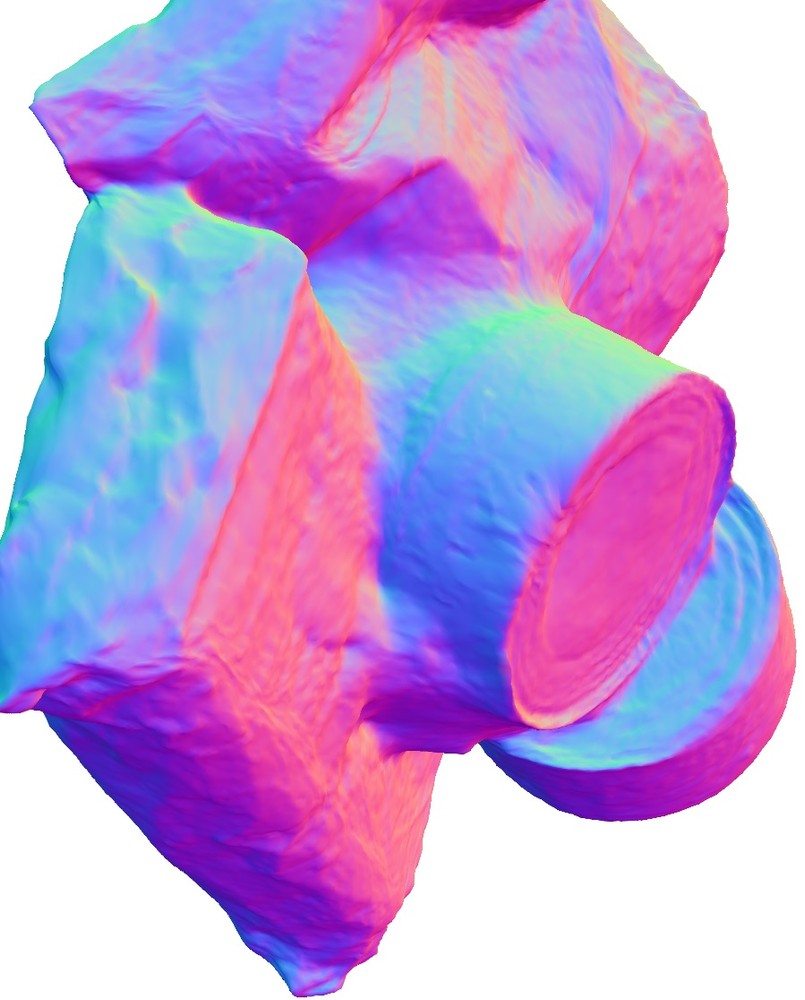}
                            &  \includegraphics[width=2cm]{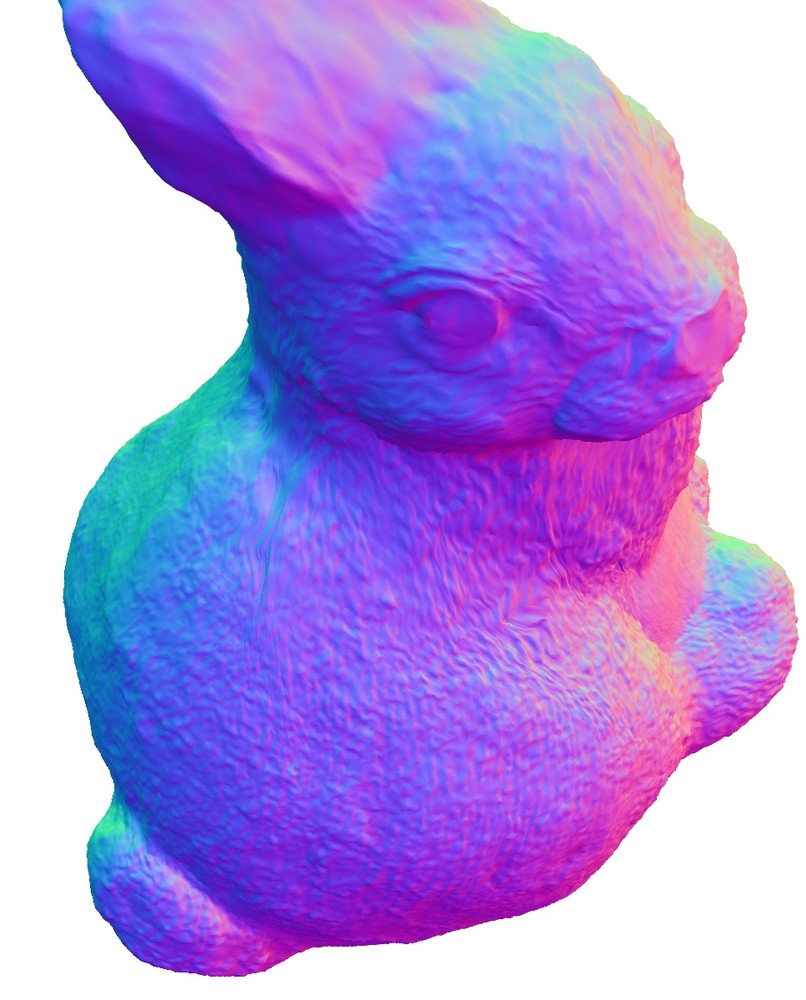}
                            & \includegraphics[width=2cm]{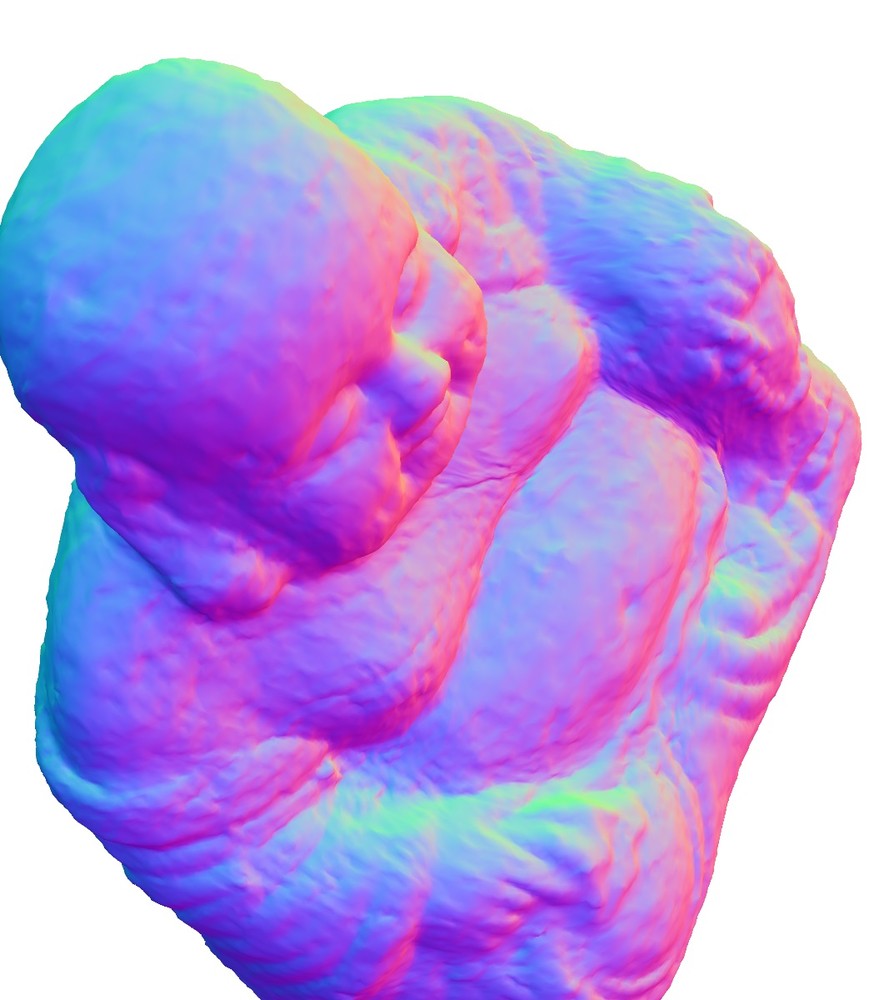}
                            &  \includegraphics[width=2cm]{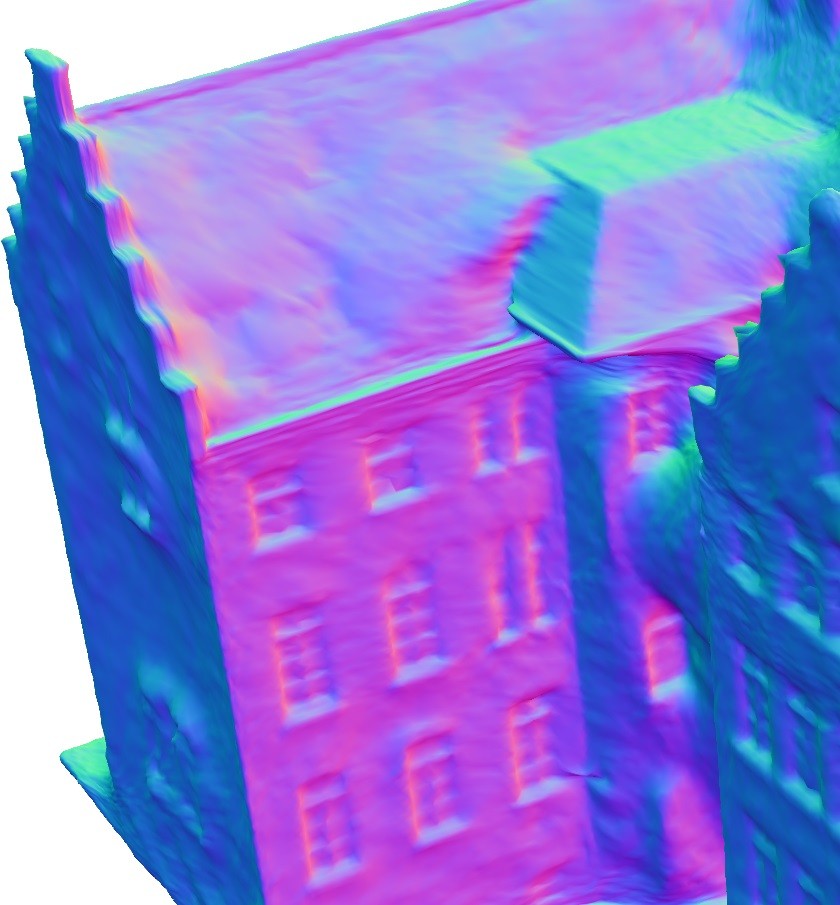}\\ 
    \rotatebox{90}{\parbox[t]{2cm}{\hspace*{\fill}$\gamma_E$ \text{Normal}\hspace*{\fill}}}\hspace*{5pt}
                            &  \includegraphics[width=2cm]{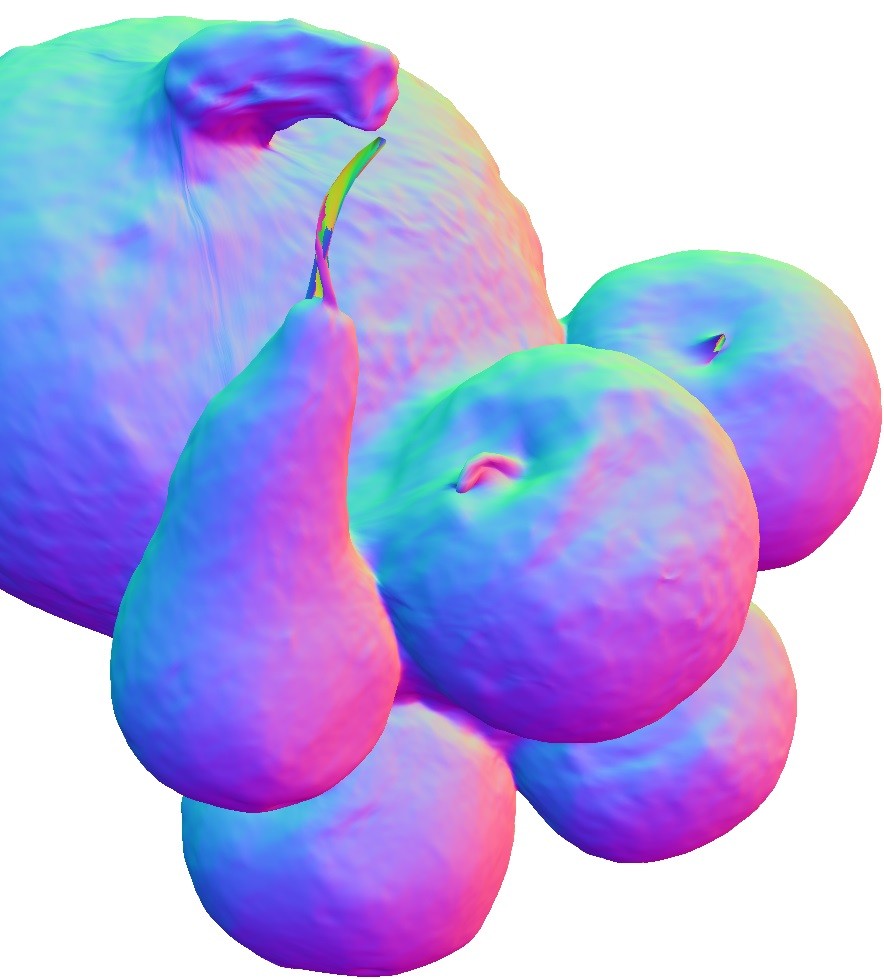}
                            &  \includegraphics[width=2cm]{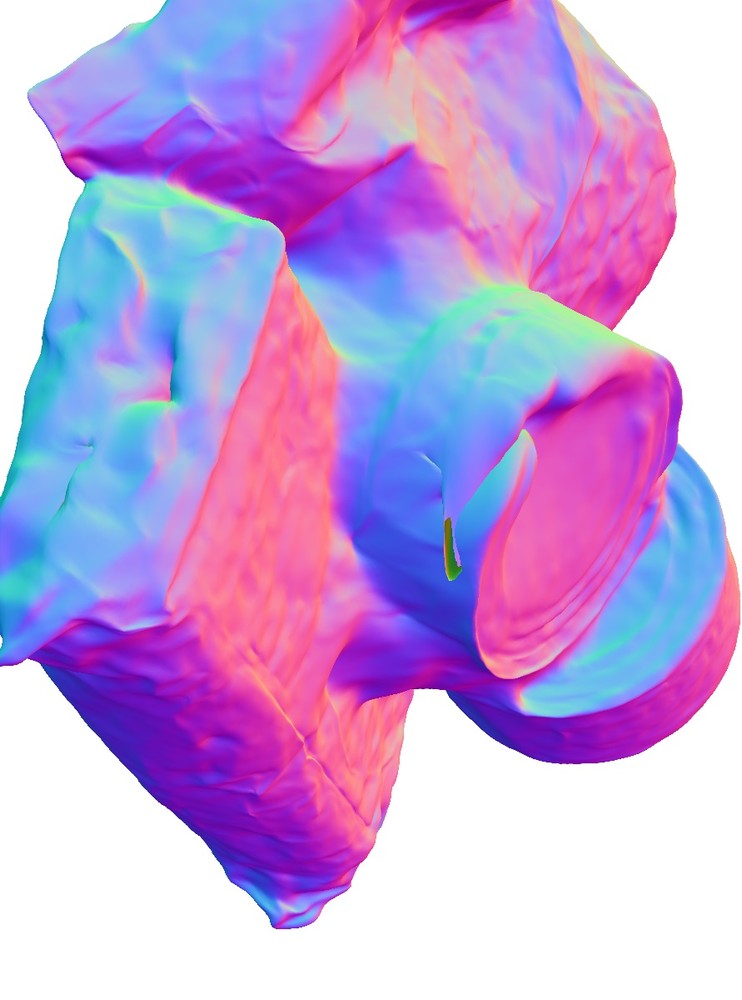}
                            &  \includegraphics[width=2cm]{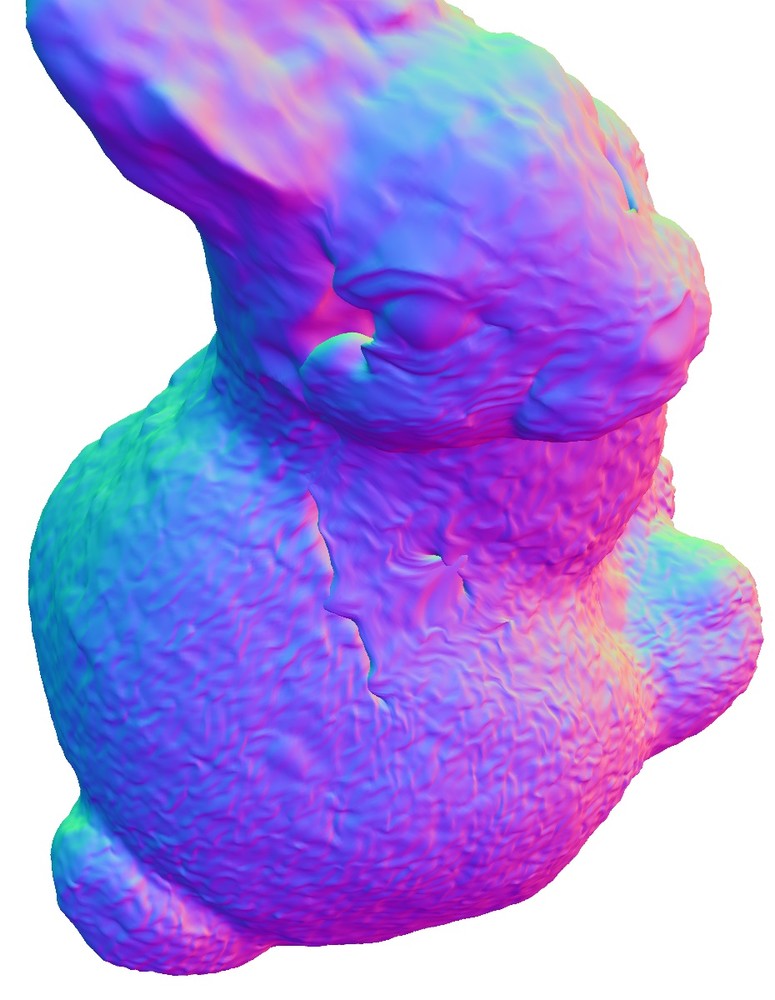}
                            & \includegraphics[width=2cm]{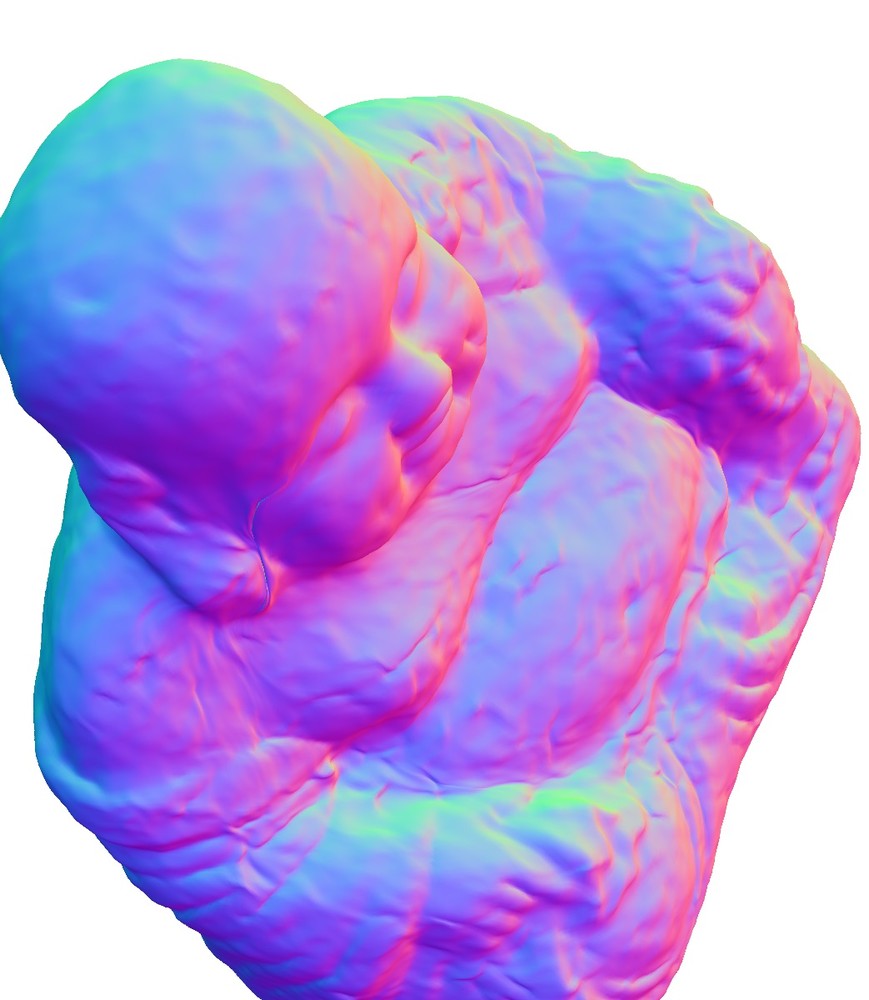}
                            &  \includegraphics[width=2cm]{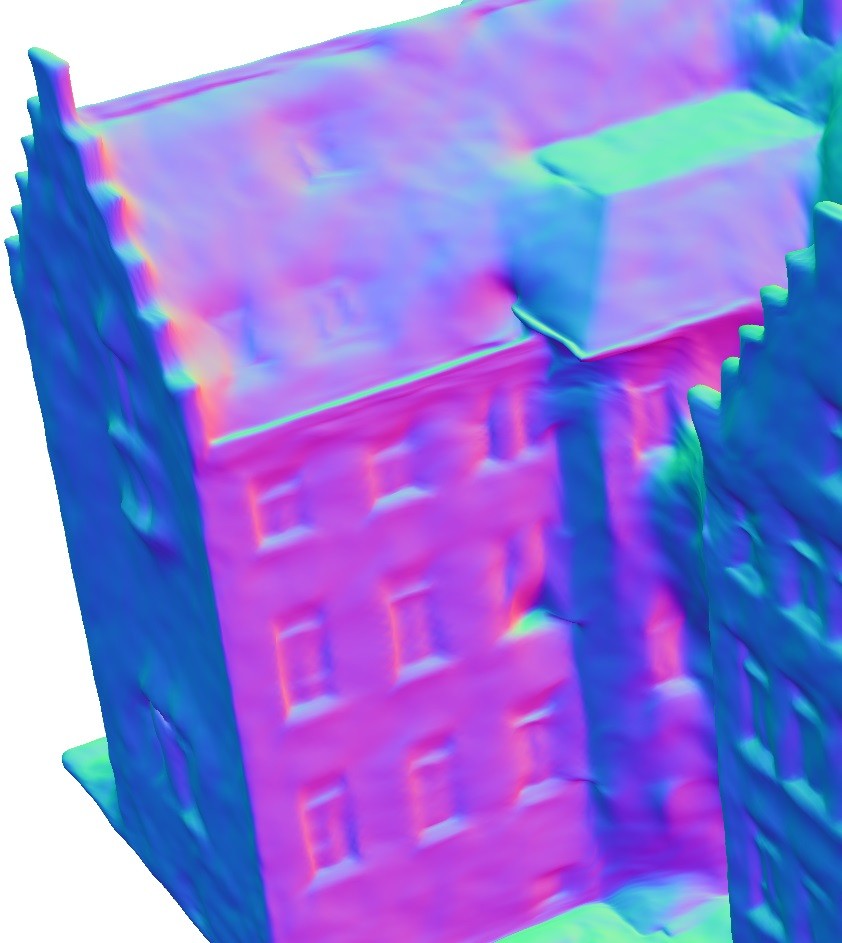}\\
    \end{tabular}
    }
    \caption{Qualitative comparison for multi-view surface reconstruction renders (top) and normal maps (bottom) between using the full model (ENS) with hybrid encoding $\gamma_H$ and an extrinsic-only encoding $\gamma_E$, best viewed in high resolution.}
    \label{fig:intrinsic_abl}
\end{figure}

\paragraph{Intrinsic Encodings}
In \Cref{fig:intrinsic_abl} we compare the renders and normal maps of the full ENS model, using a hybrid encoding $\gamma_H$ of (extrinsic) random Fourier features and (intrinsic) eigenfunctions, and a model using only extrinsic positional encoding $\gamma_E$. We observe sharper renders with $\gamma_H$, and more accurate surface details. For example, the bunny and redhouse have high-frequency textures and surface details which the extrinsic-only model $\gamma_E$ struggles to resolve (see windows). In our experiments we observed that this can result in inaccurate surface deformations and lead to poor local minima. 

\paragraph{Dual Shader Ablation} In the appearance-based loss methodology (\Cref{sec:losses}) of the main paper), we therefore define the photometric loss of our model as a combination of losses from the feature-based shader $h_z$ and the geometry-based shader $h_g$, the latter of which is scaled by coefficient $\lambda_g$.  In \Cref{fig:qual_abl_geo} we demonstrate the effect of varying the loss coefficient $\lambda_g$ on the learned geometry for scan 24 (redhouse). When using no geometry-based shader ($\lambda_g = 0$), we observe unwanted extraneous growths on the roofs where the normal-colour correlation is strong. By increasing the coefficient we clearly see the effect of the geometry shader enabling these regions to be reconstructed correctly. However, we also see this comes at the cost of less pronounced concavities. In our experiments, we found setting $\lambda_g = 0.1$ across all scenes strikes a good balance between these behaviours. nevertheless, we note this can act as a useful hyperparameter to control a prior on normal-geometry dependence.   

\subsection{Non-Spherical Topology}
\label{nonsphere}
\begin{figure}[t]
\centering
\begin{tabular}{ccc}
    \includegraphics[width=.24\linewidth]{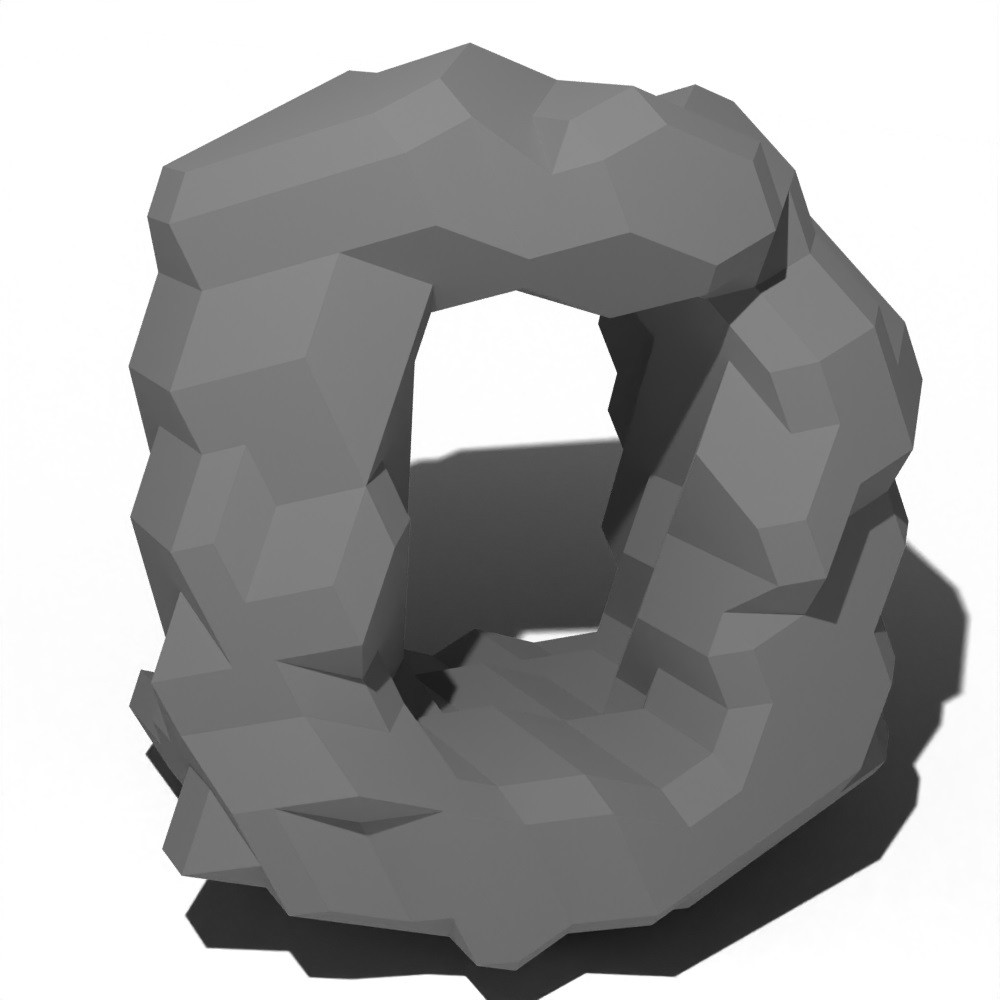} &
    \includegraphics[width=.24\linewidth]{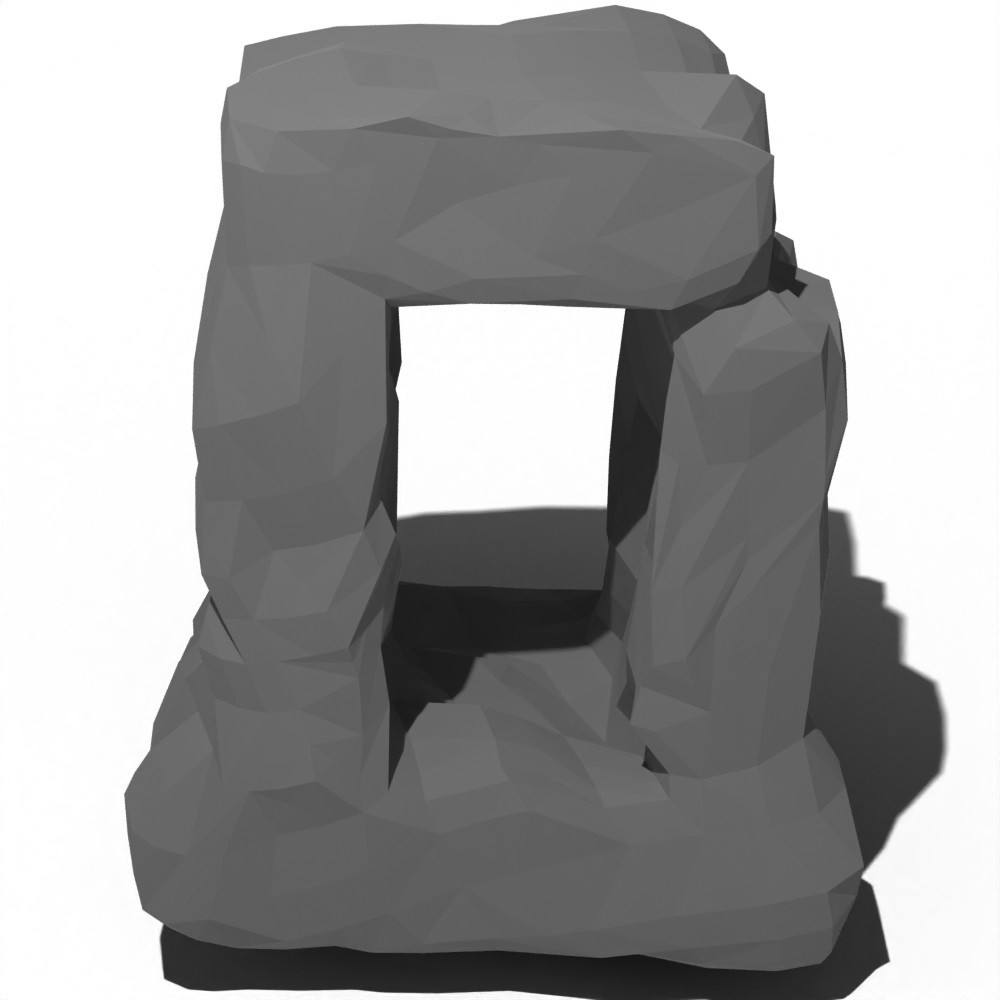} &
    \includegraphics[width=.24\linewidth]{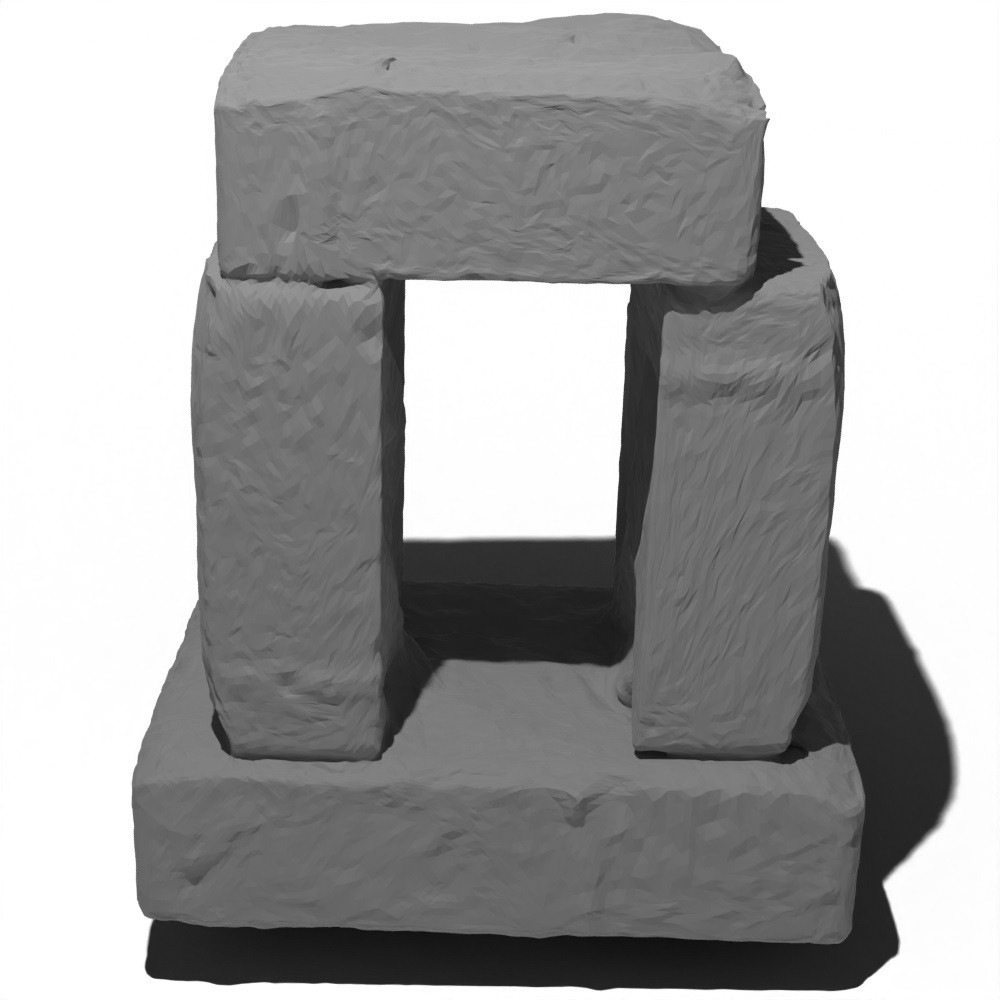} \\
    (a) Iteration 0 & (b) Iteration 100 & (c) Iteration 2000 \\
\end{tabular}
\caption{We demonstrate that a visual hull (a) can be extracted from input masks and used as the input domain with correct topology.}
\label{fig:non_sph}
\end{figure}

 Although we use a spherical input domain for most scenes, any domain $D$ could potentially be used. For scan 40 we use an initial torus mesh, and for scans 106 and 65 we follow \cite{worchel2022multi}, and generate a topologically-aligned input domain $D$ by extracting a visual hull from the ground-truth masks $\{m_i\}$, which is then linearly subdivided throughout training. The hull is extracted by projecting a $(32 \times 32 \times 32)$ volumetric grid into image space and removing any points not contained in the masks. We then run marching cubes to extract an input domain with the correct topology, see \Cref{fig:non_sph}. We 
refer the interested reader to NDS \cite{worchel2022multi} for more details. The initial hull is then linearly subdivided to $\approx 70K$ vertices. For these cases, we compute eigenfunctions on the subdivided visual hull.

\paragraph{Divergent Models}
\label{diverge}
In our ablation, we consider a model without the coarse-to-fine strategy and instead attempt to learn detailed models directly with a high-frequency deformation field. We remove $f_\text{coarse}$ and train only $f_\text{fine}$ with the highest resolution mesh. As illustrated in \Cref{fig:failed}, this model struggles to capture the coarse shape and quickly falls into bad minima as a result of its spectral bias. 
\section{Extended Remeshing and Mesh Quality Results}
\label{remesh}
We next provide examples of the meshes extracted from ENS as compared to the neural implicit model NeuS (\cite{NeuS}). We demonstrate the superior mesh quality our approach, especially at low resolutions. To illustrate the continuity of the underlying representation, we also show quad meshes extracted from ENS. Note that to get eigenfunction values on the quad mesh vertices, we interpolate them barycentrically from the original icosahedral triangle mesh.

\begin{figure}[t]
    \centering
    \includegraphics[width=\textwidth]{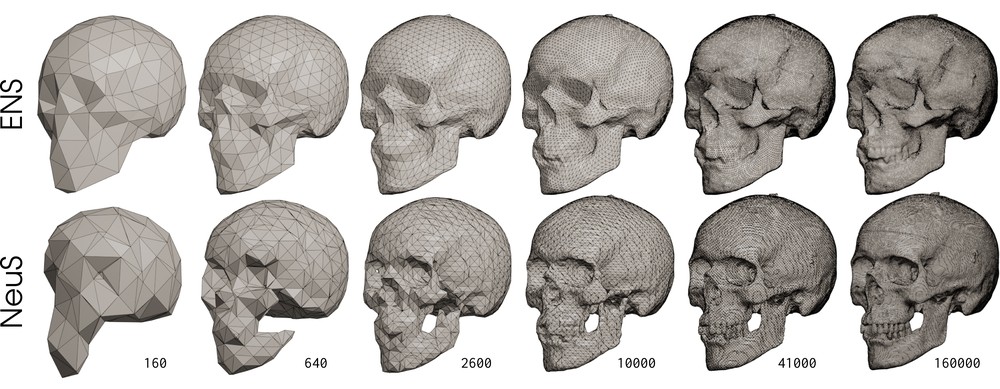}
    \vspace{3pt}
    \caption{Meshes with varying (and top-bottom comparable) number of vertices extracted from a fully-trained model for our method (top) and NeuS~\cite{NeuS} (bottom). It is evident our approach produces topologically-consistent quality meshes without spurious holes and grid artifacts (see forehead) in any resolution.}
    \label{fig:multires}

\end{figure}

\begin{figure}[p]
    \centering
    \setlength{\tabcolsep}{0pt}
    \resizebox{\textwidth}{!}{
    \begin{tabular}{ccccc}
        & \multicolumn{1}{c}{}& \multicolumn{1}{c}{}& \multicolumn{1}{c}{}\\
     \rotatebox{90}{\parbox[t]{2cm}{\hspace*{\fill}\small{\text{NeuS}}\hspace*{\fill}}}\hspace*{5pt}
                            &  \includegraphics[width=1.7cm]{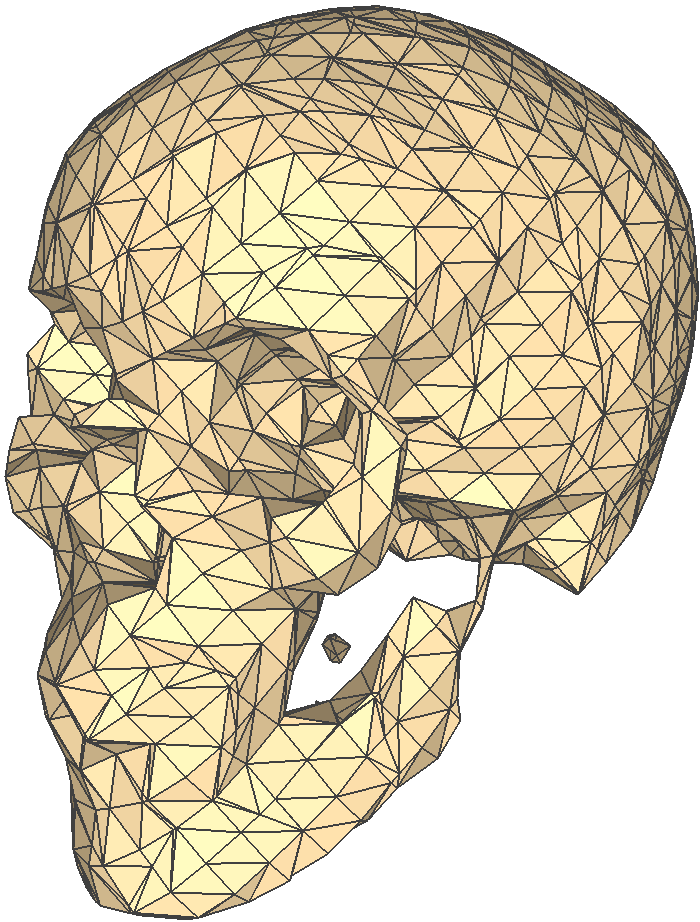} \hspace{3mm}
                            &  \includegraphics[width=2cm]{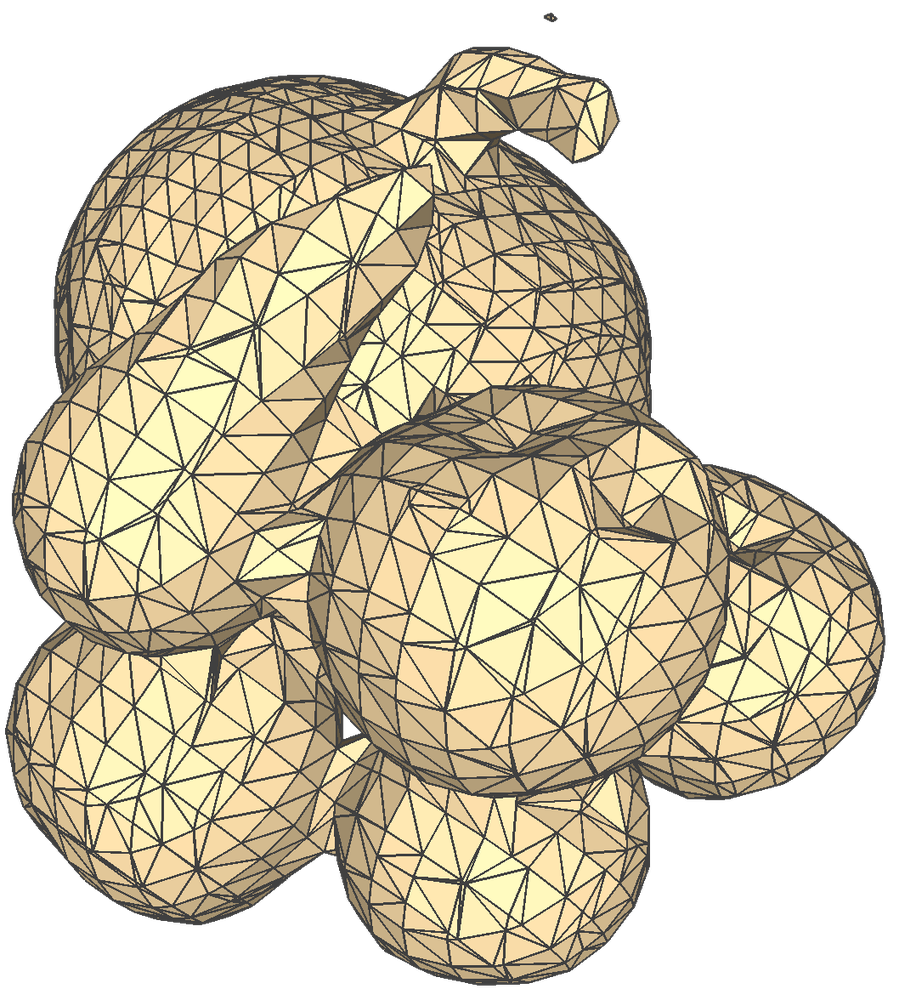} \hspace{2mm}
                            &  \includegraphics[width=2cm]{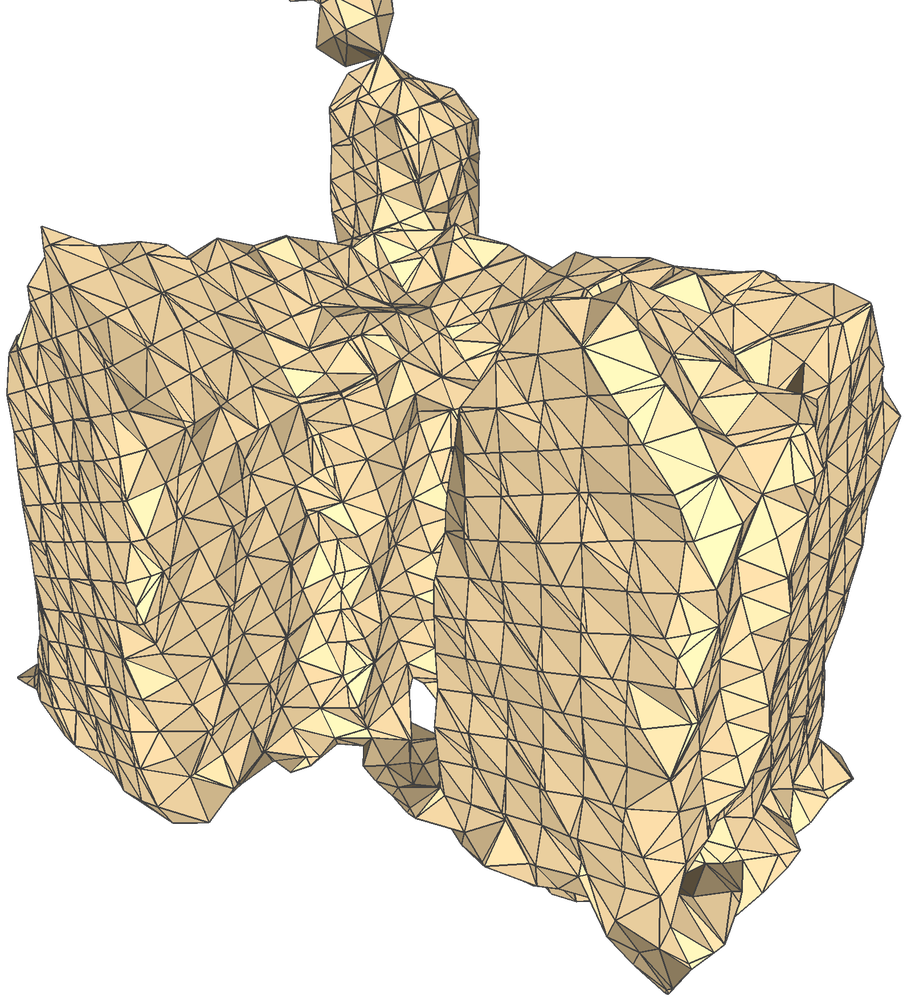}\\
     \rotatebox{90}{\parbox[t]{2cm}{\hspace*{\fill}\small{\text{ENS (Tri)}}\hspace*{\fill}}}\hspace*{5pt}
                            &  \includegraphics[width=1.7cm]{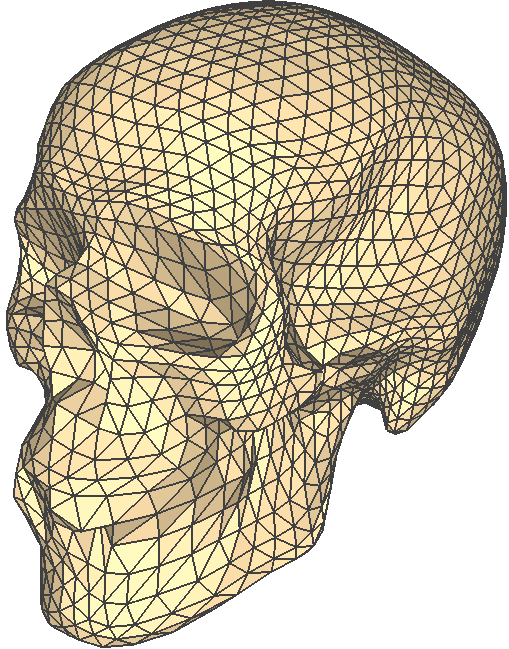}
                            \hspace{3mm}
                            &  \includegraphics[width=2cm]{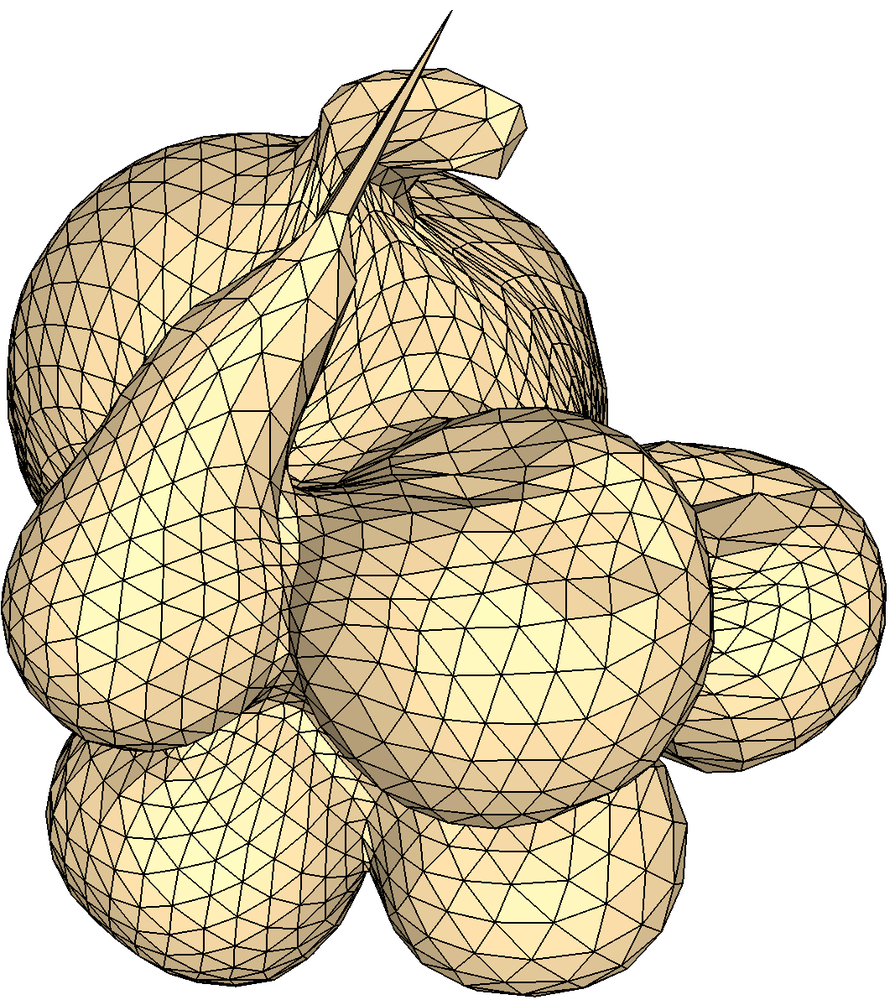}
                            \hspace{2mm}
                            &  \includegraphics[width=2cm]{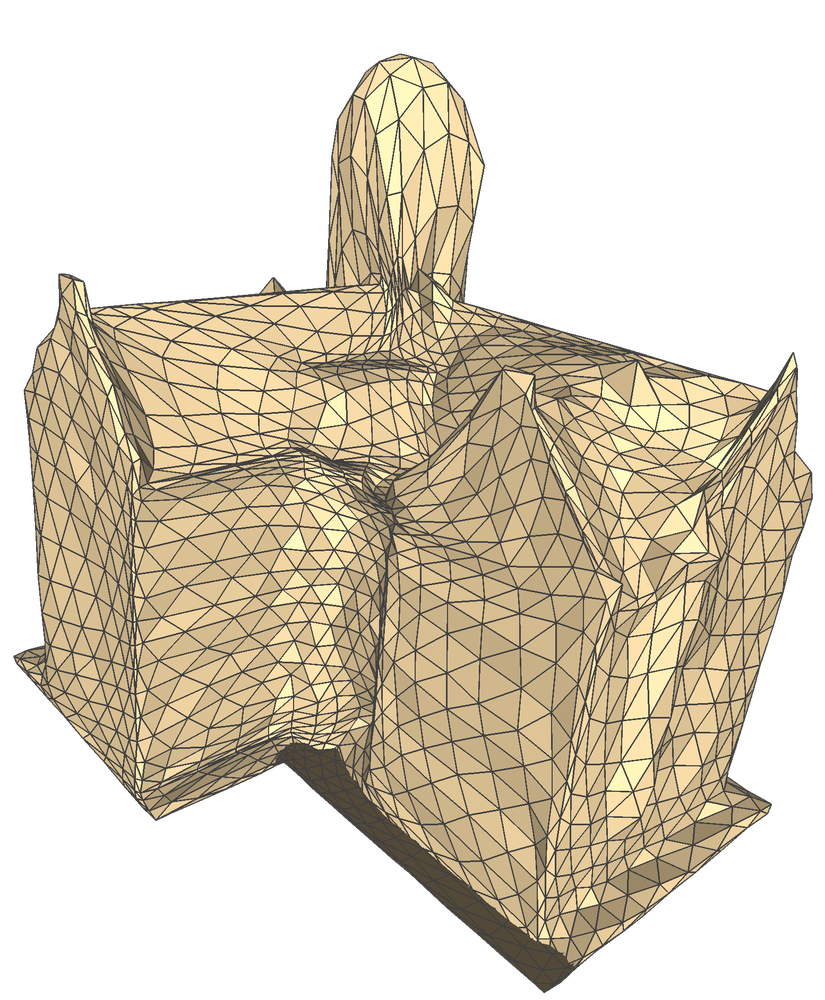}\\
      
     \rotatebox{90}{\parbox[t]{2cm}{\hspace*{\fill}\small{\text{ENS (Quad)}}\hspace*{\fill}}}\hspace*{5pt}
                            &  \includegraphics[width=1.7cm]{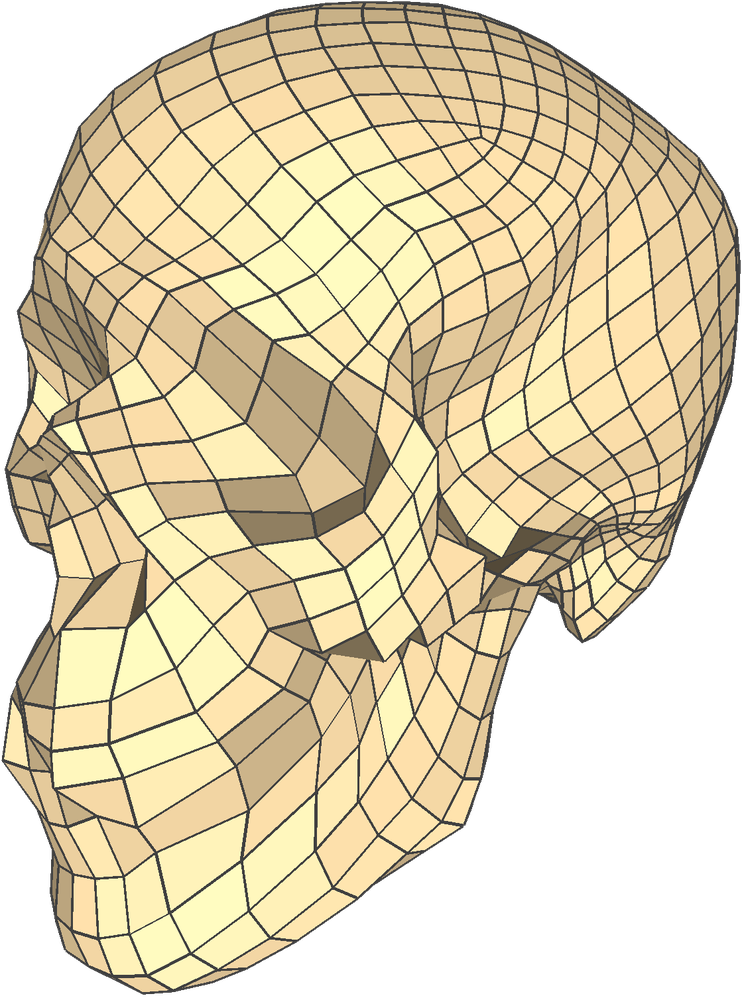}
                            \hspace{3mm}
                            &  \includegraphics[width=2cm]{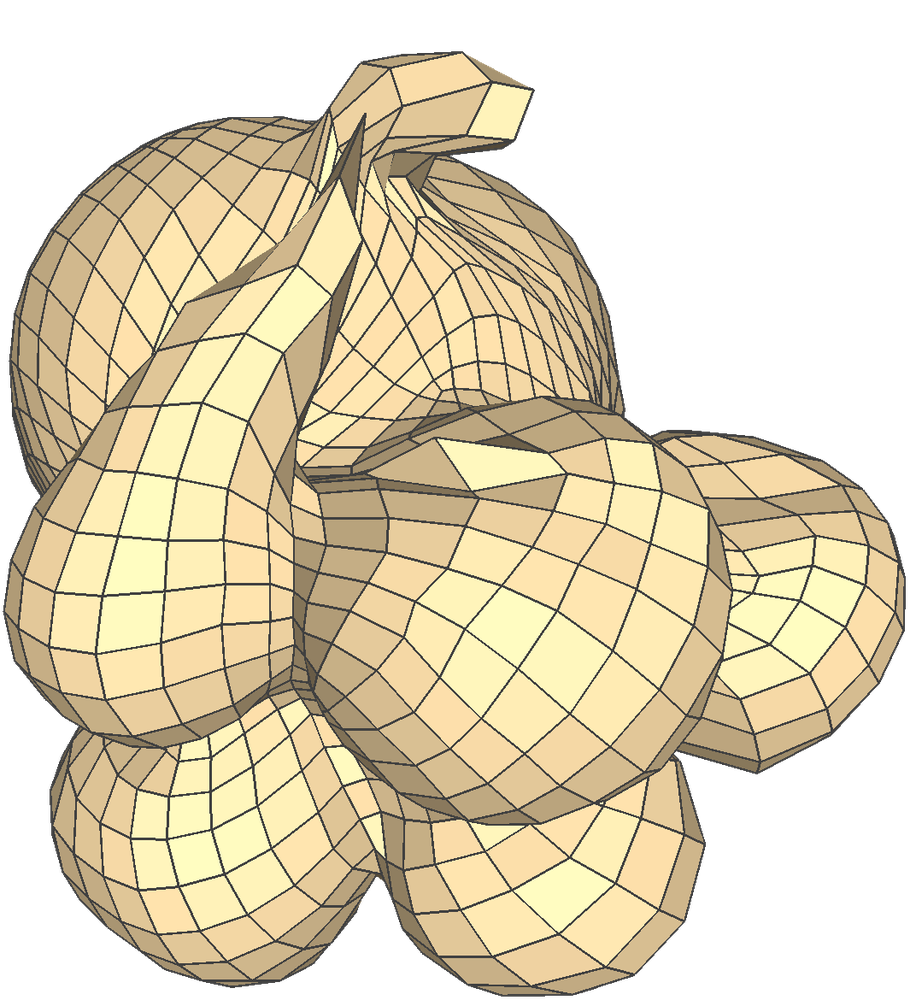}
                            \hspace{2mm}
                            &  \includegraphics[width=2cm]{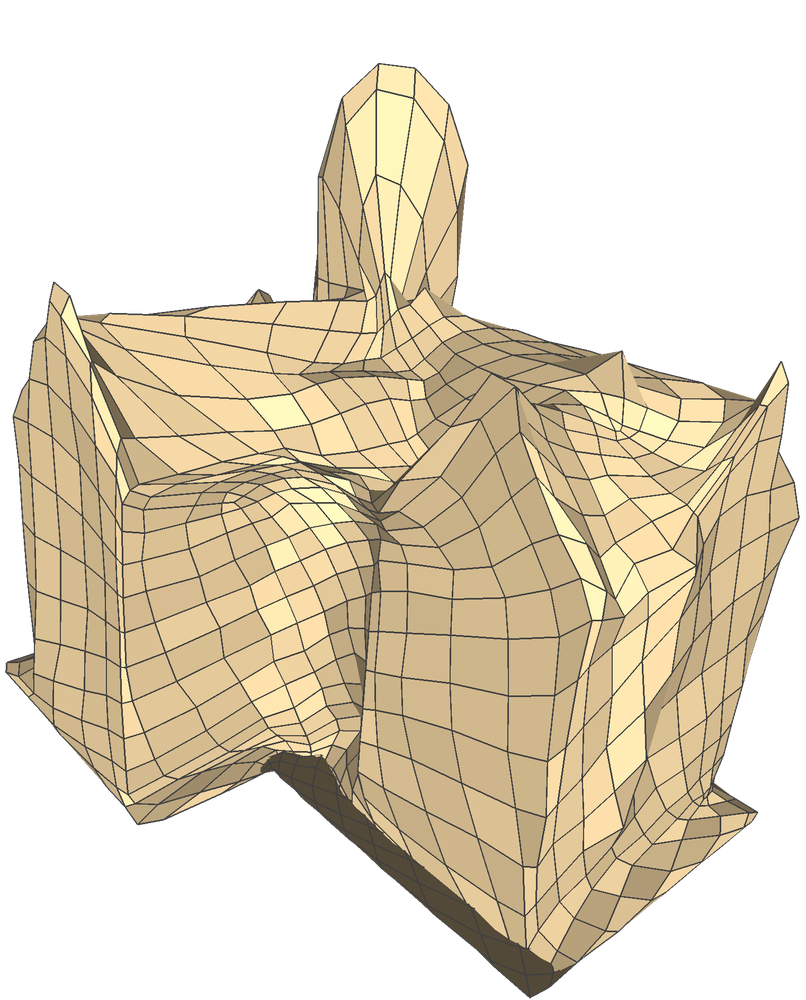}\\

    \end{tabular}
    }
    \caption{Meshes with $2.5$K vertices extracted from NeuS compared to ENS. Note that we use a smaller quad mesh with $1.5$K vertices.}
    \label{fig:2.5K}
\end{figure}

\begin{figure}[p]
    \centering
    \setlength{\tabcolsep}{0pt}
    \resizebox{\textwidth}{!}{
    \begin{tabular}{ccccc}
        & \multicolumn{1}{c}{}& \multicolumn{1}{c}{}& \multicolumn{1}{c}{}\\
     \rotatebox{90}{\parbox[t]{2cm}{\hspace*{\fill}\small{\text{NeuS}}\hspace*{\fill}}}\hspace*{5pt}
                            &  \includegraphics[width=1.7cm]{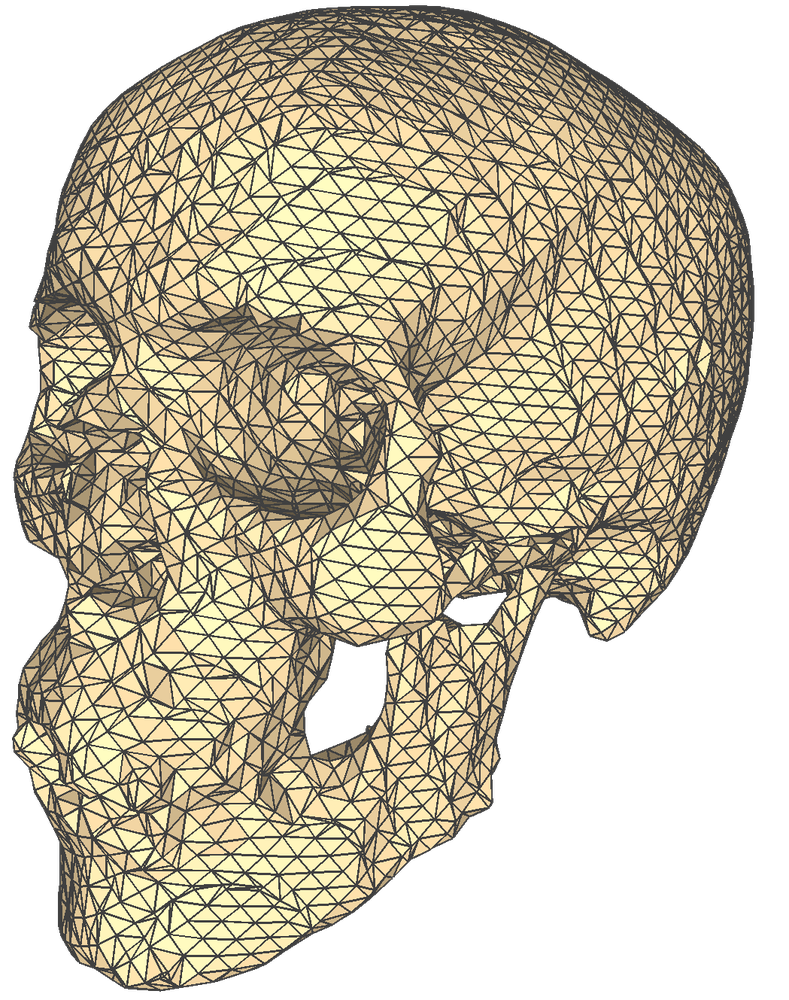} \hspace{2mm}
                            &  \includegraphics[width=2cm]{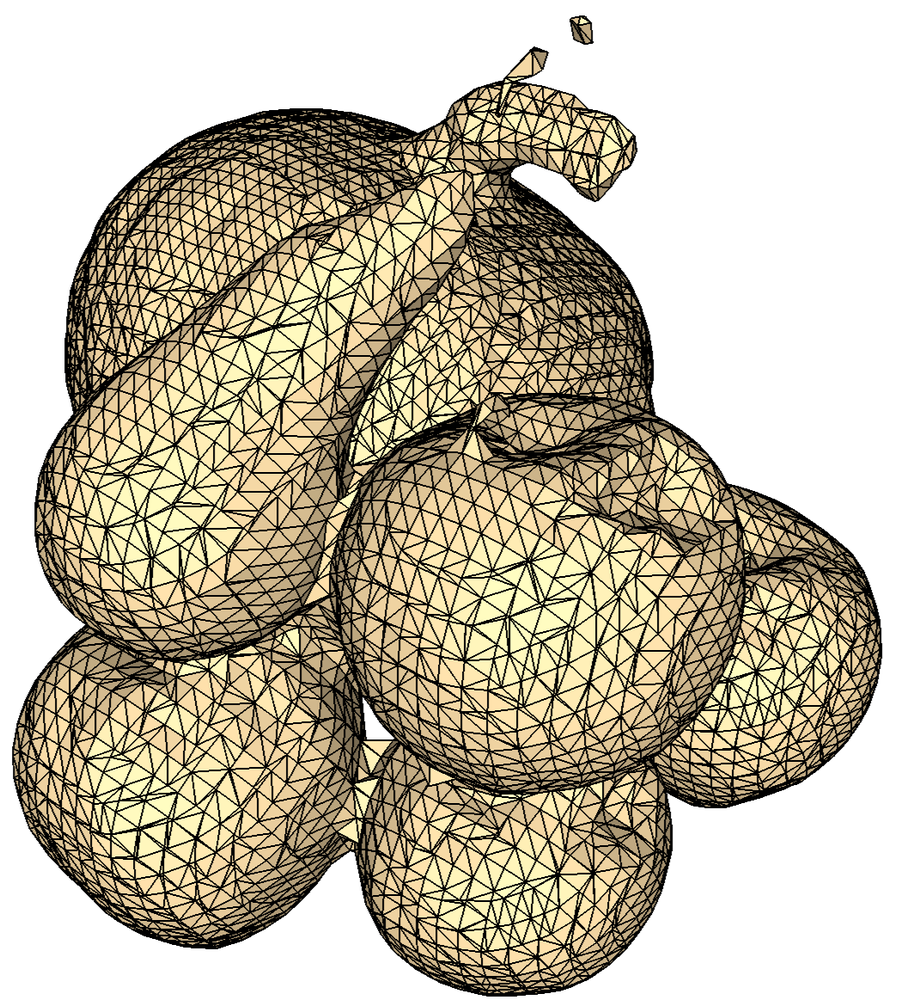} \hspace{1mm}
                            &  \includegraphics[width=2cm]{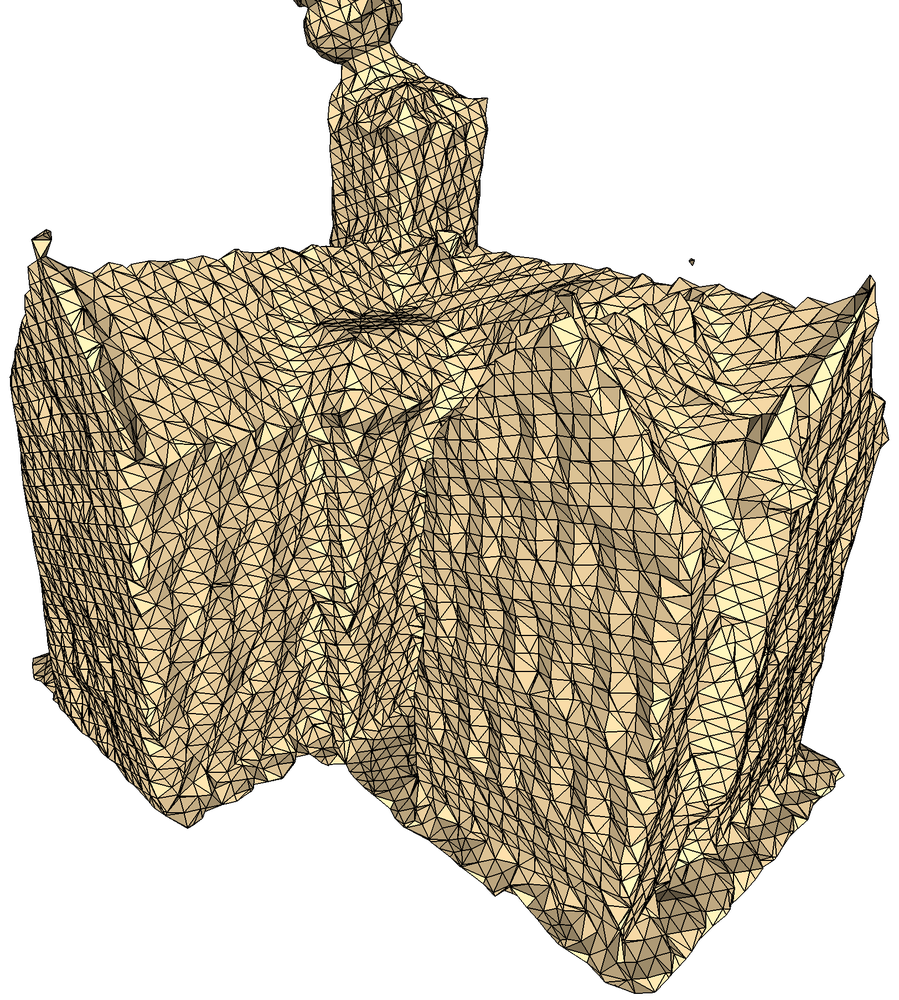}\\
     \rotatebox{90}{\parbox[t]{2cm}{\hspace*{\fill}\small{\text{ENS (Tri)}}\hspace*{\fill}}}\hspace*{5pt}
                            &  \includegraphics[width=1.7cm]{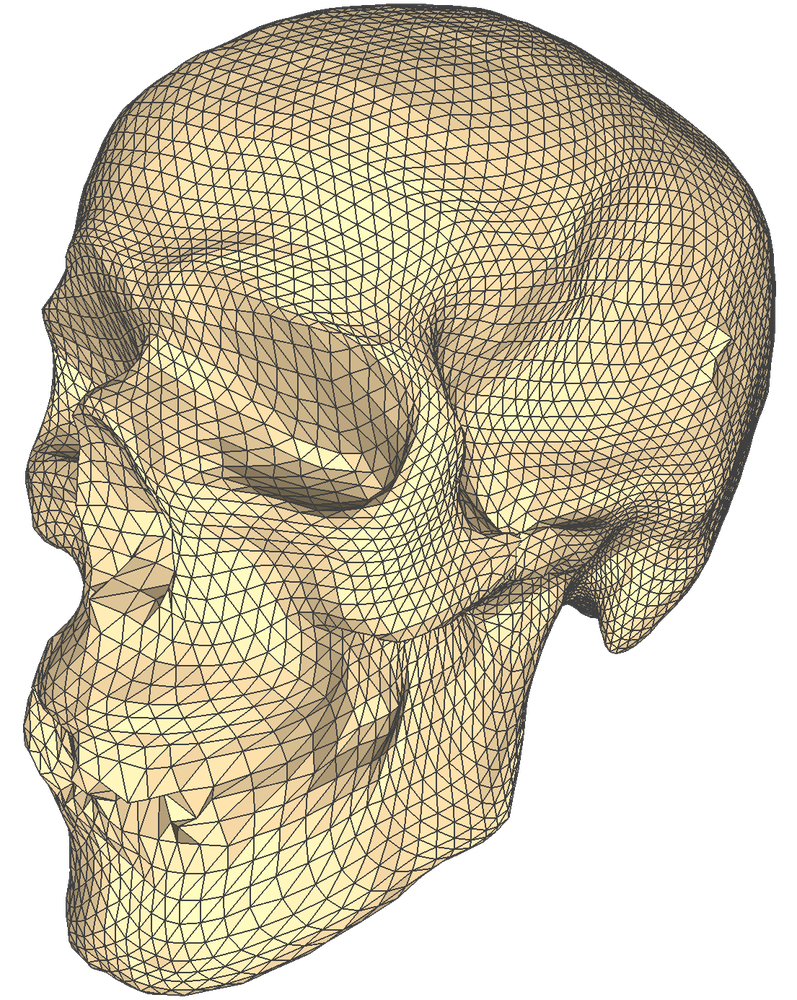} \hspace{2mm}
                            &  \includegraphics[width=2cm]{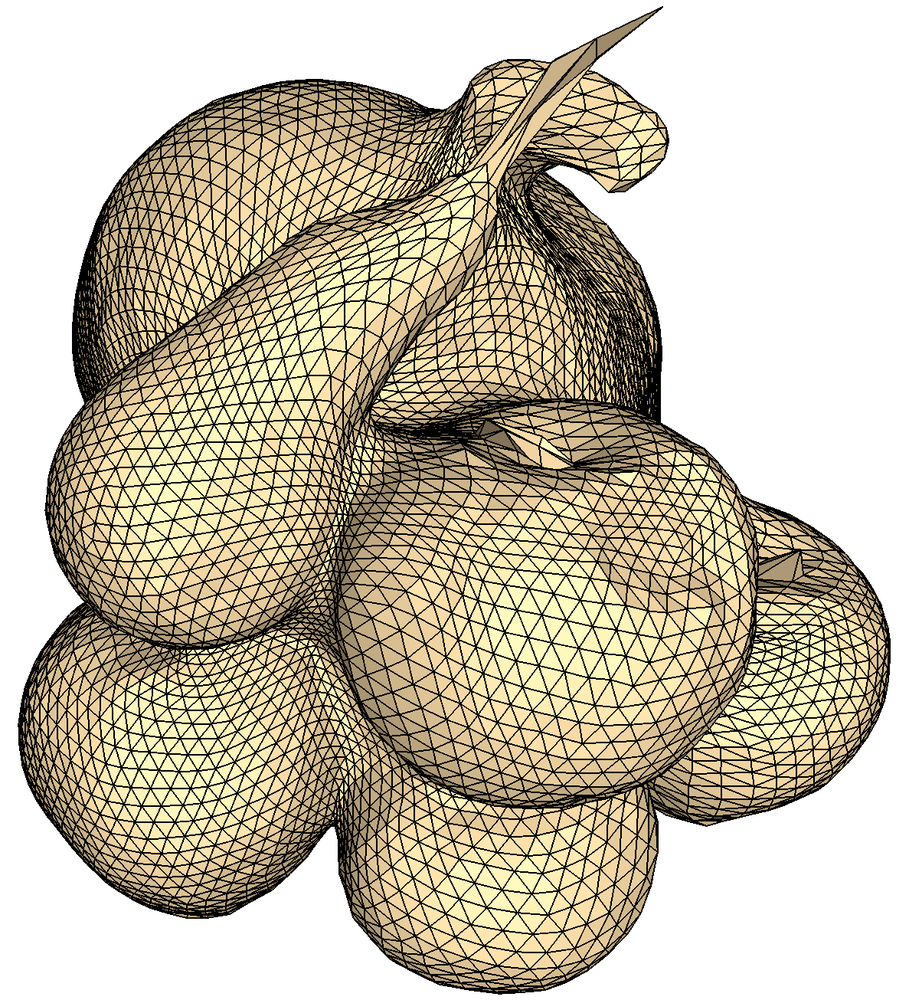} \hspace{1mm}
                            &  \includegraphics[width=2cm]{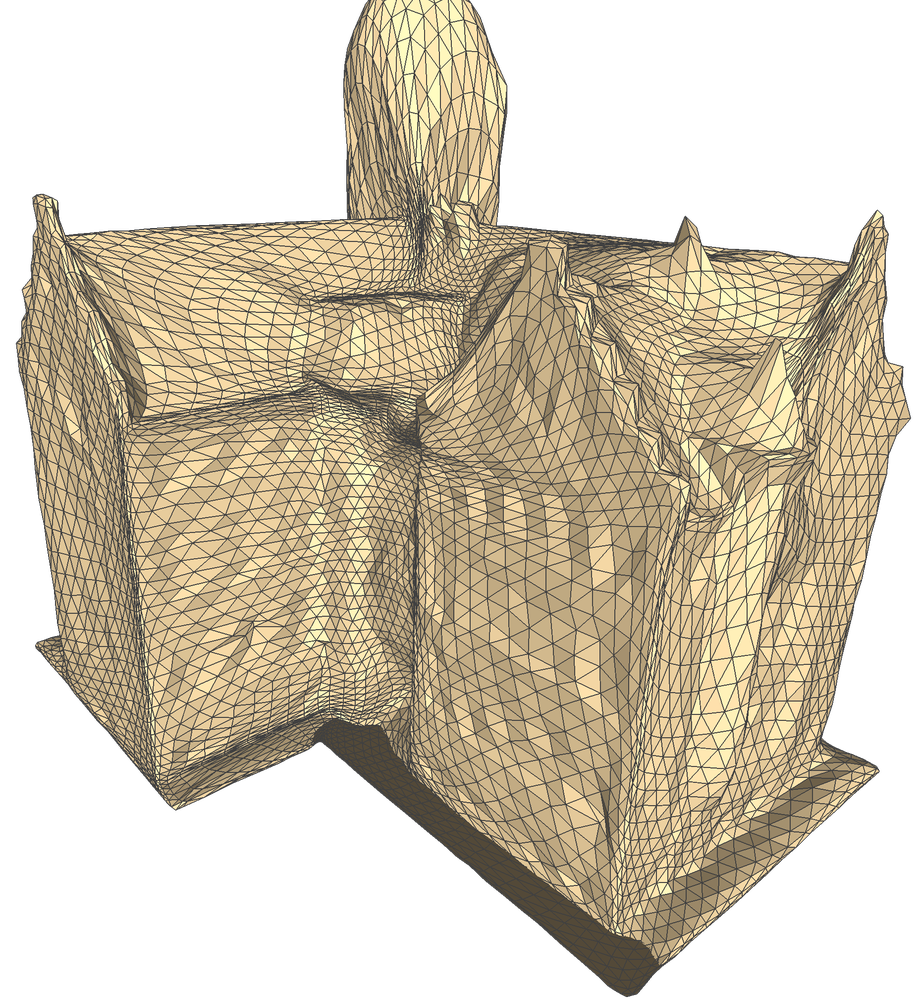}\\
      
     \rotatebox{90}{\parbox[t]{2cm}{\hspace*{\fill}\small{\text{ENS (Quad)}}\hspace*{\fill}}}\hspace*{5pt}
                            &  \includegraphics[width=1.7cm]{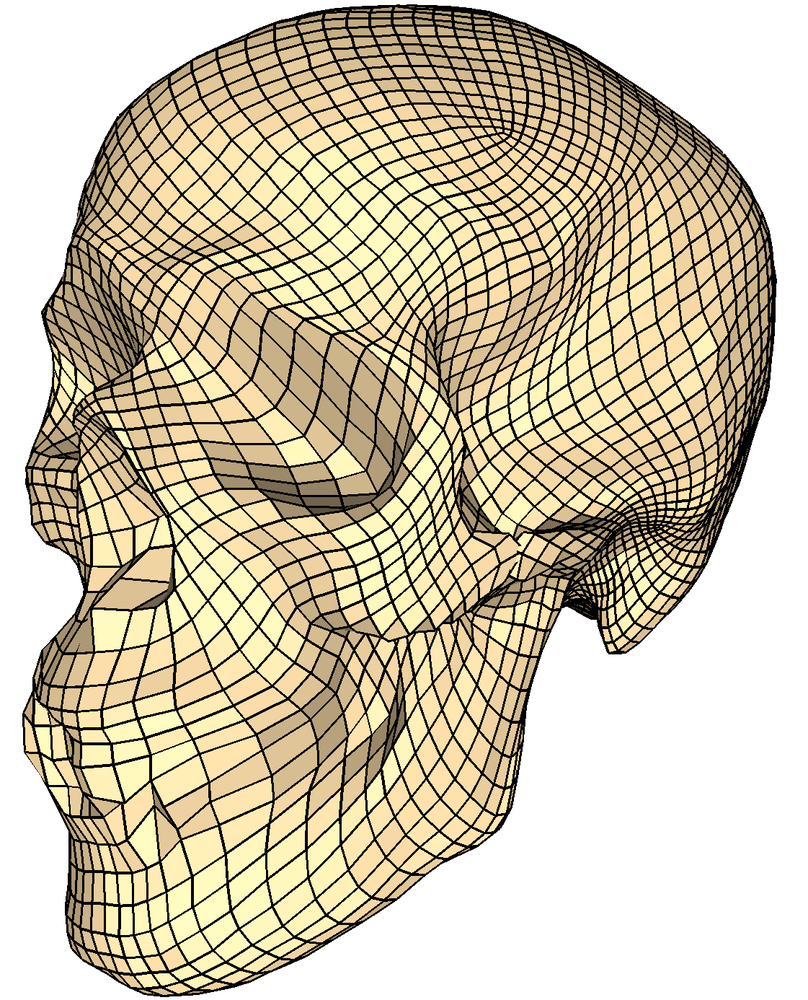} \hspace{2mm}
                            &  \includegraphics[width=2cm]{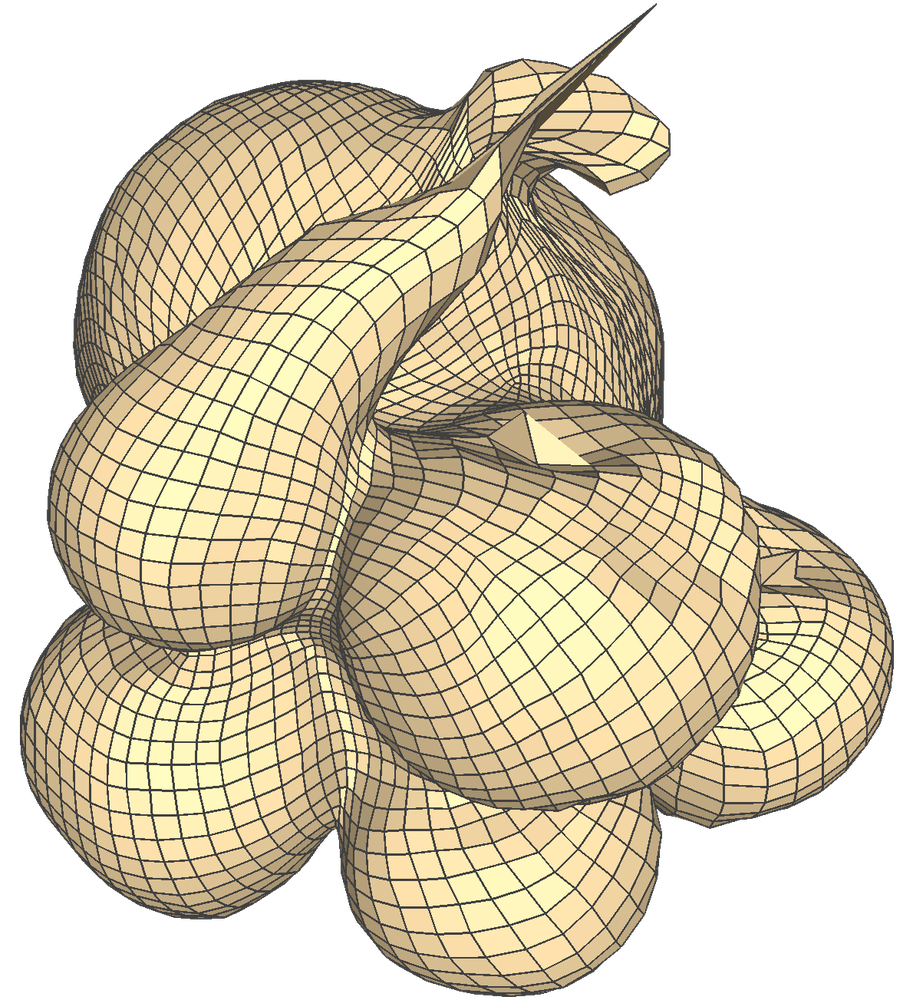} \hspace{1mm}
                            &  \includegraphics[width=2cm]{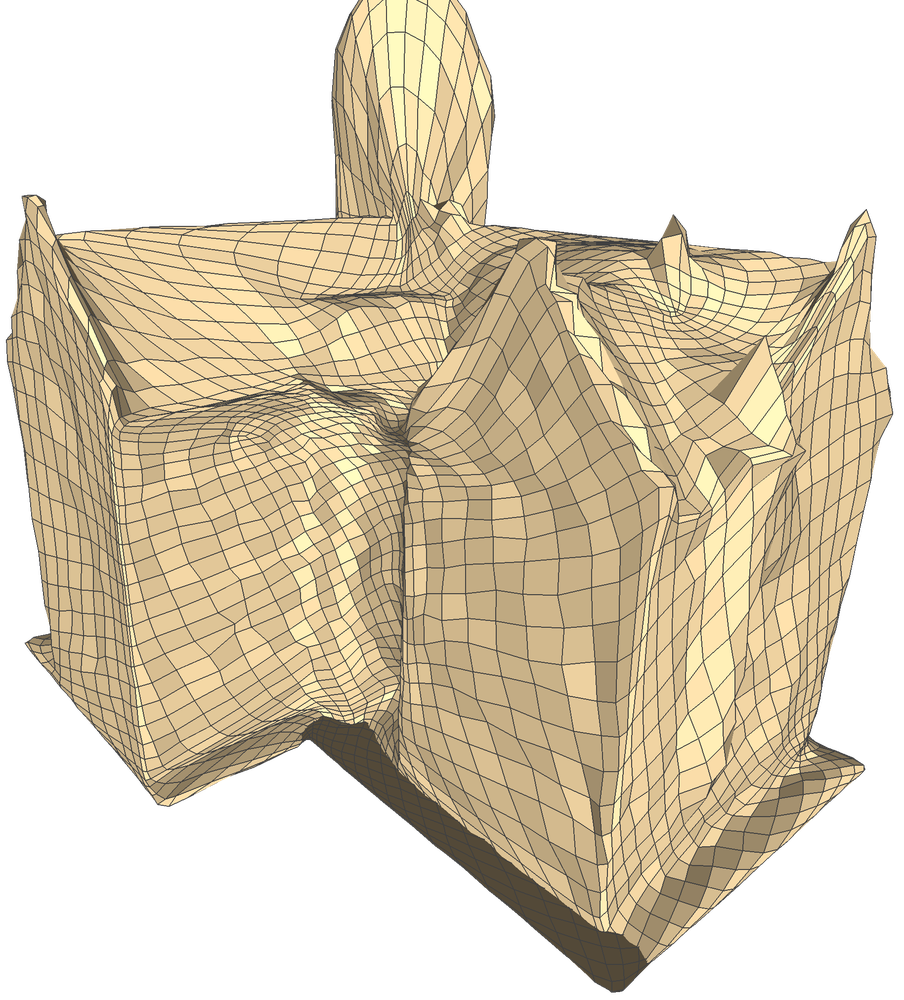}\\

    \end{tabular}
    }
    \caption{Meshes with $10$K vertices extracted from NeuS compared to ENS. Note that we use a smaller quad mesh with $6$K vertices.}
    \label{fig:10K}
\end{figure}

In \Cref{fig:2.5K} we present meshes of $\approx 2.5$K vertices extracted from NeuS and ENS on multiple scenes. Across all of the meshes extracted from NeuS we observe aliasing and reconstruction errors such as unwanted holes and floating artifacts. At this resolution, marching cubes is prone to miss sharp structures, for example the on roofs of the redhouse and or the stalk the apple. In comparison, our approach produces high-quality meshes which are suitable for downstream tasks that require coarse discretization. Increasing the resolution, in \Cref{fig:10K} we present meshes of $\approx 10$K vertices extracted from each approach. At this resolution marching cubes still produces artifacts such as floaters and noticeable ridging artifacts (see redhouse) where straight edges are not aligned to the sampling grid. Our approach can maintain straight edges for any connectivity, as we can directly sample the explicit neural surface. Note that Chamfer distance can be insensitive to these types of artifacts from marching cubes, however they have large qualitative significance.   

\begin{figure}[t]
    \centering
    \setlength{\tabcolsep}{0pt}
    \resizebox{\textwidth}{!}{
    \begin{tabular}{ccccc}
        & \multicolumn{1}{c}{}& \multicolumn{1}{c}{}& \multicolumn{1}{c}{}\\
     \rotatebox{90}{\parbox[t]{1cm}{\hspace*{\fill}\small{\text{NeuS}}\hspace*{\fill}}}\hspace*{5pt}
                            &  \includegraphics[trim={100pt 100pt 280pt 100pt
                            m},clip, width=2cm]{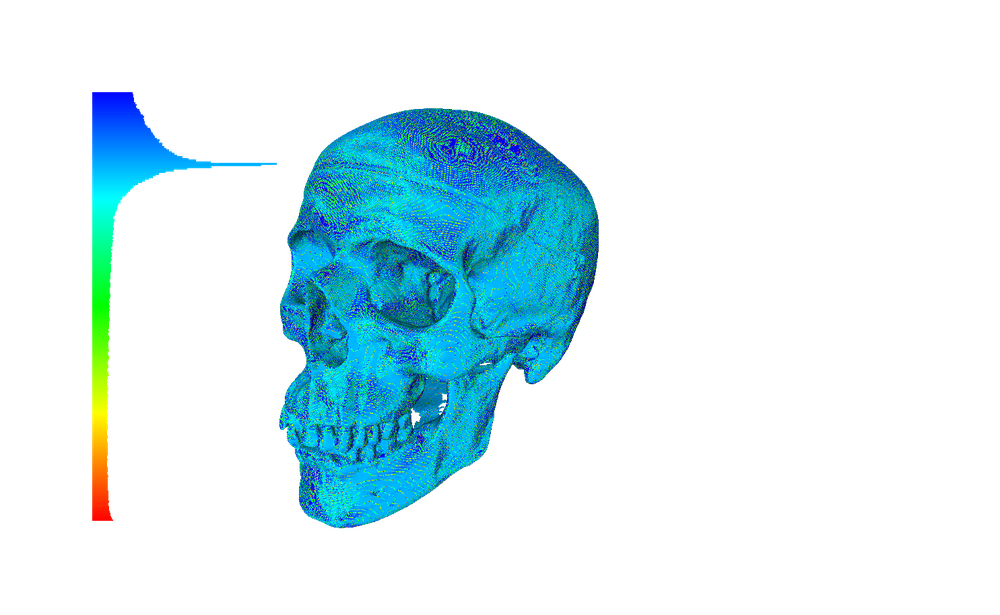}
                            &  \includegraphics[trim={100pt 100pt 280pt 100pt
                            m},clip, width=2cm]{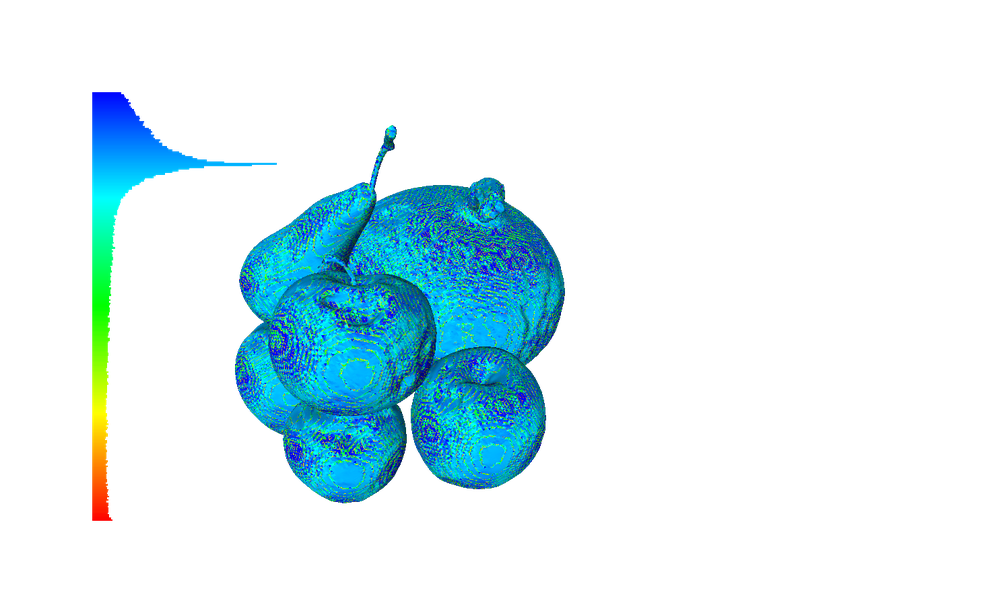}
                            &  \includegraphics[trim={100pt 100pt 280pt 100pt
                            m},clip, width=2cm]{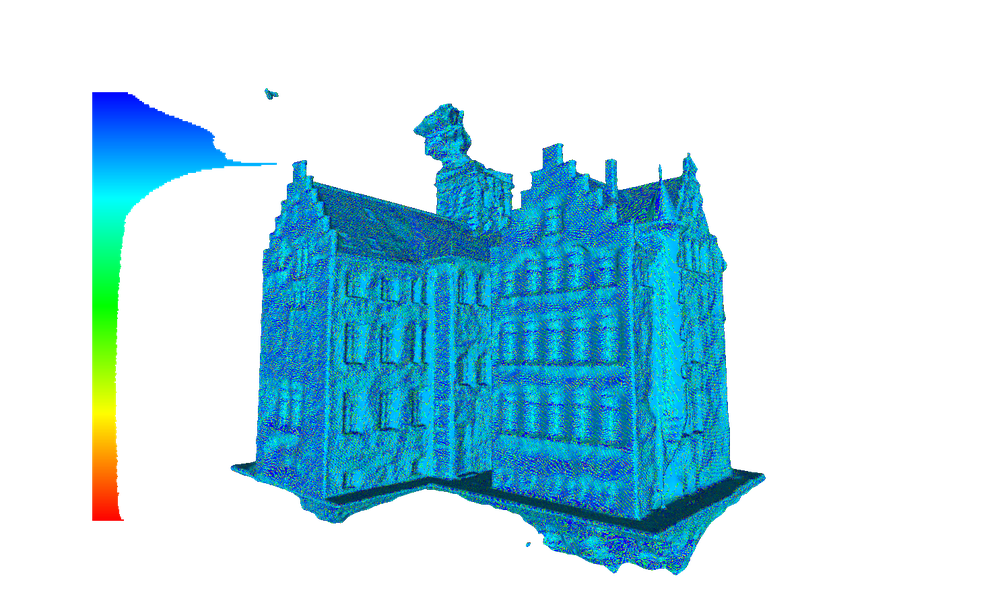}\\

                            & \tiny{Mean: 0.655} & \tiny{Mean: 0.656} & \tiny{Mean: 0.651} \\
     \rotatebox{90}{\parbox[t]{1cm}{\hspace*{\fill}\small{\text{ENS}}\hspace*{\fill}}}\hspace*{5pt}
                            &  \includegraphics[trim={100pt 100pt 280pt 100pt
                            m},clip, width=2cm]{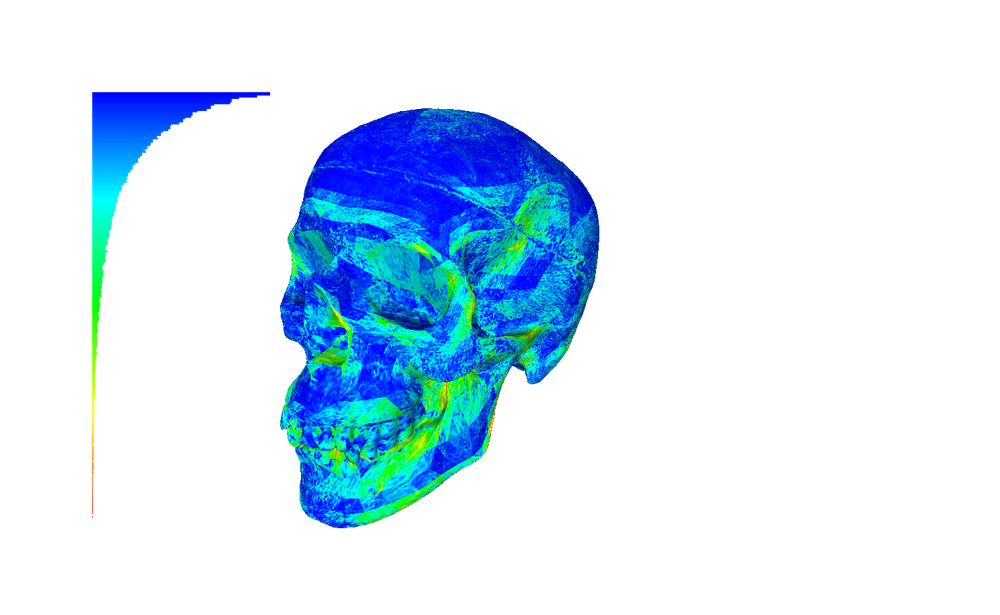}
                            &  \includegraphics[trim={100pt 100pt 280pt 100pt
                            m},clip, width=2cm]{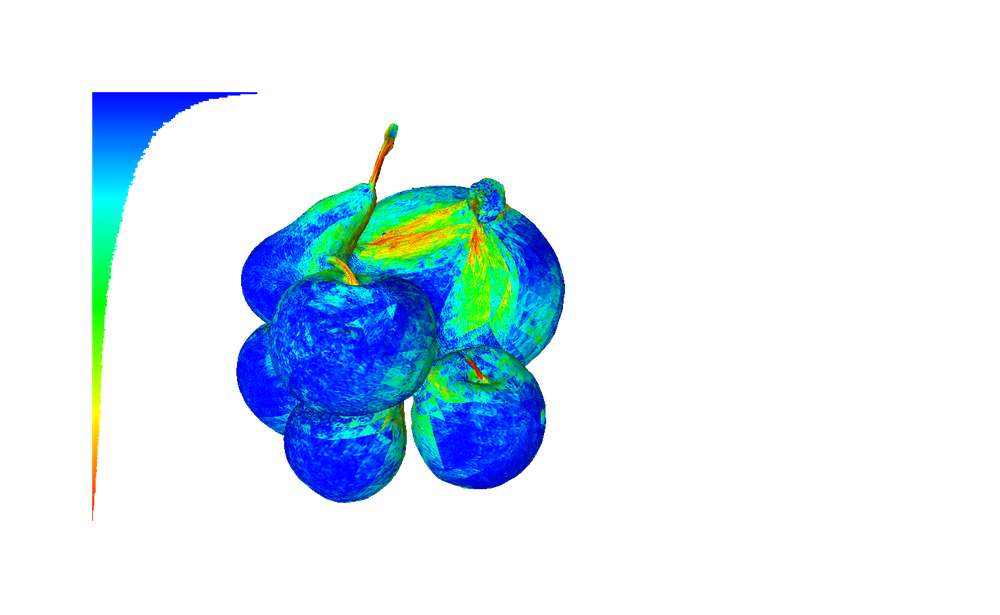}
                            &  \includegraphics[trim={100pt 100pt 280pt 100pt
                            m},clip, width=2cm]{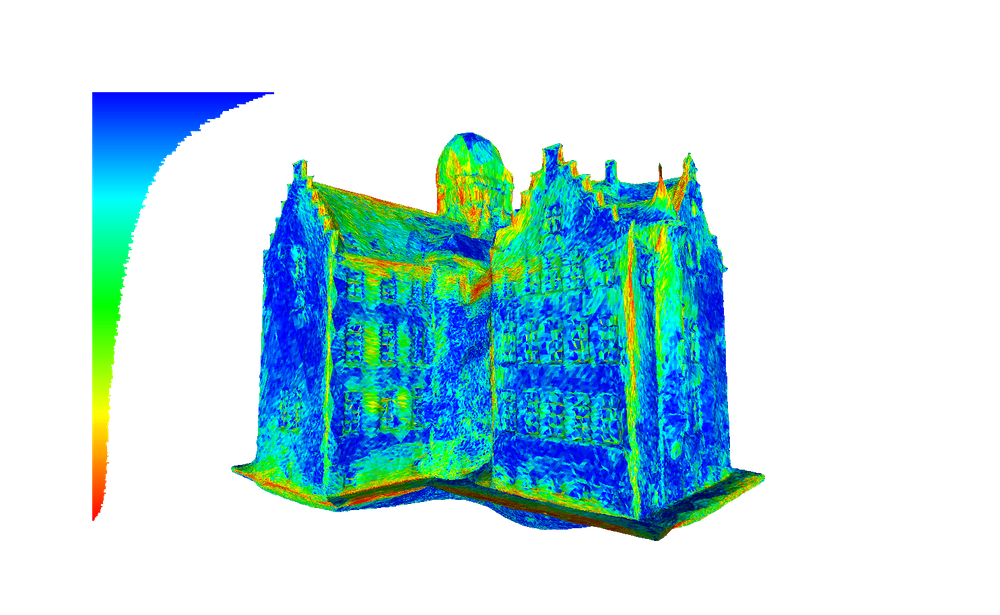}\\
                            & \tiny{Mean: 0.840} & \tiny{Mean: 0.767} & \tiny{Mean: 0.719}

    \end{tabular}
    }
    \caption{Mesh quality analysis of high-resolution meshes extracted from NeuS and ENS. Faces are coloured by normalized inradius to circumradius ratio, where ideal is 1 (blue) and degenerate is 0 (red).}
    \label{fig:mesh_qual}
\end{figure}

The artifacts of marching cubes (and the resulting poor face quality) is present even for high resolution meshes. In \Cref{fig:mesh_qual} we measure the quality of high resolution meshes ($>150$K vertices) extracted from NeuS and ENS using the per-face inradius to circumradius ratio ~\citep{shewchuk2002good}. This metric is used to measure suitability for use in finite-element-method (FEM) applications ~\citep{brandts2008equivalence}. We observe that meshes extracted from NeuS have consistently poorer face quality, as evident by the histograms and average face-quality statistics. For ENS, we note that poorer quality faces are often localized to areas of high curvature (e.g. stalks of the apples, edges of house) which require narrower fitting. For NeuS, marching cubes creates poor quality faces uniformly across the surface as a result of using a volumetric sampling grid and a fixed template of faces. When extracting meshes at low-resolution, marching cubes produces topologically inconsistent results and ridging artifacts, see \Cref{fig:multires}
    
\section{Failure Cases}

\begin{figure}[t]
\centering
\begin{tabular}{ccc}
    \includegraphics[width=0.25\linewidth]{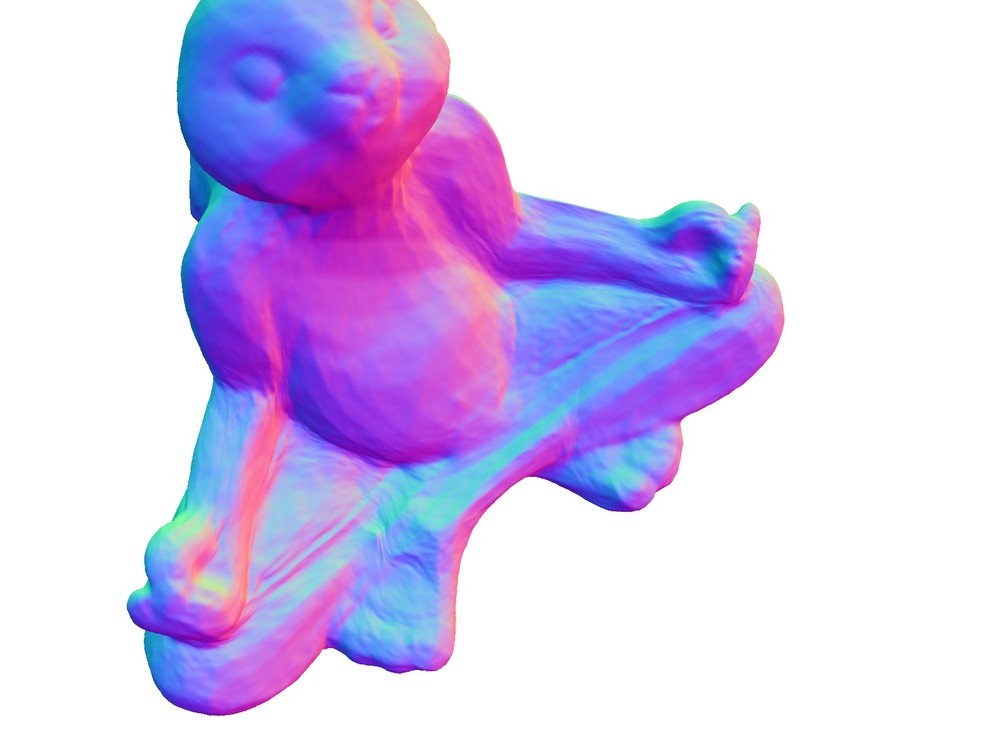} &
    \includegraphics[width=0.25\linewidth]{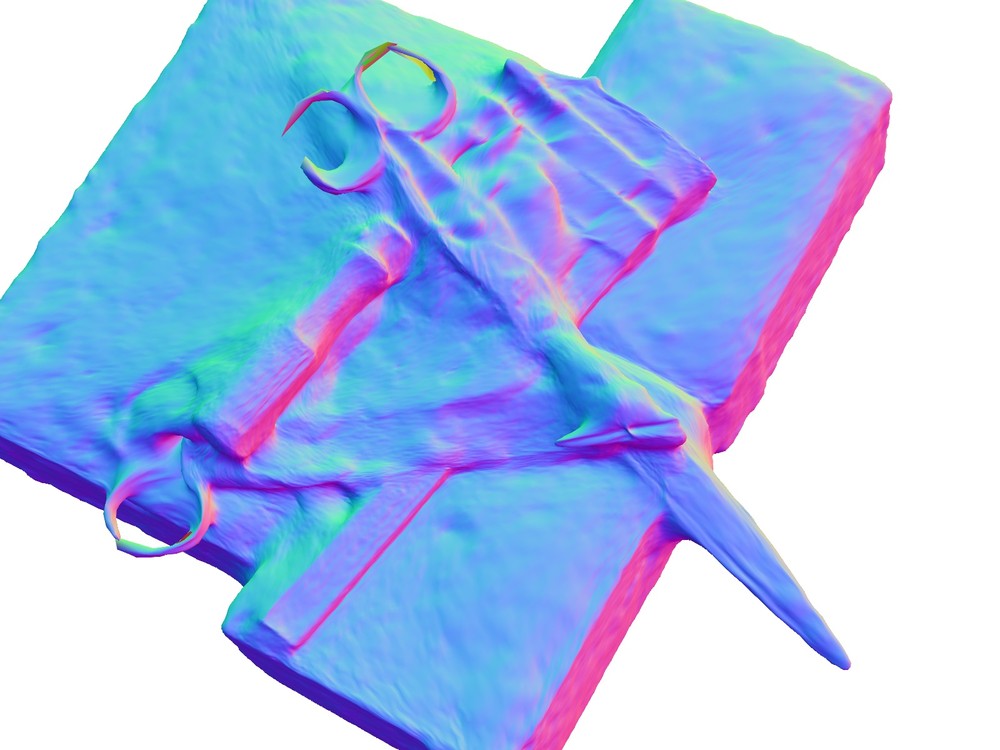} &
    \includegraphics[width=0.25\linewidth]{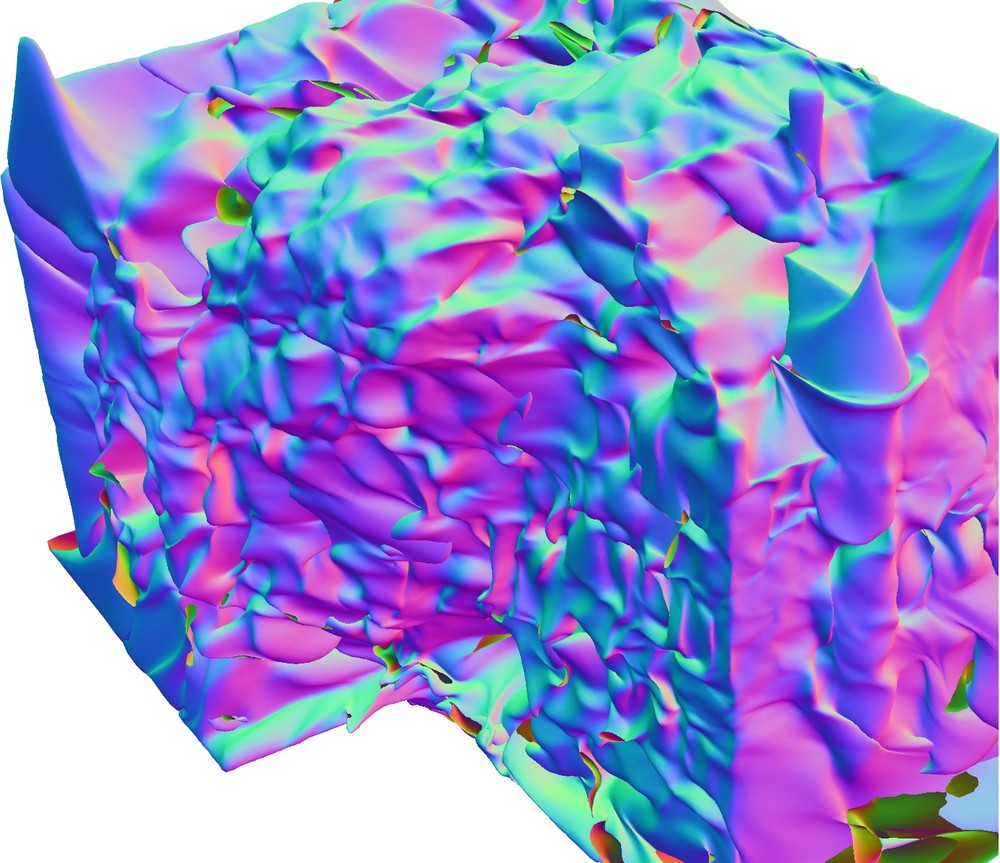} \\
    (a) 110 Rabbit & (b) 37 Scissors & (c) $f_\text{fine}$ only \\
\end{tabular}
\caption{(a)-(b) Normal maps of failed reconstructions of scene 110 and 37 of the DTU dataset. In figure (c) we show that training divergence using only a high-frequency deformation field $f_\text{fine}$.}
\label{fig:failed}
\end{figure}

In \Cref{fig:failed} we present failure cases of ENS, which contribute the largest source of Chamfer error in our quantitative performance. For scan 110 (\Cref{fig:failed} (a)) we learn a well behaved surface, however the chest area is misplaced as a result of texture ambiguities. In \Cref{fig:failed} (b) we present a failed reconstruction of scan 37. This scene has complex topology which our model struggles to approximate.

\end{document}